\documentclass{article}

\usepackage{microtype}
\usepackage{graphicx}
\usepackage{subfigure}
\usepackage{booktabs} 
\usepackage{hyperref}

\usepackage[accepted]{icml2023}

\usepackage{amsmath}
\usepackage{amssymb}
\usepackage{mathtools}
\usepackage{amsthm}

\usepackage{amsmath,amsfonts,bm}
\usepackage{amsthm}

\def\eqref#1{equation~\ref{#1}}

\def\1{\bm{1}}

\DeclareMathAlphabet{\mathsfit}{\encodingdefault}{\sfdefault}{m}{sl}
\SetMathAlphabet{\mathsfit}{bold}{\encodingdefault}{\sfdefault}{bx}{n}

\DeclareMathOperator*{\argmin}{arg\,min}

\usepackage{booktabs} 
\usepackage{amssymb}

\usepackage{enumitem}
\usepackage[capitalize,noabbrev]{cleveref}

\theoremstyle{plain}
\newtheorem{theorem}{Theorem}[section]

\newtheorem{lemma}[theorem]{Lemma}

\newtheorem{constraint}[theorem]{Constraint}
\theoremstyle{definition}
\newtheorem{definition}[theorem]{Definition}
\newtheorem{assumption}[theorem]{Assumption}
\theoremstyle{remark}

\usepackage[textsize=tiny]{todonotes}

\crefname{lemma}{lemma}{lemmas}
\Crefname{lemma}{Lemma}{Lemmas}
\crefname{thm}{theorem}{theorems}
\Crefname{thm}{Theorem}{Theorems}
\crefname{prop}{proposition}{propositions}
\Crefname{prop}{Proposition}{Propositions}
\crefname{defn}{definition}{definitions}
\Crefname{defn}{Definition}{Definitions}
\creflabelformat{equation}{#1#2#3}
\crefname{equation}{eq.}{eqs.}
\Crefname{equation}{Eq.}{Eqs.}
\Crefname{section}{\S}{\S}
\Crefname{appendix}{App.}{Apps.}
\crefname{figure}{fig.}{figs.}
\Crefname{figure}{Fig.}{Figs.}
\crefname{algorithm}{alg.}{algs.}
\Crefname{algorithm}{Alg.}{Algs.}
\crefname{assumption}{assumption}{assumptions}
\Crefname{assumption}{Assumption}{Assumptions}

\usepackage{wrapfig}

\usepackage{thmtools,thm-restate}

\newtheoremstyle{boldtheorem}
  {\topsep}
  {\topsep}
  {\itshape}
  {} 
  {\bfseries}
  {.}
  {.5em}
  {}

\theoremstyle{boldtheorem}
\newtheorem*{theorem-non}{Theorem}

\icmltitlerunning{\hfill Interventional Causal Representation Learning \hfill }

\begin{document}

\twocolumn[
\icmltitle{Interventional Causal Representation Learning}



\icmlsetsymbol{equal}{*}

\begin{icmlauthorlist}
\icmlauthor{Kartik Ahuja}{a}
\icmlauthor{Divyat Mahajan}{b}
\icmlauthor{Yixin Wang}{c}
\icmlauthor{Yoshua Bengio}{b,d}
\end{icmlauthorlist}

\icmlaffiliation{a}{FAIR (Meta AI)}
\icmlaffiliation{b}{Mila-Quebec AI Institute, Université de Montréal}
\icmlaffiliation{c}{University of Michigan}
\icmlaffiliation{d}{CIFAR Senior Fellow and CIFAR AI Chair}

\icmlcorrespondingauthor{Kartik Ahuja}{kartikahuja@meta.com}

\icmlkeywords{Machine Learning, ICML}

\vskip 0.3in
]



\printAffiliationsAndNotice{}  

\begin{abstract}
Causal representation learning seeks to extract high-level latent
factors from low-level sensory data. Most existing methods rely on
observational data and structural assumptions (e.g., conditional
independence) to identify the latent factors. However, interventional
data is prevalent across applications. Can interventional data
facilitate causal representation learning? We explore this question in
this paper. The key observation is that interventional data often
carries geometric signatures of the latent factors' support (i.e. what
values each latent can possibly take). For example, when the latent
factors are causally connected, interventions can break the dependency
between the intervened latents' support and their ancestors'.
Leveraging this fact, we prove that the latent causal factors can
be identified up to permutation and scaling given data from perfect
$do$ interventions. Moreover, we can achieve block affine
identification, namely the estimated latent
factors are only entangled with a few other latents if we have access to data from
imperfect interventions. These results highlight the unique power of
interventional data in causal representation learning; they can enable
provable identification of latent factors without any assumptions
about their distributions or dependency structure.

\end{abstract}

\section{Introduction}

Modern deep learning models like GPT-3~\citep{brown2020language} and
CLIP~\citep{radford2021learning} are remarkable representation
learners~\citep{bengio2013representation}. Despite the successes,
these models continue to be far from the human ability to adapt to new
situations (distribution shifts) or carry out new
tasks~\citep{geirhos2020shortcut,
bommasani2021opportunities, yamada2022}.  Humans encapsulate their causal
knowledge of the world in a highly reusable and recomposable
way~\citep{goyal2020inductive}, enabling them to adapt to new tasks in
an ever-distribution-shifting world.  How can we empower modern deep
learning models with this type of causal understanding? This question
is central to the emerging field of causal representation
learning~\citep{scholkopf2021towards}.

A core task in causal representation learning is \textit{provable
representation identification}, i.e., developing representation
learning algorithms that can provably identify natural latent factors
(e.g., location, shape and color of different objects in a scene).
While provable representation identification is known to be impossible
for arbitrary data-generating process
(DGP)~\citep{hyvarinen1999nonlinear, locatello2019challenging}, real
data often exhibits additional structures. For example,
\citet{hyvarinen2019nonlinear, khemakhem2020variational} consider the
conditional independence between the latents given auxiliary
information; \citet{lachapelle2022disentanglement} leverage the
sparsity of the causal connections among the latents;
\citet{locatello2020weakly, klindt2020towards, ahuja2022weakly} rely
on the sparse variation in the latents over time.

\begin{figure}[t]
\includegraphics[trim=0 3in 0 1.5in,
width=3.4in]{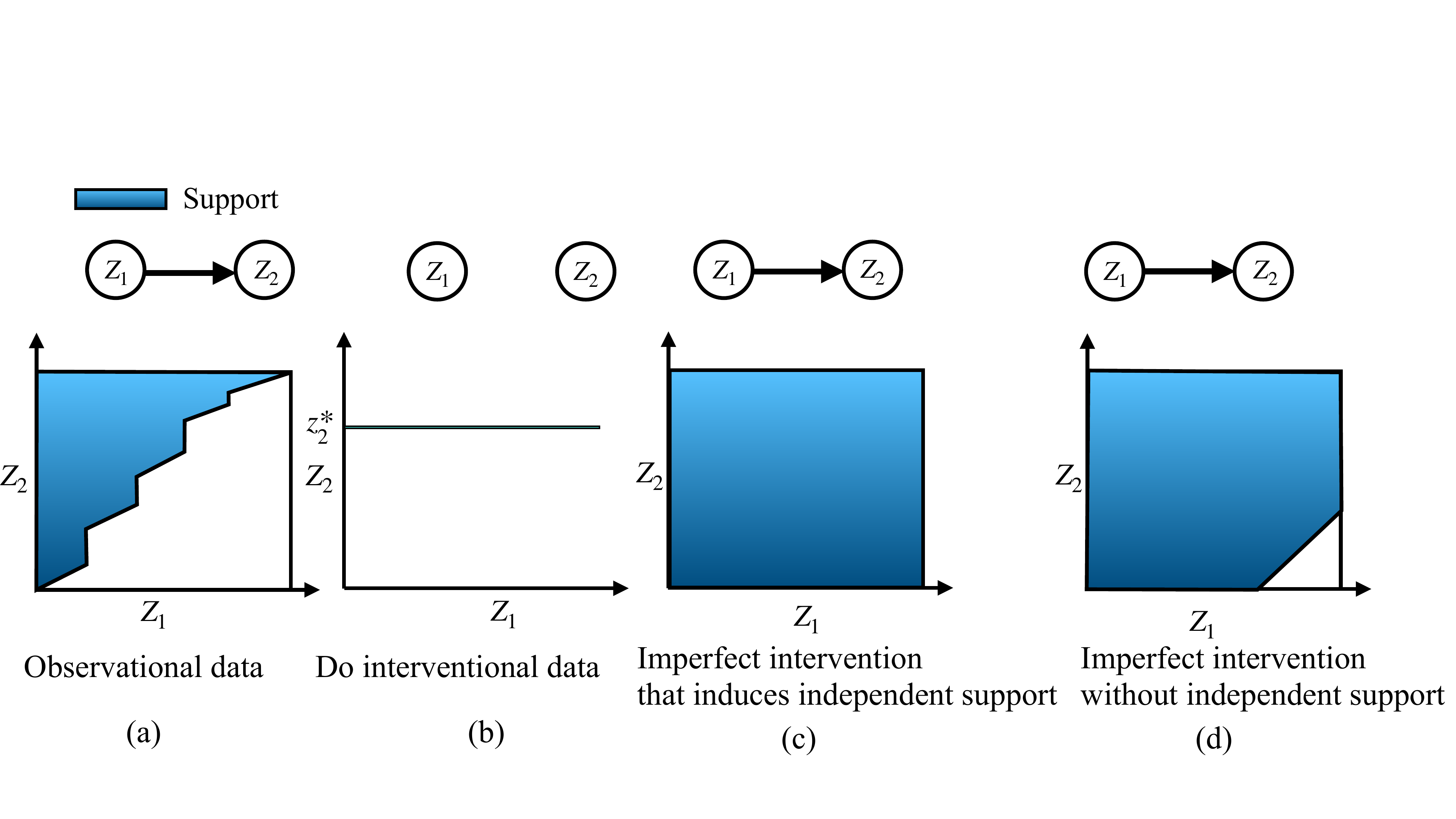}
\caption{Figure 1a) Observational data: the
support of child ($Z_2$) conditional on  parent ($Z_1$) varies with the value of parent. Figure 1b), 1c): the  support of child conditional on parent under do
intervention, perfect intervention and many imperfect interventions is
independent of the parent. Figure 1d): intervention on child reduces
the impact of the parent on it which causes the support of the child conditional on parent to
take a larger set of values.}
\label{fig1}
\end{figure}

Most existing works rely on observational data and make assumptions on the dependency structure of the latents
to achieve provable representation identification.
However, in many applications, such as robotics and genomics, there is
a wealth of interventional data available. For example, interventional
data can be obtained from experiments such as genetic
perturbations~\citep{dixit2016perturb} and electrical
stimulations~\citep{nejatbakhsh2021predicting}. \emph{Can interventional
data help identify latent factors in causal representation learning?
How can it help?} We explore these questions in this work. The key
observation is that interventional data often carries geometric
signatures of the latent factors' support (i.e., what values each
latent can possibly take). \Cref{fig1} illustrates these geometric
signatures: perfect interventions and many imperfect
interventions can make the intervened latents' support independent of
their ancestors' support. As we will show, these geometric
signatures go a long way in facilitating provable representation
identification in the absence of strong distributional assumptions. \\


\textbf{Contributions.} This work establishes representation
identification guarantees without strong distributional assumptions on the latents in the following
settings.

\begin{itemize}
\item $do$ \textbf{interventions.}  We first investigate scenarios
where the true latent factors are mapped to high-dimensional
observations through a finite-degree multivariate polynomial. When
some latent dimension undergoes a hard $do$ intervention~\citep{pearl2009causal},
we are able to identify it up to shift and
scaling. Even when the mapping is not a polynomial,
approximate identification of the intervened latent is still achievable provided we have data from multiple $do$
interventional distributions on the same latent dimension.

\item \textbf{Perfect \& imperfect interventions.} We achieve block affine identification under imperfect
interventions \citep{peters2017elements} provided the support of the intervened latent is
rendered independent of its ancestors under the intervention as shown in Figure \ref{fig1}c. This result covers all perfect interventions as a special case.

\item \textbf{Observational data and independent support.} The
independence-of-support condition above can further facilitate
representation identification with observational data. We show that,
if the support of the latents are already independent in observational
data, then these latents can be identified up to permutation, shift,
and scaling, without the need of any interventional data. This result
extends the classical identifiability results from linear independent
component analysis (ICA)~\citep{comon1994independent} to allow for
dependent latent variables. They also provide theoretical
justifications for recent proposals of performing unsupervised
disentanglement through the independent support
condition~\citep{wang2021desiderata,roth2022disentanglement}.

\end{itemize}

We summarize our results in Table \ref{table1}.
Finally, we empirically demonstrate the practical utility of our theory. From data generation mechanisms ranging from polynomials to image generation from rendering engine \citep{shinners2011pygame}, we show that interventional data  helps identification.

Also, the code repository can be accessed at: \href{https://github.com/facebookresearch/CausalRepID}{github.com/facebookresearch/CausalRepID}.

\begin{table*}
\footnotesize
	\caption{\small{Summary of results. Existing works such  as iVAE \citep{khemakhem2020variational}  use observational data and make assumptions on the graphical model of the latents to achieve identification. In contrast, we use interventional data and make no assumptions on the graph. } }
	\label{table_synth}
	\renewcommand{\arraystretch}{1.1}
	\centering
	\begin{tabular}{llll }
		\toprule
		\textbf{Input data}  & \textbf{Assm.  on}  $Z$   &  \textbf{Assm. on} $g$     &   \textbf{Identification}                     \\  \midrule
                 \textcolor{gray}{Obs} & \textcolor{gray}{$Z_r \perp Z_s | U$, $U$ aux info.}  & \textcolor{gray}{Diffeomorphic} & \textcolor{gray}{Perm \& scale (Khemakhem, 2020) } \\ 
	       Obs  &   Non-empty interior   & Injective poly & Affine (\Cref{thm1})         \\ 
               Obs &   Non-empty interior     & $\approx$ Injective poly & $\approx$ Affine (\Cref{thm1_approx_appendix})        \\  
               Obs & Independent support &  Injective poly & Perm, shift, \& scale (\Cref{thm4})\\ 
               Obs + $do$ intervn & Non-empty interior & Injective poly & Perm, shift, \& scale (\Cref{thm2})\\
               Obs + $do$ intervn & Non-empty interior & Diffeomorphic & $\approx$ Perm \& comp-wise (\Cref{thm_do_appendix})\\
               Obs + Perfect intervn & Non-empty interior & Injective poly & Block affine (\Cref{thm3})\\ 
                Obs + Imperfect intervn &  Partially indep. support & Injective poly & Block affine (\Cref{thm3}) \\ 
                \textcolor{gray}{Counterfactual } & \textcolor{gray}{Bijection w.r.t. noise} & \textcolor{gray}{Diffeomorphic} & \textcolor{gray}{Perm \& comp-wise (Brehmer, 2022)} \\ 
		\bottomrule
	\end{tabular}
 \label{table1}
\end{table*}

\section{Related Work}

Existing provable representation identification approaches often
utilize structure in time-series data, as seen in initial works by
\citet{hyvarinen2016unsupervised} and \citet{hyvarinen2017nonlinear}.
More recent studies have expanded on this approach, such as
\citet{halva2020hidden, yao2021learning, yao2022learning, yao2022learning1, lippe2022citris, lippe2022icitris, lachapelle2022disentanglement}. Other forms of
weak supervision, such as data augmentations, can also be used in
representation identification, as seen in works by
\citet{zimmermann2021contrastive, von2021self, brehmer2022weakly,
locatello2020weakly, ahuja2022weakly} that assume access to
contrastive pairs of observations $(x, \tilde{x})$. A third approach,
used in \citep{khemakhem2020variational, khemakhem2020ice}, involves
using high-dimensional observations (e.g., an image) and auxiliary
information (e.g., label) to identify representations.

To understand the factual and counterfactual knowledge used by
different works in representation identification, we can classify them
according to Pearl's ladder of causation~\citep{bareinboim2022pearl}.
In particular, our work operates with interventional data (level-two
knowledge), while other studies leverage either observational data
(level-one knowledge) or counterfactual data (level-three knowledge).
Works such as \citet{khemakhem2020variational, khemakhem2020ice,
ahuja2022towards, hyvarinen2016unsupervised, hyvarinen2017nonlinear,
ahuja2021properties} use observational data and either make 
assumptions on the structure of the underlying causal graph of latents
or rely on auxiliary information. In contrast, works like
\citet{brehmer2022weakly} use counterfactual knowledge to achieve
identification for general DAG structures; \citet{lippe2022citris,lippe2022icitris,
ahuja2022weakly, lachapelle2022disentanglement} use pre- and
post-intervention observations to achieve provable representation
identification. These latter studies use instance-level temporal interventions
that carry much more information than interventional distribution
alone. To summarize, these works require more information than is available with level two data in Pearlian ladder of causation.

Finally, a concurrent work from \citet{seigal2022linear} also studies identification of causal representations using interventional distributions. The authors focus on linear mixing of the latents and consider perfect interventions. In contrast, our results consider nonlinear mixing function and imperfect interventions.
\section{Setup: Causal Representation Learning}

Causal representation learning aims to identify latent variables from
high-dimensional observations. Begin with a data-generating process
where some high-dimensional observations $x \in \mathbb{R}^{n}$ are
generated from some latent variables $z \in \mathbb{R}^{d}$. We
consider the task of identifying latent $z$ assuming access to both
observational and interventional datasets: the observational data is
drawn from
\begin{align}
z \sim \mathbb{P}_Z; \qquad  x \leftarrow g(z), 
\label{eqn: dgp}
\end{align}
where the latent $z$ is sampled from the distribution $\mathbb{P}_Z$
and $x$ is the observed data point rendered from the underlying latent
$z$ via an injective decoder $g:\mathbb{R}^{d}\rightarrow
\mathbb{R}^{n}$. The interventional data is drawn from a similar
distribution except the latent $z$ is drawn from $\mathbb{P}^{(i)}_Z$,
namely the distribution of $z$ under intervention on   $z_i$:
\begin{align}
z \sim \mathbb{P}_Z^{(i)}; \qquad  x \leftarrow g(z).
\label{eqn: dgp_int}
\end{align}
We denote $\mathcal{Z}$ and $\mathcal{Z}^{(i)}$ as the support of
$\mathbb{P}_Z$ and $\mathbb{P}_Z^{(i)}$ respectively (support is the set where the probability density is more than zero). The support of
$x$ is thus $\mathcal{X} = g(\mathcal{Z})$ in observational data and
$\mathcal{X}^{(i)} = g\big(\mathcal{Z}^{(i)}\big)$ in interventional
data.
The goal of causal representation learning is \textit{provable
representation identification}, i.e. to learn an encoder function,
which takes in the observation $x$ as input and provably output
its underlying true latent $z$. In practice, such an encoder is often
learned via solving a reconstruction identity,
\begin{align}
    h \circ f(x) = x \qquad \forall x\in \mathcal{X}\cup\mathcal{X}^{(i)},
    \label{eqn: recons}
\end{align}
where $f:\mathbb{R}^{n} \rightarrow
\mathbb{R}^{d}$ and $h:\mathbb{R}^{d}\rightarrow
\mathbb{R}^{n}$ are a pair of encoder and decoder, which need to
jointly satisfy \Cref{eqn: recons}.
The pair $(f,h)$ together is referred to as the autoencoder. Given the
learned encoder $f$, the resulting representation is $\hat{z}
\triangleq f(x)$, which holds the encoder's estimate of the latents.

The reconstruction identity \Cref{eqn: recons} is highly
underspecified and cannot in general identify the latents. There exist
many pairs of $(f,h)$ that jointly solve \Cref{eqn: recons} but do not
provide representations $\hat{z} \triangleq f(x)$ that coincide with
the true latents $z$. For instance, applying an invertible map $b$ to
any solution $(f,h)$ will result in another valid solution $b \circ
f$, $h\circ b^{-1}$. In practical applications, however, the exact
identification of the latents is not necessary. For
example, we may not be concerned with the recovering the latent dimensions in the order they appear in $z$. Thus, in this
work, we examine conditions of under which the true latents can be identified up to
certain transformations, such as affine transformations and coordinate
permutations.

\section{Stepping Stone: Affine Representation Identification with
Polynomial Decoders}

We first establish an affine identification result, which serves as a stepping stone towards stronger identification guarantees in the next section. We begin with a few assumptions.

\begin{assumption}
\label{assm1: dgp} The interior of the support of $z$, $\mathcal{Z}\cup\mathcal{Z}^{(i)}$, is a non-empty subset of $\mathbb{R}^{d}$.\footnote{We work with $(\mathbb{R}^{d}, \|\|_2$) as the metric space. A point is in the interior of a set if there exists an~$\epsilon$ ball for some $\epsilon>0$ containing that point in the set. The set of all such points defines the interior.}
\end{assumption}

\begin{assumption}
\label{assm2: g_poly} 
The decoder $g$ is a polynomial of finite degree $p$ whose corresponding coefficient matrix $G$ has full column rank. Specifically, the decoder $g$ is determined by the coefficient matrix $G$ as follows,
 $$ g(z) = G[1, z, z\bar{\otimes} z, \cdots, \underbrace{z \bar{\otimes} \cdots \bar{\otimes}\;z}_{p \;\text{times}}]^{\top} \qquad \forall z\in \mathbb{R}^d,$$
where $ \bar{\otimes} $ represents the Kronecker product  with all distinct entries; for example, if $z = [z_1, z_2]$, then~$z\bar{\otimes} z = [z_1^2, z_1z_2, z_2^2]$. 
\end{assumption}

The assumption that the matrix $G \in \mathbb{R}^{n \times q}$ has a
full column rank of $q$ guarantees that the decoder $g$ is injective;
see Lemma~\ref{lemma1} in Appendix \ref{sec:affine_appendix} for a proof. This
injectivity condition on $g$ is common in identifiable representation
learning. Without injectivity, the problem of identification becomes
ill-defined; multiple different latent $z$'s can give rise to the same
observation $x$. We note that the full-column-rank condition for $G$ in \Cref{assm2:
g_poly} imposes an implicit constraint on the dimensionality $n$ of
the data; it requires that the dimensionality $n$ is greater than the
number of terms in the polynomial of degree $p$, namely $n \geq
\sum_{r=0}^{p} { r+d-1 \choose d-1}$. In the Appendix
(\Cref{thm1_appendix_sparse}), we show that if our data is generated
from sparse polynomials, i.e., $G$ is a sparse matrix, then $n$ is allowed to be much smaller. 

Under \Cref{assm1: dgp,assm2: g_poly}, we perform causal
representation learning with two constraints: polynomial decoder and
non-collapsing encoder.


\begin{constraint}
\label{assm3: h_poly_new} 
The learned decoder $h$ is a polynomial of degree $p$ and it is determined by its
corresponding coefficient matrix $H$ as follows, $$  h(z) = H[1, z, z\bar{\otimes} z, \cdots, \underbrace{z \bar{\otimes} \cdots \bar{\otimes}\;z}_{p \;\text{times}}]^{\top}\qquad \forall z\in \mathbb{R}^d,$$
where $ \bar{\otimes} $ represents the Kronecker product  with all distinct entries. The interior of the image of the encoder $ f(\mathcal{X} \cup \mathcal{X}^{(i)})$ is a non-empty subset of $\mathbb{R}^d$.
\end{constraint}

We now show that solving the reconstruction identity with these constraints can provably identify the true latent $z$ up to affine transformations.

\begin{restatable}{theorem}{AffineIdThm}
\label{thm1}
Suppose the observational data and interventional data are generated
from  \Cref{eqn: dgp} and \Cref{eqn: dgp_int} respectively under Assumptions \ref{assm1: dgp} and \ref{assm2: g_poly}. The autoencoder that solves the reconstruction
identity in \Cref{eqn: recons} under \Cref{assm3: h_poly_new} achieves affine identification, i.e.,
$\forall z \in \mathcal{Z} \cup \mathcal{Z}^{(i)}, \hat{z} = Az + c$,
where $\hat{z}$ is the encoder $f$'s output, $z$ is the true
latent, $A \in \mathbb{R}^{d\times d}$ is  invertible  and $c\in
\mathbb{R}^{d}$.
\end{restatable}

\Cref{thm1} drastically reduces the ambiguities in identifying latent
$z$ from arbitrary invertible transformations to only invertible affine
transformations. Moreover, \Cref{thm1} does not require any structural
assumptions about the dependency between the latents. It only requires
(i) a geometric assumption that the interior of the support is
non-empty and (ii) the map $g$ is a finite-degree polynomial.

The proof of \Cref{thm1} is in Appendix \ref{sec:affine_appendix}. The idea is to
write the representation $\hat{z} = f(x)$ as $\hat{z} = f\circ g (z) =
a(z)$ with~$a \triangleq  f \circ g$, leveraging the relationship
$x=g(z)$ in \Cref{eqn: dgp}. We then show the $a$ function must
be an affine map. To give further intuition, we consider a toy example with
one-dimensional latent $z$, three-dimensional observation $x$, and the
true decoder $g$ and the learned decoder $h$ each being a degree-two
polynomial.  We first solve the reconstruction identity on all $x$,
which gives $h(\hat{z}) = g(z)$, and equivalently $H[1,
\hat{z}, \hat{z}^2]^\top = G[1, z, z^2]^{\top}$. This equality implies
that both $\hat{z}$ and $\hat{z}^2$ must be at most degree-two
polynomials of $z$. As a consequence, $\hat{z}$ must be a degree-one
polynomial of $z$, which we next prove by contradiction. If $\hat{z}$
is a degree-two polynomial of $z$, then $\hat{z}^2$ is degree four; it
contradicts the fact that $\hat{z}^2$ is at most degree two in $z$.
Therefore, $\hat{z}$ must be a degree-one polynomial in $z$, i.e. a
linear function of $z$.

\paragraph{Beyond polynomial map $g$.}  
 \Cref{thm1_approx_appendix}
in the Appendix extends \Cref{thm1} to a class of maps $g(\cdot)$ that
are $\epsilon$-approximable by a polynomial.

\section{Provable Representation Identification with Interventional Data}
\label{sec:intervention}
In the previous section, we derived affine identification guarantees. Next, we strengthen these guarantees by leveraging geometric signals specific to many interventions.

\subsection{Representation identification with $do$ interventions}

\begin{figure}
    \centering
    \includegraphics[trim =0 7in 0 0, width=3.5in]{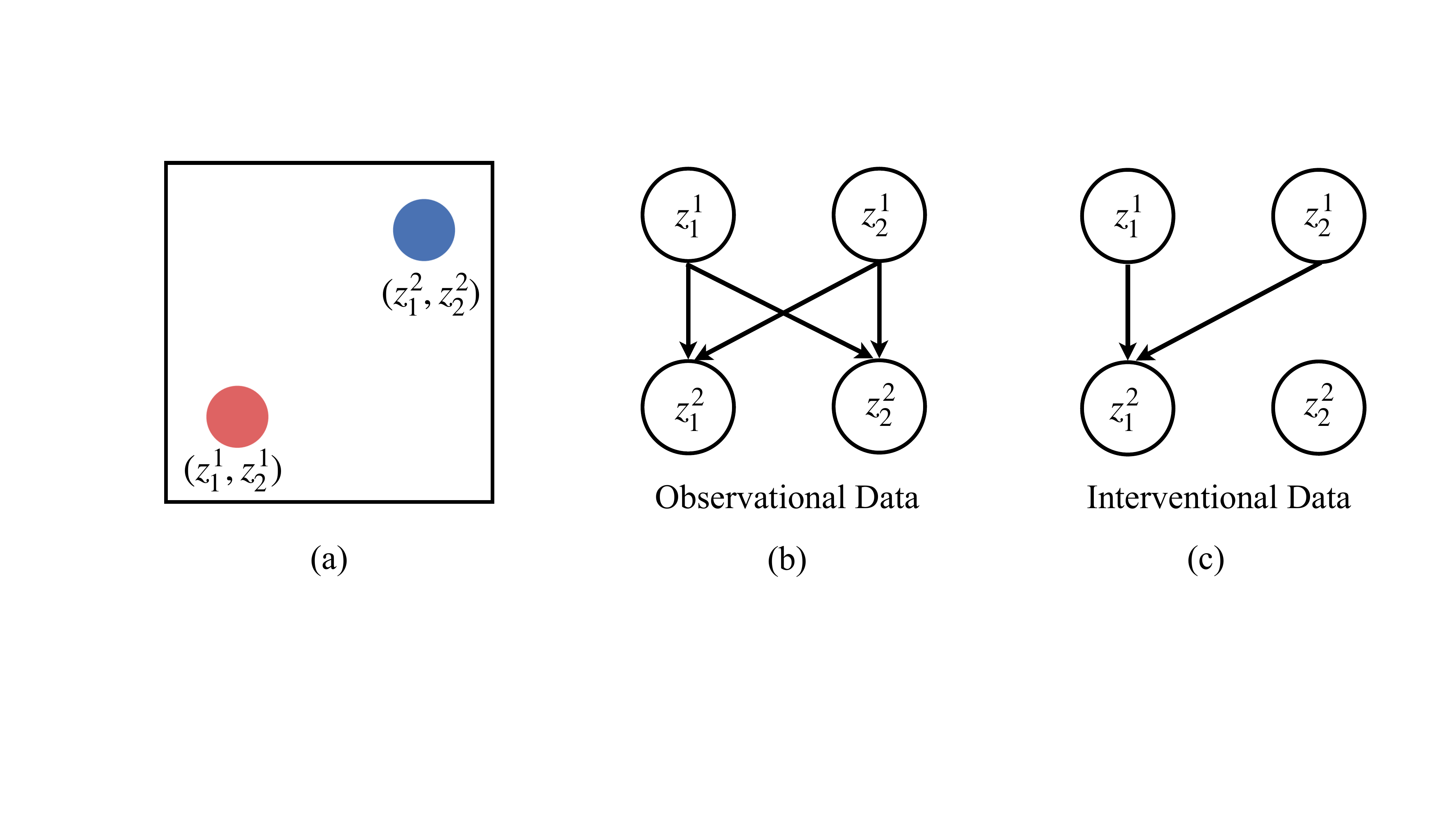}
    \caption{Illustrating $do$ interventions in image-based data in (a). The DAG of dependencies under the observational distribution (b) and a perfect intervention on $z_2^2$ in (c).}
    \label{fig2}
\end{figure}

We begin with a motivating example on images, where we  are given data with $do$ interventions on the latents.  Consider the two balls shown in \Cref{fig2}a. Ball $1$'s coordinates are $(z_1^{1}, z_2^{1})$ and Ball $2$'s coordinates are $(z_1^{2}, z_2^{2})$.  We write the latent $z = [ (z_1^{1}, z_2^{1}), (z_1^{2}, z_2^{2})]$, this latent is rendered in the form of the image $x$ shown in the \Cref{fig2}a. The latent $z$ in the observational data follows the directed acyclic graph (DAG) in \Cref{fig2}b, where Ball $1$'s coordinate cause the Ball $2$ coordinates. The latent $z$ under a $do$ intervention on $z_2^2$, then the second coordinate of Ball $2$, follows the DAG in \Cref{fig2}c. Our goal is to learn an encoder using the images $x$ in observational and interventional data, which outputs the coordinates of the balls up to permutation and scaling.

Suppose $z$ is generated from a structural causal model with an underlying DAG~\citep{pearl2009causal}. Formally, a $do$ intervention on one latent dimension fixes it to some constant value. The distribution of the children of the intervened component is affected by the intervention, while the distribution of remaining latents remains unaltered. Based on this property of $do$ intervention, we characterize the distribution $\mathbb{P}_{Z}^{(i)}$ in \Cref{eqn: dgp_int} as
 \begin{align}
         z_{i} = z^{*}; \qquad         z_{-i} \sim \mathbb{P}_{Z_{-i}}^{(i)}, 
     \label{eqn: do_int}
 \end{align}
 where  $z_i$ takes a fixed value $z^{*}$. The remaining variables in $z$, $z_{-i}$, are sampled from $\mathbb{P}_{Z_{-i}}^{(i)}$. 

 The distribution $\mathbb{P}_{Z_{-i}}^{(i)}$ in \Cref{eqn: do_int} encompasses many settings in practice, including (i) the $do$
 interventions on causal DAGs~\citep{pearl2009causal}, i.e., $\mathbb{P}_{Z_{-i}}^{(i)} =
 \mathbb{P}_{Z_{-i}|\textit{do}(z_i=z^{*})}$, ii) the $do$ interventions on cyclic graphical models~\citep{mooij2013cyclic}, and (iii) sampling $z_{-i}$ from its conditional in the observational data  $\mathbb{P}_{Z_{-i}}^{(i)} = \mathbb{P}_{Z_{-i}|z_i=z^{*}}$ (e.g., subsampling images in observational data with a fixed background color).

Given interventional data from $do$ interventions, we perform causal
representation learning by leveraging the geometric signature of the
$do$ intervention in search of the autoencoder. In particular,
we enforce the following constraint while solving the reconstruction
identity in \Cref{eqn: recons}.

\begin{constraint} \label{eqn: intv_cons}
The encoder's $k^{th}$ component $f_k(x)$ denoted as $\hat{z}_k$  is required to take some fixed value $z^{\dagger}$ for all $x \in \mathcal{X}^{(i)}$. Formally stated $f_k(x) = z^{\dagger}, \forall x \in \mathcal{X}^{(i)}$.
\end{constraint}

In \Cref{eqn: intv_cons}, we do not need to know which component is intervened and the value it takes, i.e., $k\not=i$ and $z^{\dagger}\not=z^{*}$. We next show how this constraint
helps identify the intervened latent $z_i$ under an additional  assumption on the support of the unintervened
latents stated below.

 \begin{assumption}
\label{assm:span} The interior of support of distribution of unintervened latents  $\mathbb{P}_{Z_{-i}}^{(i)}$ is a non-empty subset of $\mathbb{R}^{d-1}$. 
\end{assumption}

\begin{restatable}{theorem}{DoItvID}
\label{thm2} 
Suppose the observational data and interventional data are generated from  \Cref{eqn: dgp} and \Cref{eqn: dgp_int} respectively under \Cref{assm1: dgp,assm2: g_poly}, where $\mathbb{P}_{Z}^{(i)}$ follows  \Cref{eqn: do_int}. The autoencoder that solves \Cref{eqn: recons} under \Cref{assm3: h_poly_new}, \Cref{eqn: intv_cons} identifies the intervened latent $z_{i}$  up to shift and scaling, i.e., $\hat{z}_k = e z_i + b$,  where $e \in \mathbb{R}, b \in \mathbb{R}$. 
\end{restatable}

\Cref{thm2} immediately extends to settings when multiple
interventional distributions are available, with each corresponding to
a hard $do$ intervention on a distinct latent variable.  Under the
same assumptions of \Cref{thm2}, each of the intervened latents can be
identified up to permutation, shift, and scaling. Notably, \Cref{thm2} does
not rely on any distributional assumptions (e.g., parametric
assumptions) on $z$; nor does it rely on the nature of the graphical model
for $z$ (e.g., cyclic, acyclic). \Cref{thm2} makes these key geometric
assumptions: (i) support of $z$ in observational data, (ii) support of
unintervened latents $z_{-i}$ has a non-empty interior.

\Cref{thm2} combines the affine identification guarantee we derived in
\Cref{thm1} with the geometric signature of $do$ interventions. For
example, in \Cref{fig1}b, the
support of the true latents is axis-aligned (parallel to x-axis). In this case, the interventional constraint also forces
the support of $\hat{z}$ to be axis-aligned (parallel to x-axis or y-axis).  The proof of \Cref{thm2} is in Appendix \ref{sec: rep_id_do}. We provide
some intuition here. First, given \Cref{assm1: dgp,assm2:
g_poly,assm3: h_poly_new}, \Cref{thm1} already guarantees affine
identification. It implies $\hat{z}_k = a_{-i}^{\top}z_{-i} + ez_i
+ b$, where $z_{-i}$ includes all entries of $z$ other than $z_i$, and
  $a_{-i}$ is a vector of the corresponding coefficients. As a result,
  $a_{-i}^{\top}z_{-i}$ must also take a fixed value for all values of
  $z_{-i}$ in the support of $\mathbb{P}_{Z_{-i}}^{(i)}$, since both
  $\hat{z}_k$ and $z_i$ are set to a fixed value. We argue
  $a_{-i}=0$ by contradiction. If $a_{-i}\ne 0$, then any changes to
  $z_{-i}$ in the direction of $a_{-i}$ will also reflect as a change
  in $\hat{z}_k$; it contradicts the fact that $\hat{z}_k$ takes a
  fixed value. Therefore, $a_{-i}=0$ and $z_i$ is identified up to
  shift and scaling.

\paragraph{Beyond polynomial map $g$.} In \Cref{thm2}, we assume that the
map $g$ is a polynomial. In the Appendix (\Cref{thm_do_appendix}) we
show that, even when $g$ is not a polynomial but a general
diffeomorphism, the intervened latent can be approximately identified
up to an invertible transform provided sufficiently many $do$ interventional distributions
per latent are available. That said, one interventional distribution per latent no
longer suffices, unlike the polynomial $g$ case. Our experiments on images in \Cref{sec: emp} further support this argument. We state \Cref{thm_do_appendix} informally below.

\begin{theorem-non}(Informal)
 Suppose the observational data is generated from  \Cref{eqn: dgp} and suppose we gather multiple interventional datasets for latent $z_i$, where in each interventional dataset, $z_i$ is set to a distinct fixed value under $do$ intervention following \Cref{eqn: do_int}. If the number of $do$ interventional datasets is sufficiently large and the support of the latents satisfy certain regularity conditions (detailed in \Cref{thm_do_appendix}), then the autoencoder that solves \Cref{eqn: recons} under multiple constraints of the form \Cref{eqn: intv_cons} identifies $z_i$ up to an invertible transform approximately. 
\end{theorem-non}

\subsection{General perfect and imperfect interventions}
In the discussion so far, we focused on $do$ interventions.  In this section, our goal is to build identification guarantees under imperfect interventions. In the example that follows, we motivate the class of imperfect interventions we consider. 

\paragraph{Motivating example of perfect \& imperfect interventions on images.} First, we revisit perfect interventions in causal DAGs \citep{peters2017elements}. Under a perfect  intervention, the intervened latent is disconnected from its parents and $do$ interventions are a special case of perfect interventions.   Consider the two balls shown in \Cref{fig2}a. Suppose Ball 1 has a strong influence on Ball 2 in the observational DAG shown in \Cref{fig2}b. As a result, the position of Ball 1 determines the region where Ball 2 can be located inside the box in \Cref{fig2}a. Now imagine if a perfect intervention is carried out as shown in \Cref{fig2}c. Under this intervention the second coordinate of Ball 2 is not restricted by Ball 1 and it takes all possible values in the box. Do we need  perfect interventions to ensure that  Ball 2 can be located anywhere in the box? Even an imperfect intervention that reduces the strength of influence of Ball 1 on Ball 2 can suffice to ensure that Ball 2 takes all possible locations in the box. In this section, we consider such imperfect interventions that guarantee that the range of values the intervened latent takes  does not depend on its non-descendants. We formalize this below.

\begin{definition} \label{def: sup_ind} \citep{wang2021desiderata}
Consider a random variable $V = [V_1, V_2]$ sampled from $\mathbb{P}_V$. $V_1,V_2$ are said to have independent support if 
$\mathcal{V} = \mathcal{V}_1 \times \mathcal{V}_2$
where $\mathcal{V}$ is the support of $\mathbb{P}_V$, $\mathcal{V}_j$  are the supports of marginal distribution of $V_j$ for $j\in \{1,2\}$ and $\times$ is the Cartesian product. 
\end{definition}

Observe that two random variables can be dependent but have independent support. Suppose $z$ is generated from a structural causal model with an underlying DAG and $z_i$ undergoes an imperfect intervention.   We consider imperfect interventions such that each pair $(z_i,z_j)$ satisfies support independence (\Cref{def: sup_ind}), where $z_j$ is a non-descendant of $z_i$ in the underlying DAG. Below we characterize imperfect interventions that satisfy support independence.

\paragraph{Characterizing imperfect interventions that lead to support independence.}  Suppose $z_i \leftarrow w(\mathsf{Pa}(z_{i}), u)$, where $\mathsf{Pa}(z_{i})$ is the value of the set of parents of $z_i$, $u \in \mathcal{U}$ is a noise variable that is independent of the ancestors of $z_i$, and $w$ is the map that  generates $z_i$. We carry out an imperfect intervention on $z_i$ and change the map $w$ to $v$.  If the range of values assumed by $v$ for any two values assumed by the parents are equal, then the support of $z_i$ is independent of all its non-descendants. Formally stated the condition is $v(\mathsf{Pa}(z_{i}), \mathcal{U})= v(\mathsf{Pa}^{'}(z_{i}), \mathcal{U})$,  where  $\mathsf{Pa}(z_{i})$ and $\mathsf{Pa}^{'}(z_{i})$ are any two sets of values assumed by the parents. 

We are now ready to describe the geometric properties we require of the interventional distribution $\mathbb{P}_{Z}^{(i)}$ in \Cref{eqn: dgp_int}. We introduce some notation before that.  Let $[d] \coloneqq \{1,\cdots, d\}$. For each $j \in [d]$, we define the supremum and infimum of each component $z_j$ in the interventional distribution. Define  $ \alpha_{\sup}^{j}$ ($ \alpha_{\inf}^{j}$) to be the supremum (infimum) of the set $\mathcal{Z}_{j}^{(i)}$.

\begin{assumption}
\label{assm: soft_intv}
Consider $z$ sampled from the interventional distribution $\mathbb{P}_{Z}^{(i)}$ in \Cref{eqn: dgp_int}.    $\exists \mathcal{S}\subseteq [d]$ such that the support of $z_i$ is independent of $z_j$ for all $j\in \mathcal{S}$. For all $j \in \mathcal{S}$
\begin{equation}
\mathcal{Z}^{(i)}_{i,j} = \mathcal{Z}^{(i)}_{i} \times \mathcal{Z}^{(i)}_{j}
\end{equation}
For all $j \in [d]$, $-\infty <\alpha_{\inf}^{j} \leq \alpha_{\sup}^{j} < \infty$. $\exists\; \zeta>0$ such that $\Pi_{j\in [d]}(\alpha_{\sup}^{j} -\zeta, \alpha_{\sup}^{j}) \cup (\alpha_{\inf}^{j}, \alpha_{\inf}^{j}+\zeta) \subseteq \mathcal{Z}^{(i)}$ and $\Pi$ denotes the Cartesian product. 
\end{assumption}

The  distribution $\mathbb{P}_{Z}^{(i)}$ above is quite general in several ways as it encompasses  i) all perfect interventions since  they render the intervened latent independent of its non-descendants and ii) imperfect interventions that lead to independent support as characterized above. The latter part of the above assumption is a regularity condition on the geometry of the support. It ensures the support of $z$ has a $\zeta$-thick boundary for a $\zeta>0$.

We now describe a constraint on the encoder that leverages the geometric signature of imperfect interventions in \Cref{assm: soft_intv}.  Recall $\hat{z}_k = f_k(x)$. Let $\hat{\mathcal{Z}} = f(\mathcal{X})$ and $\hat{\mathcal{Z}}^{(i)} = f(\mathcal{X}^{(i)})$ represent the support of encoder $f$'s output on observational data and interventional data respectively.  $ \hat{\mathcal{Z}}^{(i)}_{k,m} $ represents the joint  support of  $(\hat{z}_k,\hat{z}_m)$  and $ \hat{\mathcal{Z}}_{k}^{(i)} $ is the support of $\hat{z}_k$ in interventional data. Similarly, we define $ \hat{\mathcal{Z}}_{k,m} $  and $ \hat{\mathcal{Z}}_{k} $  for observational data.

\begin{constraint}
     \label{eqn: intv_cons_soft}
Given a set $\mathcal{S}^{'}$. For each $m \in \mathcal{S}^{'}$, $(\hat{z}_k,\hat{z}_m)$  satisfies support independence on interventional data, i.e.,  
 $$ \hat{\mathcal{Z}}^{(i)}_{k,m} = \hat{\mathcal{Z}}^{(i)}_{k} \times \hat{\mathcal{Z}}^{(i)}_{m}, \forall m \in \mathcal{S}^{'}.$$
\end{constraint}

In the above \Cref{eqn: intv_cons_soft}, the index $k$ and set $\mathcal{S}^{'}$ are not necessarily the same as $i$ and $\mathcal{S}$ from \Cref{assm: soft_intv}. In the theorem that follows, we require $|\mathcal{S}^{'}| \leq |\mathcal{S}|$ to guarantee that a solution to \Cref{eqn: intv_cons_soft} exists. In the Appendix \ref{sec: rep_id_imp}, we explain that this requirement can be easily relaxed.  Note that \Cref{eqn: intv_cons_soft} bears similarity to \Cref{eqn: intv_cons} from the case of $do$ interventions. Both constraints ensure that the support of the $k^{th}$ component is independent of all other components. 
In the theorem that follows, we show that the above \Cref{eqn: intv_cons_soft} helps achieve block affine identification, which we formally define below.

\begin{definition}
    If $\hat{z} = \tilde{\Lambda}\Pi  z + c$ for all $z\in \mathcal{Z} \cup \mathcal{Z}^{(i)}$, where $\Pi$ is a permutation matrix, $\tilde{\Lambda}$ is an invertible matrix such that there is a submatrix of $\tilde{\Lambda}$ which is zero, then $\hat{z}$ is said to block-affine identify $z$.
\end{definition}

\begin{restatable}{theorem}{GeneralItvID}
\label{thm3}  
Suppose the observational data and interventional data are generated from  \Cref{eqn: dgp} and \Cref{eqn: dgp_int} respectively under Assumptions \ref{assm1: dgp}, \ref{assm2: g_poly}, \ref{assm: soft_intv}. The autoencoder that solves \Cref{eqn: recons} under Constraint \ref{assm3: h_poly_new}, \ref{eqn: intv_cons_soft} (with $|\mathcal{S}^{'}| \leq |\mathcal{S}|$)  achieves block affine identification. More specifically,  $\forall z \in \mathcal{Z} \cup \mathcal{Z}^{(i)}$ $$\hat{z}_k = a_{k}^{\top} z + c_k, \hat{z}_{m} = a_m^{\top} z  + c_{m}, \forall m \in \mathcal{S}^{'},$$
where $a_k$ contains at most $d-|\mathcal{S}^{'}|$ non-zero elements and  each component  of $a_m$ is zero whenever the corresponding component of $a_k$ is non-zero for all $m\in \mathcal{S}^{'}$.
\end{restatable}

Firstly,  from \Cref{thm1}, $\hat{z}=Az+c$. From the above theorem, it follows that  $\hat{z}_k$ linearly depends on at most $d-|\mathcal{S}^{'}|$ latents and not all the latents. Each $\hat{z}_m$ with $m\in \mathcal{S}^{'}$ does not depend on any of the latents that $\hat{z}_k$ depends on.  As a result, $|\mathcal{S}^{'}|+1$ rows of $A$ (from \Cref{thm1}) are sparse.  Observe that if $|\mathcal{S}^{'}|=|\mathcal{S}| = d-1$, then as a result of the above theorem, $\hat{z}_k$ identifies some $z_j$ up to scale and shift. Further, remaining components $\hat{z}_{-k}$ linearly depend on $z_{-j}$ and do not depend on $z_j$. The proof of \Cref{thm3} is in Appendix \ref{sec: rep_id_imp}.

\section{Extensions to Identification with
Observational Data \& Independent Support}
\label{sec: obs}

In the previous section, we showed that interventions induce geometric structure (independence of supports) in the support of the latents that helps  achieve strong identification guarantees. In this section, we consider a special case where such geometric structure is already present in the support of the latents in the observational data. Since we only work with observational data in this section, we set the interventional supports $\mathcal{Z}^{(i)} = \mathcal{X}^{(i)} = \emptyset$, where $\emptyset$ is the empty set. For each $j \in [d]$, define  $ \beta_{\sup}^{j}$ to be the supremum of the support of $z_j$, i.e., $\mathcal{Z}_{j}$.  Similarly, for each $j \in [d]$, define  $ \beta_{\inf}^{j}$ to be the infimum of the set $\mathcal{Z}_{j}$.

\begin{assumption}
\label{assm: obs_fact}
The support of $\mathbb{P}_Z$ in \Cref{eqn: dgp} satisfies pairwise support independence between all the pairs of latents. Formally stated, 
\begin{equation}
\mathcal{Z}_{r,s} = \mathcal{Z}_{r} \times \mathcal{Z}_{s}, \forall r \not =s,  r, s \in [d]
\end{equation}
For all $r \in [d]$, $-\infty <\beta_{\inf}^{r} \leq \beta_{\sup}^{r} < \infty$. $\exists\;\zeta>0$ such that  $\Pi_{r\in [d]}(\beta_{\sup}^{r} -\zeta, \beta_{\sup}^{r}) \cup (\beta_{\inf}^{r}, \beta_{\inf}^{r}+\zeta) \subseteq \mathcal{Z}$ and $\Pi$ denotes the Cartesian product.
\end{assumption}

Following previous sections, we state a constraint, where the learner leverages the geometric structure in the support in \Cref{assm: obs_fact} to search for the autoencoder.

\begin{constraint}
    \label{eqn: obs_supp_ind}
Each pair $(\hat{z}_k, \hat{z}_m)$, where $k,m \in [d]$ and $k\not=m$ satisfies support independence on observational data, i.e., $  \hat{\mathcal{Z}}_{k,m} = \hat{\mathcal{Z}}_{k} \times \hat{\mathcal{Z}}_{m}$, where $\hat{\mathcal{Z}}_{k,m}$ is the joint support of $(\hat{z}_k, \hat{z}_m)$ and $\hat{\mathcal{Z}}_{k}$ is support of $\hat{z}_k$.
\end{constraint}

\begin{restatable}{theorem}{ObsSuppID}
\label{thm4}  
Suppose the observational data is generated from  \Cref{eqn: dgp} under Assumption \ref{assm1: dgp}, \ref{assm2: g_poly}, and \ref{assm: obs_fact}, The autoencoder that the solves \Cref{eqn: recons}   under \Cref{eqn: obs_supp_ind} achieves permutation, shift and scaling identification. Specifically,  $\forall z \in \mathcal{Z}, \hat{z} = \Lambda \Pi z + c,$  where $\hat{z}$ is the output of the encoder $f$ and $z$ is the true latent and $\Pi$ is a permutation matrix and $\Lambda$ is an invertible diagonal matrix.
\end{restatable}

The proof of \Cref{thm4} is in Appendix \ref{sec: rep_id_obs}. \Cref{thm4} says that the independence between the latents' support is
sufficient to achieve identification up to permutation, shift, and
scaling in observational data. 
\Cref{thm4} has important implications for the seminal works on linear
ICA~\citep{comon1994independent}, considering the simple case of a
linear $g$. \citet{comon1994independent} shows that, if the latent
variables are independent and non-Gaussian, then the latent variables
can be identified up to permutation and scaling. However, \Cref{thm4}
states that, even if the latent variables are dependent, the latent
variables can be identified up to permutation, shift and scaling, as
long as they are bounded (hence non-Gaussian) and satisfy pairwise
support independence.

Finally, \Cref{thm4} provides a first general theoretical
justification for recent proposals of unsupervised disentanglement via
the independent support
condition~\citep{wang2021desiderata,roth2022disentanglement}.

\section{Learning Representations from Geometric Signatures: Practical Considerations}
\label{sec: method}
In this section, we describe practical algorithms to solve the constrained representation learning problems in \Cref{sec:intervention,sec: obs}.

To perform constrained representation learning with $do$-intervention
data, we proceed in two steps. In the first step, we carry out minimization of the reconstruction objective $f^{\dagger},h^{\dagger} =  \argmin_{f,h} \mathbb{E} \big[\|h \circ f (X) -X\|^2\big] $, where $h$ is the decoder, $f$ is the encoder and expectation is taken over observational data and interventional data.  In the experiments, we restrict $h$ to be a polynomial and show that affine identification is achieved by the learned $f^{\dagger}$ as proved in \Cref{thm1}. 

In the second step, we learn a linear map to transform the learned
representations and enforce \Cref{eqn:
intv_cons}. For each interventional distribution,
$\mathbb{P}_X^{(i)}$, we learn a different linear map $\gamma_{i}$
that projects the representation such that it takes an arbitrary fixed
value $z^{\dagger}_{i}$ on the support of $\mathbb{P}_X^{(i)}$. We
write this objective as
\begin{equation}
\min_{\{\gamma_{i}\}}\sum_{i} \mathbb{E}_{X \sim \mathbb{P}_X^{(i)}}\bigg[\big\|\gamma_{i}^{\top} f^{\dagger} (X) -z^{\dagger}_{i}\big\|^2\bigg] .
\label{eqn: cons_int}
\end{equation}
Construct a matrix $\Gamma$ with different $\gamma_{i}^{\top}$ as the rows. The final output representation is $\Gamma f^{\dagger}(X)$. In the experiments, we show that this representation achieves permutation, shift and scaling identification as predicted by Theorem \ref{thm2}. A few remarks in order. i) $z_i^{\dagger}$ is arbitrary and learner does not know the true do intervention value, ii) for ease of exposition, \Cref{eqn: cons_int} assumes the knowledge of index of intervened  and can be easily relaxed by multiplying $\Gamma$ with a permutation matrix.

We next describe an algorithm that learns representations to enforce
independence of support (leveraged in Theorem \ref{thm3} and
\ref{thm4}).  To measure the (non)-independence of the latents'
support, we follow \citet{wang2021desiderata, roth2022disentanglement}
and measure the distance between the sets in terms of Hausdorff
distance: the Hausdorff distance $\mathsf{HD}$ between the sets
$\mathcal{S}_1,
\mathcal{S}_2$ is
$\mathsf{HD}(\mathcal{S}_1, \mathcal{S}_2) =  \sup_{z\in
\mathcal{S}_2}\bigg(\inf_{z^{'}\in \mathcal{S}_1}(\|z-z^{'}\|)\bigg)$,
where $\mathcal{S}_1 \subseteq \mathcal{S}_2$.

To further enforce the independent support constraint, we again follow a
two-step algorithm. The first step remains the same, i.e., we minimize
the reconstruction objective.  In the second step, we transform the
learned representations ($f^{\dagger}(x)$) with an invertible map $\Gamma \in
\mathbb{R}^{d\times d}$. The joint support obtained post
transformation is a function of the parameters $\Gamma$ and is denoted
as $\hat{\mathcal{Z}}(\Gamma)$. Following the notation introduced
earlier, the joint support along dimensions $k,m$ is
$\hat{\mathcal{Z}}_{k,m}(\Gamma)$ and the marginal support along $k$
is $\hat{\mathcal{Z}}_{k}(\Gamma)$. We translate the problem in
\Cref{eqn: obs_supp_ind} as follows. We find a $\Gamma$ to minimize
\begin{equation}\min_{\Gamma} \sum_{k\not=m } \mathsf{HD}\big(\hat{\mathcal{Z}}_{k,m}(\Gamma), \hat{\mathcal{Z}}_{k}(\Gamma) \times \hat{\mathcal{Z}}_{m}(\Gamma)\big). \label{eqn: hd_2}
\end{equation}
\Cref{eqn: intv_cons_soft} can be similarly translated.

\section{Empirical Findings} 
\label{sec: emp}

In this section, we analyze how the practical implementation of the theory holds up in different settings ranging from data generated from polynomial decoders to images generated from PyGame rendering engine \citep{shinners2011pygame}. The code to reproduce the experiments can be found at \url{https://github.com/facebookresearch/CausalRepID}.


\paragraph{Data generation process.} \textit{Polynomial decoder data:} The latents for the observational data are sampled from $\mathbb{P}_{Z}$. $\mathbb{P}_Z$  can be i) independent uniform, ii) an SCM with sparse connectivity (SCM-S), iii) an SCM with dense connectivity (SCM-D)~\citep{brouillard2020differentiable}.  The latent variables are then mapped to $x$ using a multivariate polynomial. We use a $n=200$ dimensional $x$. We use two possible dimensions for the latents ($d$) -- six and ten. We use polynomials of degree ($p$) two and three. Each element in $G$ to generate $x$ is sampled from a standard normal distribution.   

\textit{Image data:} For image-based experiments, we used the PyGame~\citep{pygame} rendering engine. We generate $64 \times 64 \times 3$ pixel images of the form in  \Cref{fig2} and consider a setting with two balls. We consider three distributions for latents: i) independent uniform, ii) a linear SCM with DAG in \Cref{fig2}, iii) a non-linear SCM with DAG in \Cref{fig2}, where the coordinates of Ball $1$ are at the top layer in the DAG and coordinates of Ball $2$ are at the bottom layer in the DAG.   

For both settings above,  we carry out $do$ interventions on each latent dimension to generate interventional data.


\paragraph{Model parameters and evaluation metrics.} We follow the two step training procedures described in \Cref{sec: method}.
For image-based experiments we use a ResNet-18 as the encoder \citep{he2016deep} and for all other experiments, we use an MLP with three hidden layers and two hundred units per layer. We learn a polynomial decoder $h$ as the theory prescribes to use a polynomial decoder (\Cref{assm3: h_poly_new}) when $g$ is a polynomial. In \Cref{sec:res_poly_decoder_app}, we also present results when we use an MLP decoder.    To check for affine identification (from \Cref{thm1}), we measure the $R^2$ score for linear regression between the output representation and the true representation. If the score is high, then it guarantees affine identification. To verify permutation, shift and scaling identification (from \Cref{thm4}), we check the mean correlation coefficient (MCC \citep{khemakhem2020variational}). For further details on data generation, models, hyperparamters, and supplementary experiments refer to the \Cref{sec:emp_find_append}.

\begin{table}[!h]
\small
    \centering
\begin{tabular}{ccccc}
\toprule
      $\mathbb{P}_Z$ &   $d$ &  $p$ &                     $R^2$ &                     MCC (IOS)\\
\midrule
         Uniform &           $6$ &            $2$ &   $1.00 \pm 0.00$ & $99.3 \pm 0.07$ \\
         Uniform &           $6$ &            $3$ &   $1.00 \pm 0.00$ & $99.4 \pm 0.06$ \\
         Uniform &          $10$ &            $2$ &   $1.00 \pm 0.00$ & $90.7 \pm 2.92$ \\
         Uniform &          $10$ &            $3$ &  $0.99 \pm 0.00$ &  $94.6 \pm 1.50$ \\
         \midrule
           SCM-S &           $6$ &            $2$ & $0.96 \pm 0.02$ & $72.6 \pm 1.48$ \\
           SCM-S &           $6$ &            $3$ & $0.87 \pm 0.07$ & $70.6 \pm 1.54$ \\
           SCM-S &          $10$ &            $2$ &  $0.99 \pm 0.00$ & $65.9 \pm 1.32$ \\
           SCM-S &          $10$ &            $3$ &  $0.90 \pm 0.05$ & $58.8 \pm 1.27$ \\
         \midrule
           SCM-D &           $6$ &            $2$ & $0.97 \pm 0.01$ & $61.6 \pm 4.36$ \\
           SCM-D &           $6$ &            $3$ & $0.81 \pm 0.11$ &  $65.2 \pm 2.70$ \\
           SCM-D &          $10$ &            $2$ &  $0.83 \pm 0.10$ & $69.6 \pm 3.09$ \\
           SCM-D &          $10$ &            $3$ & $0.72 \pm 0.15$ &  $60.1 \pm 1.16$ \\
\bottomrule
\end{tabular}

\caption{Observational data with polynomial decoder $g$: Mean ± S.E. (5 random seeds). $R^2$ and MCC(IOS) (for uniform) have high values as predicted in \Cref{thm1} and \Cref{thm4} respectively. }
    \label{table1_results}
\end{table}

\begin{table}[!h]
\small
    \centering
\begin{tabular}{cccccc}
\toprule
     $\mathbb{P}_Z$ &  $d$  &  $p$ &                      MCC  &            MCC (IL)\\
\midrule
         Uniform &           $6$ &            $2$ & $69.1 \pm 1.11$ & $100.0 \pm 0.00$ \\
         Uniform &           $6$ &            $3$ & $73.4 \pm 0.49$ &  $100.0 \pm 0.00$ \\
         Uniform &          $10$ &            $2$ & $59.9 \pm 2.03$ &  $100.0 \pm 0.00$ \\
         Uniform &          $10$ &            $3$ &  $65.9 \pm 0.80$ & $\;99.9 \pm 0.03 $\\
         \midrule
           SCM-S &           $6$ &            $2$ &  $68.4 \pm 0.90$ & $99.5 \pm 0.38$ \\
           SCM-S &           $6$ &            $3$ & $74.1 \pm 2.32$ & $99.3 \pm 0.34$ \\
           SCM-S &          $10$ &            $2$ & $68.0 \pm 2.36$ & $99.9 \pm 0.03$ \\
           SCM-S &          $10$ &            $3$ &  $66.8 \pm 1.10$ &  $98.8 \pm 0.13$ \\
         \midrule
           SCM-D &           $6$ &            $2$ &  $71.8 \pm 3.77$ & $99.6 \pm 0.12$ \\
           SCM-D &           $6$ &            $3$ & $79.5 \pm 3.45$ & $98.2 \pm 1.07$ \\
           SCM-D &          $10$&            $2$ & $70.8 \pm 1.89$ &  $95.3 \pm 2.24$ \\
           SCM-D &          $10$ &            $3$ &  $70.1 \pm 2.80$ & $97.2 \pm 0.88$ \\
\bottomrule
\end{tabular}
    \caption{Interventional data with polynomial decoder $g$:  Mean ± S.E.  (5 random seeds). MCC(IL) is high as shown in \Cref{thm2}.}
    \label{table2_results}
\end{table}

\paragraph{Results for polynomial decoder.} 
\textit{Observational data:} We consider the setting when the true decoder $g$ is a polynomial and the learned decoder $h$ is also a polynomial. In Table \ref{table1_results}, we report the $R^2$ between the representation learned after the first step, where we only minimize reconstruction loss. $R^2$ values are high as predicted in \Cref{thm1}. In the second step, we learn a map $\Gamma$ and enforce independence of support constraint by minimizing Hausdorff distance from \Cref{eqn: hd_2}. Among the distributions $\mathbb{P}_Z$ only the uniform distribution satisfies support independence from \Cref{assm: obs_fact} and following \Cref{thm4}, we expect MCC to be high in this case only. In Table \ref{table1_results}, we report the MCC obtained by enforcing independence of support in  MCC (IOS). In the \Cref{sec:res_poly_decoder_app}, we also carry out experiments on correlated uniform distributions and observe high MCC (IOS).

\textit{Interventional data:} We now consider the case when we also have access to $do$ intervention data in addition to observational data. We consider the setting with one $do$ intervention per latent dimension. We follow the two step procedure described in \Cref{sec: method}. In Table \ref{table2_results}, we first show the MCC values of the representation obtained after the first step in the MCC column. In the second step, we learn $\Gamma$ by minimizing the interventional loss (IL) in \Cref{eqn: cons_int}. We report the MCC of the representation obtained in the MCC (IL) column in Table \ref{table2_results}; the values are close to one as predicted by Theorem \ref{thm2}. 


\paragraph{Results for image dataset.} We follow the two step procedure described in \Cref{sec: method} except now in the second step, we learn a non-linear map (using an MLP) to minimize the interventional loss (IL) in  \Cref{eqn: cons_int}. In Table \ref{table4_results}, we show the MCC values achieved by the learned representation as we vary the number of $do$ interventional distributions per latent dimension. As shown in  \Cref{thm_do_appendix}, more interventional distributions per latent dimension improve the MCC. 
\begin{table}[t]
\small
    \centering
\begin{tabular}{llll}
\toprule
    \#interv dist.   &  Uniform  & SCM linear  &   SCM non-linear                 \\
\midrule
          $1$ &        $34.2 \pm 0.24$    &    $12.8 \pm 0.28$    &  $19.7 \pm 0.31$  \\
          $3$ &        $73.9 \pm 0.38$    &   $73.2 \pm 0.33$      &  $59.7 \pm 0.28$  \\
         $5$&         $73.6 \pm 0.21$  &     $83.4 \pm 0.21$      &  $62.8 \pm 0.2$ \\
         $7$ &         $72.5 \pm 0.34$  &    $84.2 \pm 0.25$       &  $69.3 \pm 0.34$ \\
         $9$ &         $73.1 \pm 0.47$ &    $86.2 \pm 0.17$      &  $71.4 \pm 0.26$ \\
\bottomrule
\end{tabular}
    \caption{Interventional data in image-based experiments: Mean ± S.E (5 random seeds). MCCs increase with the number of $do$ interventional distributions per latent dimension (\Cref{thm_do_appendix}).}
    \label{table4_results}
\end{table}
\section{Conclusions}

In this work, we lay down the theoretical foundations for learning causal representations in the presence of interventional data. We show that geometric signatures such as support independence that are induced under many interventions are useful for provable representation identification.  Looking forward, we believe that exploring representation learning with real interventional data \citep{lopez2022learning, liu2023causal}  is a fruitful avenue for future work. 

\section*{Acknowledgments}
Yixin Wang acknowledges grant support from the National Science Foundation and the Office of Naval Research. 
Yoshua Bengio acknowledges the support from CIFAR and IBM. We thank Anirban Das for insightful feedback that helped us correctly state the precise $\zeta$-thickness conditions.

\clearpage

\bibliography{ICRL_camera_ready}

\begin{thebibliography}{50}
\providecommand{\natexlab}[1]{#1}
\providecommand{\url}[1]{\texttt{#1}}
\expandafter\ifx\csname urlstyle\endcsname\relax
  \providecommand{\doi}[1]{doi: #1}\else
  \providecommand{\doi}{doi: \begingroup \urlstyle{rm}\Url}\fi

\bibitem[Ahuja et~al.(2021)Ahuja, Hartford, and Bengio]{ahuja2021properties}
Ahuja, K., Hartford, J., and Bengio, Y.
\newblock Properties from mechanisms: an equivariance perspective on
  identifiable representation learning.
\newblock \emph{arXiv preprint arXiv:2110.15796}, 2021.

\bibitem[Ahuja et~al.(2022{\natexlab{a}})Ahuja, Hartford, and
  Bengio]{ahuja2022weakly}
Ahuja, K., Hartford, J., and Bengio, Y.
\newblock Weakly supervised representation learning with sparse perturbations.
\newblock \emph{arXiv preprint arXiv:2206.01101}, 2022{\natexlab{a}}.

\bibitem[Ahuja et~al.(2022{\natexlab{b}})Ahuja, Mahajan, Syrgkanis, and
  Mitliagkas]{ahuja2022towards}
Ahuja, K., Mahajan, D., Syrgkanis, V., and Mitliagkas, I.
\newblock Towards efficient representation identification in supervised
  learning.
\newblock \emph{arXiv preprint arXiv:2204.04606}, 2022{\natexlab{b}}.

\bibitem[Ash et~al.(2000)Ash, Robert, Doleans-Dade, and
  Catherine]{ash2000probability}
Ash, R.~B., Robert, B., Doleans-Dade, C.~A., and Catherine, A.
\newblock \emph{Probability and measure theory}.
\newblock Academic press, 2000.

\bibitem[Bareinboim et~al.(2022)Bareinboim, Correa, Ibeling, and
  Icard]{bareinboim2022pearl}
Bareinboim, E., Correa, J.~D., Ibeling, D., and Icard, T.
\newblock On pearl’s hierarchy and the foundations of causal inference.
\newblock In \emph{Probabilistic and Causal Inference: The Works of Judea
  Pearl}, pp.\  507--556. 2022.

\bibitem[Bengio et~al.(2013)Bengio, Courville, and
  Vincent]{bengio2013representation}
Bengio, Y., Courville, A., and Vincent, P.
\newblock Representation learning: A review and new perspectives.
\newblock \emph{IEEE transactions on pattern analysis and machine
  intelligence}, 35\penalty0 (8):\penalty0 1798--1828, 2013.

\bibitem[Bommasani et~al.(2021)Bommasani, Hudson, Adeli, Altman, Arora, von
  Arx, Bernstein, Bohg, Bosselut, Brunskill,
  et~al.]{bommasani2021opportunities}
Bommasani, R., Hudson, D.~A., Adeli, E., Altman, R., Arora, S., von Arx, S.,
  Bernstein, M.~S., Bohg, J., Bosselut, A., Brunskill, E., et~al.
\newblock On the opportunities and risks of foundation models.
\newblock \emph{arXiv preprint arXiv:2108.07258}, 2021.

\bibitem[Brehmer et~al.(2022)Brehmer, De~Haan, Lippe, and
  Cohen]{brehmer2022weakly}
Brehmer, J., De~Haan, P., Lippe, P., and Cohen, T.
\newblock Weakly supervised causal representation learning.
\newblock \emph{arXiv preprint arXiv:2203.16437}, 2022.

\bibitem[Brouillard et~al.(2020)Brouillard, Lachapelle, Lacoste,
  Lacoste-Julien, and Drouin]{brouillard2020differentiable}
Brouillard, P., Lachapelle, S., Lacoste, A., Lacoste-Julien, S., and Drouin, A.
\newblock Differentiable causal discovery from interventional data.
\newblock \emph{Advances in Neural Information Processing Systems},
  33:\penalty0 21865--21877, 2020.

\bibitem[Brown et~al.(2020)Brown, Mann, Ryder, Subbiah, Kaplan, Dhariwal,
  Neelakantan, Shyam, Sastry, Askell, et~al.]{brown2020language}
Brown, T., Mann, B., Ryder, N., Subbiah, M., Kaplan, J.~D., Dhariwal, P.,
  Neelakantan, A., Shyam, P., Sastry, G., Askell, A., et~al.
\newblock Language models are few-shot learners.
\newblock \emph{Advances in neural information processing systems},
  33:\penalty0 1877--1901, 2020.

\bibitem[Burgess et~al.(2018)Burgess, Higgins, Pal, Matthey, Watters,
  Desjardins, and Lerchner]{burgess2018understanding}
Burgess, C.~P., Higgins, I., Pal, A., Matthey, L., Watters, N., Desjardins, G.,
  and Lerchner, A.
\newblock Understanding disentangling in beta-vae.
\newblock \emph{arXiv preprint arXiv:1804.03599}, 2018.

\bibitem[Comon(1994)]{comon1994independent}
Comon, P.
\newblock Independent component analysis, a new concept?
\newblock \emph{Signal processing}, 36\penalty0 (3):\penalty0 287--314, 1994.

\bibitem[Dixit et~al.(2016)Dixit, Parnas, Li, Chen, Fulco, Jerby-Arnon,
  Marjanovic, Dionne, Burks, Raychowdhury, et~al.]{dixit2016perturb}
Dixit, A., Parnas, O., Li, B., Chen, J., Fulco, C.~P., Jerby-Arnon, L.,
  Marjanovic, N.~D., Dionne, D., Burks, T., Raychowdhury, R., et~al.
\newblock Perturb-seq: dissecting molecular circuits with scalable single-cell
  rna profiling of pooled genetic screens.
\newblock \emph{cell}, 167\penalty0 (7):\penalty0 1853--1866, 2016.

\bibitem[Geirhos et~al.(2020)Geirhos, Jacobsen, Michaelis, Zemel, Brendel,
  Bethge, and Wichmann]{geirhos2020shortcut}
Geirhos, R., Jacobsen, J.-H., Michaelis, C., Zemel, R., Brendel, W., Bethge,
  M., and Wichmann, F.~A.
\newblock Shortcut learning in deep neural networks.
\newblock \emph{Nature Machine Intelligence}, 2\penalty0 (11):\penalty0
  665--673, 2020.

\bibitem[Goyal \& Bengio(2020)Goyal and Bengio]{goyal2020inductive}
Goyal, A. and Bengio, Y.
\newblock Inductive biases for deep learning of higher-level cognition.
\newblock \emph{arXiv preprint arXiv:2011.15091}, 2020.

\bibitem[H{\"a}lv{\"a} \& Hyvarinen(2020)H{\"a}lv{\"a} and
  Hyvarinen]{halva2020hidden}
H{\"a}lv{\"a}, H. and Hyvarinen, A.
\newblock Hidden markov nonlinear ica: Unsupervised learning from nonstationary
  time series.
\newblock In \emph{Conference on Uncertainty in Artificial Intelligence}, pp.\
  939--948. PMLR, 2020.

\bibitem[He et~al.(2016)He, Zhang, Ren, and Sun]{he2016deep}
He, K., Zhang, X., Ren, S., and Sun, J.
\newblock Deep residual learning for image recognition.
\newblock In \emph{Proceedings of the IEEE conference on computer vision and
  pattern recognition}, pp.\  770--778, 2016.

\bibitem[Hyvarinen \& Morioka(2016)Hyvarinen and
  Morioka]{hyvarinen2016unsupervised}
Hyvarinen, A. and Morioka, H.
\newblock Unsupervised feature extraction by time-contrastive learning and
  nonlinear {ICA}.
\newblock \emph{Advances in neural information processing systems}, 29, 2016.

\bibitem[Hyvarinen \& Morioka(2017)Hyvarinen and
  Morioka]{hyvarinen2017nonlinear}
Hyvarinen, A. and Morioka, H.
\newblock Nonlinear {ICA} of temporally dependent stationary sources.
\newblock In \emph{Artificial Intelligence and Statistics}, pp.\  460--469.
  PMLR, 2017.

\bibitem[Hyv{\"a}rinen \& Pajunen(1999)Hyv{\"a}rinen and
  Pajunen]{hyvarinen1999nonlinear}
Hyv{\"a}rinen, A. and Pajunen, P.
\newblock Nonlinear independent component analysis: Existence and uniqueness
  results.
\newblock \emph{Neural networks}, 12\penalty0 (3):\penalty0 429--439, 1999.

\bibitem[Hyvarinen et~al.(2019)Hyvarinen, Sasaki, and
  Turner]{hyvarinen2019nonlinear}
Hyvarinen, A., Sasaki, H., and Turner, R.
\newblock Nonlinear ica using auxiliary variables and generalized contrastive
  learning.
\newblock In \emph{The 22nd International Conference on Artificial Intelligence
  and Statistics}, pp.\  859--868. PMLR, 2019.

\bibitem[Khemakhem et~al.(2020)Khemakhem, Monti, Kingma, and
  Hyvarinen]{khemakhem2020ice}
Khemakhem, I., Monti, R., Kingma, D., and Hyvarinen, A.
\newblock Ice-beem: Identifiable conditional energy-based deep models based on
  nonlinear {ICA}.
\newblock \emph{Advances in Neural Information Processing Systems},
  33:\penalty0 12768--12778, 2020.

\bibitem[Khemakhem et~al.(2022)Khemakhem, Kingma, Monti, and
  Hyvarinen]{khemakhem2020variational}
Khemakhem, I., Kingma, D., Monti, R., and Hyvarinen, A.
\newblock Variational autoencoders and nonlinear {ICA}: A unifying framework.
\newblock In \emph{International Conference on Artificial Intelligence and
  Statistics}, pp.\  2207--2217. PMLR, 2022.

\bibitem[Klindt et~al.(2020)Klindt, Schott, Sharma, Ustyuzhaninov, Brendel,
  Bethge, and Paiton]{klindt2020towards}
Klindt, D., Schott, L., Sharma, Y., Ustyuzhaninov, I., Brendel, W., Bethge, M.,
  and Paiton, D.
\newblock Towards nonlinear disentanglement in natural data with temporal
  sparse coding.
\newblock \emph{arXiv preprint arXiv:2007.10930}, 2020.

\bibitem[Lachapelle et~al.(2022)Lachapelle, Rodriguez, Sharma, Everett,
  Le~Priol, Lacoste, and Lacoste-Julien]{lachapelle2022disentanglement}
Lachapelle, S., Rodriguez, P., Sharma, Y., Everett, K.~E., Le~Priol, R.,
  Lacoste, A., and Lacoste-Julien, S.
\newblock Disentanglement via mechanism sparsity regularization: A new
  principle for nonlinear {ICA}.
\newblock In \emph{Conference on Causal Learning and Reasoning}, pp.\
  428--484. PMLR, 2022.

\bibitem[Lippe et~al.(2022{\natexlab{a}})Lippe, Magliacane, L{\"o}we, Asano,
  Cohen, and Gavves]{lippe2022icitris}
Lippe, P., Magliacane, S., L{\"o}we, S., Asano, Y.~M., Cohen, T., and Gavves,
  E.
\newblock icitris: Causal representation learning for instantaneous temporal
  effects.
\newblock \emph{arXiv preprint arXiv:2206.06169}, 2022{\natexlab{a}}.

\bibitem[Lippe et~al.(2022{\natexlab{b}})Lippe, Magliacane, L{\"o}we, Asano,
  Cohen, and Gavves]{lippe2022citris}
Lippe, P., Magliacane, S., L{\"o}we, S., Asano, Y.~M., Cohen, T., and Gavves,
  S.
\newblock Citris: Causal identifiability from temporal intervened sequences.
\newblock In \emph{International Conference on Machine Learning}, pp.\
  13557--13603. PMLR, 2022{\natexlab{b}}.

\bibitem[Liu et~al.(2023)Liu, Alahi, Russell, Horn, Zietlow, Sch{\"o}lkopf, and
  Locatello]{liu2023causal}
Liu, Y., Alahi, A., Russell, C., Horn, M., Zietlow, D., Sch{\"o}lkopf, B., and
  Locatello, F.
\newblock Causal triplet: An open challenge for intervention-centric causal
  representation learning.
\newblock \emph{arXiv preprint arXiv:2301.05169}, 2023.

\bibitem[Locatello et~al.(2019)Locatello, Bauer, Lucic, Raetsch, Gelly,
  Sch{\"o}lkopf, and Bachem]{locatello2019challenging}
Locatello, F., Bauer, S., Lucic, M., Raetsch, G., Gelly, S., Sch{\"o}lkopf, B.,
  and Bachem, O.
\newblock Challenging common assumptions in the unsupervised learning of
  disentangled representations.
\newblock In \emph{international conference on machine learning}, pp.\
  4114--4124. PMLR, 2019.

\bibitem[Locatello et~al.(2020)Locatello, Poole, R{\"a}tsch, Sch{\"o}lkopf,
  Bachem, and Tschannen]{locatello2020weakly}
Locatello, F., Poole, B., R{\"a}tsch, G., Sch{\"o}lkopf, B., Bachem, O., and
  Tschannen, M.
\newblock Weakly-supervised disentanglement without compromises.
\newblock In \emph{International Conference on Machine Learning}, pp.\
  6348--6359. PMLR, 2020.

\bibitem[Lopez et~al.(2022)Lopez, Tagasovska, Ra, Cho, Pritchard, and
  Regev]{lopez2022learning}
Lopez, R., Tagasovska, N., Ra, S., Cho, K., Pritchard, J.~K., and Regev, A.
\newblock Learning causal representations of single cells via sparse mechanism
  shift modeling.
\newblock \emph{arXiv preprint arXiv:2211.03553}, 2022.

\bibitem[Mityagin(2015)]{mityagin2015zero}
Mityagin, B.
\newblock The zero set of a real analytic function.
\newblock \emph{arXiv preprint arXiv:1512.07276}, 2015.

\bibitem[Mooij \& Heskes(2013)Mooij and Heskes]{mooij2013cyclic}
Mooij, J. and Heskes, T.
\newblock Cyclic causal discovery from continuous equilibrium data.
\newblock \emph{arXiv preprint arXiv:1309.6849}, 2013.

\bibitem[Nejatbakhsh et~al.(2021)Nejatbakhsh, Fumarola, Esteki, Toyoizumi,
  Kiani, and Mazzucato]{nejatbakhsh2021predicting}
Nejatbakhsh, A., Fumarola, F., Esteki, S., Toyoizumi, T., Kiani, R., and
  Mazzucato, L.
\newblock Predicting perturbation effects from resting activity using
  functional causal flow.
\newblock \emph{bioRxiv}, pp.\  2020--11, 2021.

\bibitem[Pearl(2009)]{pearl2009causal}
Pearl, J.
\newblock Causal inference in statistics: An overview.
\newblock \emph{Statistics surveys}, 3:\penalty0 96--146, 2009.

\bibitem[Pedregosa et~al.(2011)Pedregosa, Varoquaux, Gramfort, Michel, Thirion,
  Grisel, Blondel, Prettenhofer, Weiss, Dubourg, Vanderplas, Passos,
  Cournapeau, Brucher, Perrot, and Duchesnay]{scikit-learn}
Pedregosa, F., Varoquaux, G., Gramfort, A., Michel, V., Thirion, B., Grisel,
  O., Blondel, M., Prettenhofer, P., Weiss, R., Dubourg, V., Vanderplas, J.,
  Passos, A., Cournapeau, D., Brucher, M., Perrot, M., and Duchesnay, E.
\newblock Scikit-learn: Machine learning in {P}ython.
\newblock \emph{Journal of Machine Learning Research}, 12:\penalty0 2825--2830,
  2011.

\bibitem[Peters et~al.(2017)Peters, Janzing, and
  Sch{\"o}lkopf]{peters2017elements}
Peters, J., Janzing, D., and Sch{\"o}lkopf, B.
\newblock \emph{Elements of causal inference: foundations and learning
  algorithms}.
\newblock The MIT Press, 2017.

\bibitem[Radford et~al.(2021)Radford, Kim, Hallacy, Ramesh, Goh, Agarwal,
  Sastry, Askell, Mishkin, Clark, et~al.]{radford2021learning}
Radford, A., Kim, J.~W., Hallacy, C., Ramesh, A., Goh, G., Agarwal, S., Sastry,
  G., Askell, A., Mishkin, P., Clark, J., et~al.
\newblock Learning transferable visual models from natural language
  supervision.
\newblock In \emph{International Conference on Machine Learning}, pp.\
  8748--8763. PMLR, 2021.

\bibitem[Roth et~al.(2022)Roth, Ibrahim, Akata, Vincent, and
  Bouchacourt]{roth2022disentanglement}
Roth, K., Ibrahim, M., Akata, Z., Vincent, P., and Bouchacourt, D.
\newblock Disentanglement of correlated factors via hausdorff factorized
  support.
\newblock \emph{arXiv preprint arXiv:2210.07347}, 2022.

\bibitem[Sch{\"o}lkopf et~al.(2021)Sch{\"o}lkopf, Locatello, Bauer, Ke,
  Kalchbrenner, Goyal, and Bengio]{scholkopf2021towards}
Sch{\"o}lkopf, B., Locatello, F., Bauer, S., Ke, N.~R., Kalchbrenner, N.,
  Goyal, A., and Bengio, Y.
\newblock Towards causal representation learning 2021.
\newblock \emph{arXiv preprint arXiv:2102.11107}, 2021.

\bibitem[Seigal et~al.(2022)Seigal, Squires, and Uhler]{seigal2022linear}
Seigal, A., Squires, C., and Uhler, C.
\newblock Linear causal disentanglement via interventions.
\newblock \emph{arXiv preprint arXiv:2211.16467}, 2022.

\bibitem[Shinners(2011)]{pygame}
Shinners, P.
\newblock Pygame.
\newblock \url{http://pygame.org/}, 2011.

\bibitem[Shinners et~al.(2011)]{shinners2011pygame}
Shinners, P. et~al.
\newblock Pygame.
\newblock \emph{Dostupn{\'e} z: http://pygame. org/[Online (2011)}, 2011.

\bibitem[Von~K{\"u}gelgen et~al.(2021)Von~K{\"u}gelgen, Sharma, Gresele,
  Brendel, Sch{\"o}lkopf, Besserve, and Locatello]{von2021self}
Von~K{\"u}gelgen, J., Sharma, Y., Gresele, L., Brendel, W., Sch{\"o}lkopf, B.,
  Besserve, M., and Locatello, F.
\newblock Self-supervised learning with data augmentations provably isolates
  content from style.
\newblock \emph{Advances in neural information processing systems},
  34:\penalty0 16451--16467, 2021.

\bibitem[Wang \& Jordan(2021)Wang and Jordan]{wang2021desiderata}
Wang, Y. and Jordan, M.~I.
\newblock Desiderata for representation learning: A causal perspective.
\newblock \emph{arXiv preprint arXiv:2109.03795}, 2021.

\bibitem[Yamada et~al.(2022)Yamada, Tang, and Ilker]{yamada2022}
Yamada, Y., Tang, T., and Ilker, Y.
\newblock When are lemons purple? the concept association bias of clip.
\newblock \emph{arXiv preprint arXiv:2212.12043}, 2022.

\bibitem[Yao et~al.(2021)Yao, Sun, Ho, Sun, and Zhang]{yao2021learning}
Yao, W., Sun, Y., Ho, A., Sun, C., and Zhang, K.
\newblock Learning temporally causal latent processes from general temporal
  data.
\newblock \emph{arXiv preprint arXiv:2110.05428}, 2021.

\bibitem[Yao et~al.(2022{\natexlab{a}})Yao, Chen, and Zhang]{yao2022learning}
Yao, W., Chen, G., and Zhang, K.
\newblock Learning latent causal dynamics.
\newblock \emph{arXiv preprint arXiv:2202.04828}, 2022{\natexlab{a}}.

\bibitem[Yao et~al.(2022{\natexlab{b}})Yao, Sun, Ho, Sun, and
  Zhang]{yao2022learning1}
Yao, W., Sun, Y., Ho, A., Sun, C., and Zhang, K.
\newblock Learning temporally causal latent processes from general temporal
  data.
\newblock In \emph{International Conference on Learning Representations},
  2022{\natexlab{b}}.
\newblock URL \url{https://openreview.net/forum?id=RDlLMjLJXdq}.

\bibitem[Zimmermann et~al.(2021)Zimmermann, Sharma, Schneider, Bethge, and
  Brendel]{zimmermann2021contrastive}
Zimmermann, R.~S., Sharma, Y., Schneider, S., Bethge, M., and Brendel, W.
\newblock Contrastive learning inverts the data generating process.
\newblock In \emph{International Conference on Machine Learning}, pp.\
  12979--12990. PMLR, 2021.

\end{thebibliography}
\bibliographystyle{icml2023}

\newpage
\appendix
\onecolumn
\hrule
\begin{center}
  \textbf{\Large Interventional Causal Representation Learning}
  \vskip 0.25cm
  {\large Appendices}
\end{center}
\hrule

\section*{Contents}
We organize the Appendix as follows. 

\begin{itemize}
\item In \Cref{sec:proof_and_details}, we present the proofs for the theorems that were presented in the main body of the paper. 
\begin{itemize} 
\item In \Cref{sec:affine_appendix}, we derive the affine identification guarantees and its approximations in various settings. (Theorem~\ref{thm1})
\item In \Cref{sec: rep_id_do}, we derive the $do$ intervention based identification guarantees and its extensions. (Theorem~\ref{thm2})
\item In \Cref{sec: rep_id_imp}, we present representation identification guarantees for imperfect interventions. (Theorem~\ref{thm3})
\item In \Cref{sec: rep_id_obs}, we present representation identification guarantees for observational data with independent support. (Theorem~\ref{thm4})
\end{itemize}
\item In \Cref{sec:emp_find_append}, we present supplementary materials for the experiments. 
\begin{itemize}
\item In \Cref{sec:method_det_append}, we present the pseudocode for the method used to learn the representations. 
\item In \Cref{sec:setup_poly_decoder_app}, we present the details of the setup used in the experiments with the polynomial decoder $g$.
\item In \Cref{sec:res_poly_decoder_app}, we present supplementary results for the setting with polynomial decoder $g$.
\item In \Cref{sec:setup_images_app}, we present the details of the setup used in the experiments with image data.
\item In \Cref{sec:results_images_app}, we present supplementary results for the setting with image data. 
\end{itemize}
\end{itemize}

\newpage 
\section{Proofs and Technical Details}
\label{sec:proof_and_details}

In this section, we provide the proofs for the theorems. We restate the theorems for convenience. 

\paragraph{Preliminaries and notation.} We state the formal definition of support of a random variable. In most of the work, we operate on the following measure space $(\mathbb{R}^{d}, \mathcal{B}, \lambda)$, $\mathcal{B}$ is the Borel sigma field over $\mathbb{R}^{d}$ and $\lambda$ is the Lebesgue measure over completion of Borel sets on $\mathbb{R}^d$ \citep{ash2000probability}. For a random variable $X$, the support $\mathcal{X} = \{x \in \mathbb{R}^{d}, d\mathbb{P}_X(x) >0\}$, where $d\mathbb{P}_X(x)$ is the Radon-Nikodym derivative of $\mathbb{P}$ w.r.t Lebesgue measure over completion of Borel sets on $\mathbb{R}^d$.  For random variable $Z$, $\mathcal{Z}$ is the support of $Z$ in the observational data. The support of the component $Z_j$ of $Z$ is  $\mathcal{Z}_j$. For random variable $Z$, $\mathcal{Z}^{(i)}$ is the support of $Z$ when $Z_i$ is intervened. The support of the component $Z_j$ of $Z$ in intervened data is  $\mathcal{Z}_j^{(i)}$. 


\subsection{Affine Identification}
\label{sec:affine_appendix}

\begin{lemma}
\label{lemma1}
If the matrix $G$ that defines the polynomial $g$ is full rank and $p>0$, then $g$ is injective.  
\end{lemma}

\begin{proof}
Suppose this is not the case and $g(z_1) = g(z_2)$ for some $z_1\not=z_2$. Thus 
\begin{equation}
\begin{split}
    & G \begin{bmatrix} 1 \\ z_1 \\ z_1\bar{\otimes} z_1 \\ \vdots \\ \underbrace{z_1 \bar{\otimes} \cdots \bar{\otimes}\;z_1}_{p \;\text{times}} \end{bmatrix}  = G \begin{bmatrix} 1 \\ z_2 \\ z_2\bar{\otimes} z_2 \\ \vdots \\ \underbrace{z_2 \bar{\otimes} \cdots \bar{\otimes}\;z_2}_{p \;\text{times}}  \end{bmatrix} \\ 
    \implies & G \begin{bmatrix} 0 \\ (z_1-z_2) \\ z_1\bar{\otimes} z_1 - z_2\bar{\otimes} z_2\\ \vdots \\ \underbrace{z_1 \bar{\otimes} \cdots \bar{\otimes}\;z_1}_{p \;\text{times}} -\underbrace{z_2 \bar{\otimes} \cdots \bar{\otimes}\;z_2}_{p \;\text{times}} \end{bmatrix}  =0
\end{split}
\end{equation}
Since $z_1\not= z_2$ we find a non-zero vector in the null space of $G$ which contradicts the fact that $G$ has full column rank. Therefore, it cannot be the case that $g(z_1) = g(z_2)$ for some $z_1\not=z_2$. Thus $g$ has to be injective. 
\end{proof}

\begin{lemma}
\label{lemma4}
If $v_1$ is a polynomial of degree $k_1$ and $v_2$ is a polynomial of degree $k_2$,  then $v_1 v_2$ is a polynomial of degree $k_1+k_2$. 
\end{lemma}

\begin{proof}

We separate $v_{i}(z)$ into two parts -- the terms with degree $k_{i}$ ($u_{i}(z)$) and the terms with degree less than $k_i$ ($w_{i}(z)$) for $i\in \{1,2\}$. We obtain the following expression.
    \begin{equation}
        v_1(z) v_2(z) = (u_1(z)+w_1(z)) (u_2(z)+w_2(z)) = u_1(z)u_2(z) + u_1(z)w_2(z)  + u_2(z)w_1(z) + w_1(z)w_2(z) 
    \end{equation}
The maximum degree achieved by $u_1(z)u_2(z)$ is $k_1+k_2$. For the other terms, the maximum is bounded above by $k_1+k_2-1$. To prove the result, we need to show that $u_1(z)u_2(z)$ has a degree $k_1+k_2$. 

We first start with a simple case. Suppose $u_1(z)$ and $u_2(z)$ do not share any component of $z$ that they both depend on. In such a case, if we take the leading degree term in $u_1$ and $u_2$ respectively and multiply them then we obtain distinct terms of degree $k_1+k_2$. 

Suppose $u_1$ and $u_2$ both depend on $z_1$.  We write $u_1(z)$ as 
$$ u_1(z) = \sum_{\sum_i d_{ji} =k_1} \theta_j \prod_{i=1}^{d} z_{i}^{d_{ji}} = \sum_{\sum_i d_{ji} =k_1} \theta_j c_j(z)$$
where $c_j(z) = \prod_{i} z_{i}^{d_{ji}}$ is a degree $k_1$ polynomial. Note that for each $j$, $c_j$ is a different polynomial, i.e. for $j\not=q$, $c_j\not=c_q$. 
We write $u_2(z)$ as 
$$ u_2(z) = \sum_{\sum_i d_{ji} =k_2} \beta_j \prod_{i=1}^{d} z_{i}^{d_{ji}} = \sum_{\sum_i d_{ji} =k_2} \beta_j c_j(z)$$

We collect all the terms in $u_1$ that have the highest degree associated with $z_1$ such that  the coefficient $\theta_j$ is non-zero.  We denote the highest degree as $r$ and  write these terms as 
$$\sum_{q}\theta_q z_1^{r}  \prod_{i=2}^{d} z_{i}^{d_{qi}} = \sum_{q}\theta_q z_1^{r} \omega_q(z)$$
where $\omega_q(z) =\prod_{i=2}^{d} z_{i}^{d_{qi}}$, $q\not=l \implies \omega_{q}\not= \omega_l$, and $r\geq 1$

From $u_2(z)$, collect the terms with the highest degree for $z_1$ such that  the coefficient $\beta_j$ is non-zero to obtain.  We denote the highest degree as $s$ and  write these terms as

$$\sum_{t}\beta_t z_1^{s}  \prod_{i=2}^{d} z_{i}^{d_{ti}} = \sum_{t}\beta_t z_1^{s} \eta_t(z)$$
where $\eta_t(z) =\prod_{i=2}^{d} z_{i}^{d_{ti}}$, $t\not=l \implies \eta_{t}\not= \eta_l$, and $s\geq 1$.   

As a result, $u_1(z) u_2(z)$ will contain the term 
$$z_1^{r + s}  \sum_{q}\theta_q \omega_q(z) \sum_{t}\beta_t  \eta_t(z) $$
$$z_1^{r + s}  \delta_1(z)  \delta_2(z)$$

where $\delta_1(z) =  \sum_{q}\theta_q \omega_q(z) $ and $\delta_2(z) = \sum_{t}\beta_t  \eta_t(z) $. We will use principle of induction on the degree of polynomial  to prove the claim.

We first establish the base case for $k_1=1$ and $k_2=1$. 
Consider two polynomials $\rho_1^{\top} z$ and $\rho_2^{\top} z$. We multiply the two to obtain 
$ \sum_{i,j} \rho_{1i} \rho_{2j} z_i z_j$. Consider two cases. In case 1, the two polynomials have at least one non-zero coefficient for the same component $z_i$. In that case, we obtain the only non-zero term with $\rho_{1i}^2 z_i^2$, which establishes the base case. In the second case, the two polynomials have no shared non-zero coefficients. In such a case, each term with a non-zero coefficient is of the form $\rho_{1i} \rho_{2j} z_iz_j$. This establishes the base case. The other cases with $k_1=0$ and $k_2=1$ or $k_2=0$ and $k_1=1$ or both $k_1=0$, $k_2=0$ are trivially true. Thus we have established the base case for all polynomials (with arbitrary dimension for $z$) of degree less than $k_1=1$ and $k_2=1$.

We can now assume that the claim is true for all polynomials $v_1$ with degree less than $k_1-1$ and all polynomials $v_2$ with degree less than $k_2-1$. As a result, the degree of $\delta_1(z) \delta_2(z)$ is $k_1+k_2-r-s$.

We can write $\delta_1 \delta_2$ in terms of the terms with degree equal to $k_1+k_2-r-s$ ($\delta^{'}(z)$) and terms that have a degree less than $k_1+k_2-r-s$ ($\delta^{*}(z)$). As a result, we can simplify $z_1^{r + s}  \delta_1(z)  \delta_2(z)$ to obtain 

\begin{equation}
    z_1^{r+s} (\delta^{'}(z) + \delta^{*}(z)) 
\end{equation}

 The degree of  $z_1^{r+s} \delta^{*}(z)$ is at most $k_1+k_2-1$. The degree of $z_1^{r+s} (\delta^{'}(z))$ has to be $k_1+k_2$ since  $\delta^{'}(z)$ does not depend on $z_1$, $\delta^{'}(z)$ is of degree $k_1+k_2-r-s$. Note that this  is the only term in the entire polynomial $u_1(z)u_2(z)$ that is associated with the highest degree for $z_1$ ($z_1^{r+s}$) since other terms ($c_j, c_j^{'}$) have a smaller degree associated with $z_1$ thus the coefficient of this term cannot be cancelled to zero. Therefore, the degree of the polynomial $u_1u_2$ and hence the degree of $v_1v_2$ is $k_1+k_2$.

\end{proof}

Recall $\hat{z} \triangleq f(x), a \triangleq  f \circ g$. Since $f(x) = f \circ g(z) = a(z) \implies \hat{z} = a(z), $ where $a:\mathcal{Z} \cup \mathcal{Z}^{(i)} \rightarrow \hat{\mathcal{Z}} \cup \hat{\mathcal{Z}}^{(i)}$, and $\hat{\mathcal{Z}} = f(\mathcal{X})$ and $\hat{\mathcal{Z}}^{(i)} = f(\mathcal{X}^{(i)})$. We now show that $a$ is bijective. 

\begin{lemma}
\label{lem:abij}
   Suppose the observational data and interventional data are generated
 from  \Cref{eqn: dgp} and \Cref{eqn: dgp_int} respectively. The mapping $a$ that relates the output of the encoder $f$  written as $\hat{z} $, which solves the reconstruction identity \Cref{eqn: recons}, is related to the true latent $z$ is bijective, where $\hat{z}=a(z)$. 
\end{lemma}

\begin{proof}
Observe that $a$ is surjective by construction. We now need to prove that $a$ is injective. Suppose $a$ is not injective. 
Therefore, there exists $z_1 \in \mathcal{Z}$ and $z_2 \in \mathcal{Z}$, where $z_1\not=z_2$ and  $\hat{z}_1 = a(z_1) = \hat{z}_2= a(z_2)$. Note that $a(z_1) = f (x_1)$, where $x_1 =g(z_1)$ and 
$a(z_2) = f (x_2)$, where $x_2 =g(z_2)$. This implies that $f(x_1) = f(x_2)$. We know that the decoder encoder pair satisfy reconstruction, which means $h \circ f(x_1) = x_1$ and $h \circ f(x_2) = x_2$. Since $f(x_1)=f(x_2)$, we obtain that $x_1=x_2$, which implies that $z_1=z_2$ since $g$ is injective. This contradicts the fact that $z_1\not=z_2$. Therefore, $\hat{z}=a(z)$ is bijective. 
\end{proof}
\AffineIdThm*


\begin{proof}
We start by restating the reconstruction identity. For all $x\in \mathcal{X} \cup \mathcal{X}^{(i)}$
\begin{equation}
\begin{split}
&    h \circ f(x) = x   \\ 
& h(\hat{z}) = g(z)  \\ 
& H \begin{bmatrix} 1\\ \hat{z} \\ \hat{z}\bar{\otimes} \hat{z} \\ \vdots \\ \underbrace{\hat{z} \bar{\otimes} \cdots \bar{\otimes}\;\hat{z}}_{p \;\text{times}}\end{bmatrix} =  G \begin{bmatrix}1 \\ z \\ z\bar{\otimes} z \\ \vdots \\\underbrace{z \bar{\otimes} \cdots \bar{\otimes}\;z}_{p \;\text{times}} \end{bmatrix} \\ 
\end{split}
\label{proof1:eqn1}
\end{equation}

Following the assumptions, $h$ is restricted to be polynomial but $f$ bears no restriction. If $H=G$ and $f=g^{-1}$, we get the ideal solution $\hat{z}=z$, thus a solution to the above identity exists.

Since $G$ has full column rank, we can select $q$ rows of $G$ such that $\tilde{G} \in \mathbb{R}^{q\times q}$ and $\mathsf{rank}(\tilde{G}) = q$. Denote the corresponding matrix $H$ that select the same rows as $\tilde{H}$. We restate the identity in \Cref{proof1:eqn1} in terms of $\tilde{H}$ and $\tilde{G}$ as follows.  For all $z\in \mathcal{Z}\cup \mathcal{Z}^{(i)}$

\begin{equation}
    \begin{split}
        &\tilde{H} \begin{bmatrix} 1 \\ \hat{z} \\ \hat{z}\bar{\otimes} \hat{z} \\ \vdots \\ \underbrace{\hat{z} \bar{\otimes} \cdots \bar{\otimes}\;\hat{z}}_{p \;\text{times}} \end{bmatrix} =  \tilde{G} \begin{bmatrix}1 \\ z \\ z\bar{\otimes} z \\ \vdots \\ \underbrace{z \bar{\otimes} \cdots \bar{\otimes}\;z}_{p \;\text{times}} \end{bmatrix} \\ 
        &\tilde{G}^{-1}\tilde{H}\begin{bmatrix} 1 \\ \hat{z} \\ \hat{z}\bar{\otimes} \hat{z} \\ \vdots \\ \underbrace{\hat{z} \bar{\otimes} \cdots \bar{\otimes}\;\hat{z}}_{p \;\text{times}}  \end{bmatrix} =   \begin{bmatrix}1 \\ z \\ z\bar{\otimes} z \\ \vdots \\ \underbrace{z \bar{\otimes} \cdots \bar{\otimes}\;z}_{p \;\text{times}} \end{bmatrix} \\ 
        &z = \tilde{A} \begin{bmatrix}1 \\ \hat{z} \\ \hat{z}\bar{\otimes} \hat{z} \\ \vdots \\ \underbrace{\hat{z} \bar{\otimes} \cdots \bar{\otimes}\;\hat{z}}_{p \;\text{times}}\end{bmatrix}  \\
        & z = \tilde{A}_1 \hat{z} + \tilde{A}_2 \; \hat{z}\bar{\otimes} \hat{z} + \cdots \tilde{A}_p\; \underbrace{\hat{z} \bar{\otimes} \cdots \bar{\otimes}\;\hat{z}}_{p \;\text{times}} + c, 
    \end{split}
    \label{proof1:eqn2}
\end{equation}
where $\tilde{A}$ is a submatrix of $\tilde{G}^{-1}\tilde{H}$ that describes the relationship between $z$ and polynomial of $\hat{z}$, $\{\tilde{A}_i\}_{i=1}^{p}$ correspond to blocks of rows of $\tilde{A}$. Suppose at least one of $\tilde{A}_2, \cdots, \tilde{A}_p$ is non-zero. Among the matrices $\tilde{A}_2, \cdots, \tilde{A}_p$ which are non-zero, pick the matrix  $\tilde{A}_k$ with largest index $k$. Suppose row $i$ of $\tilde{A}_k$ has some non-zero element. Now consider the element in the row in the RHS of \eqref{proof1:eqn2} corresponding to $z_{i}^{p}$. Observe that $z_{i}^{p}$ is a polynomial of $\hat{z}$ of degree  $kp$, where $k\geq2$ (follows from Lemma \ref{lemma4}). In the LHS, we have a polynomial of degree at most $p$. In the LHS, we have a polynomial of degree at most $p$. The equality between LHS and RHS is true for all $\hat{z}\in f(\mathcal{X} \cup \mathcal{X}^{(i)})$. The difference of LHS and RHS is an analytic function.  From \Cref{assm3: h_poly_new} $f(\mathcal{X} \cup \mathcal{X}^{(i)})$ has a measure greater than zero. Therefore, we leverage \citet{mityagin2015zero} to conclude that the LHS is equal to RHS on entire $\mathbb{R}^d$.   If two polynomials are equal everywhere, then their respective coefficients have to be the same. Based on supposition, RHS has non zero coefficient for terms with degree $kp$ while LHS has zero coefficient for terms higher than degree $p$. This leads to a contradiction.  As a result, none of $\tilde{A}_2, \cdots, \tilde{A}_p$ can be non-zero. Thus $z = \tilde{A}_1 \hat{z} +c$. Next, we show that $\tilde{A}_1$ is invertible, which immediately follows from Lemma  \ref{lem:abij}.



\end{proof}

\subsubsection{Extensions to sparse polynomial $g(\cdot)$}
\label{sec:sparse_poly_app}

Suppose $g(\cdot)$ is a degree $p$ polynomial. Let us define the basis that generates $g$ as 
$$ u(z) = \begin{bmatrix}1 \\ z \\ z\bar{\otimes} z \\ \vdots \\\underbrace{z \bar{\otimes} \cdots \bar{\otimes}\;z}_{p \;\text{times}} \end{bmatrix}$$

Note that the number of terms in $u(z)$ grows as $q=\sum_{r=0}^{p} { r+d-1 \choose d-1}$.  In the previous proof, we worked with 

$$ g(z) = G \begin{bmatrix}1 \\ z \\ z\bar{\otimes} z \\ \vdots \\\underbrace{z \bar{\otimes} \cdots \bar{\otimes}\;z}_{p \;\text{times}} \end{bmatrix} = Gu(z) $$

where $G \in \mathbb{R}^{n \times q}$ was full rank. As a result, $n$ has to be greater than $q$ and also grow at least as $\sum_{r=0}^{p} { r+d-1 \choose d-1}$. In real data, we can imagine that the $g(\cdot)$ has a high degree. However, $g$ can exhibit some structure, for instance sparsity. We now show that our entire analysis continues to work even for sparse polynomials thus significantly reducing the requirment on $n$ to grow as the number of non-zero basis terms in the sparse polynomial. 
We write the basis for the sparse polynomial of degree $p$ as $u^{'}(z)$. $u^{'}(z)$ consists of a subset of terms in $u(z)$. 
We write the sparse polynomial $g(\cdot)$ as 
$$ g(z)  = G u^{'}(z)$$

We formally state the assumption on the decoder in this case as follows.

\begin{assumption}
\label{assm2: g_poly_sparse} 
The decoder $g$ is a polynomial of degree $p$ whose corresponding coefficient matrix $G$ (a.k.a. the weight matrix) has full column rank. Specifically, the decoder $g$ is determined by the coefficient matrix $G$ as follows,
\begin{equation}
    g(z) = Gu^{'}(z)
\end{equation}
where $u^{'}(z)$ consists of a subset of terms in $u(z)$.  $u^{'}(z)$  consists of the degree one term, i.e., $z$ and at least one term of the form $z_i^{o}$, where $o\geq \frac{p+1}{2}$ 
\end{assumption}

\begin{theorem}
\label{thm1_appendix_sparse}
Suppose the observational data and interventional data are generated from  \Cref{eqn: dgp} and \Cref{eqn: dgp_int} respectively under Assumptions \ref{assm1: dgp}, \ref{assm2: g_poly_sparse}. The autoencoder that solves reconstruction identity in \Cref{eqn: recons} under \Cref{assm3: h_poly_new} achieves affine identification, i.e., $\forall z \in \mathcal{Z} \cup \mathcal{Z}^{(i)}, \hat{z} = Az + c$, where $\hat{z}$ is the output of the encoder $f$, $z$ is the true latent, $A$ is an invertible $d\times d$ matrix and $c\in \mathbb{R}^{d}$.
\end{theorem}

\begin{proof}
We start by restating the reconstruction identity. For all $x\in \mathcal{X} \cup \mathcal{X}^{(i)}$
\begin{equation}
\begin{split}
&    h \circ f(x) = x  \\ 
& h(\hat{z}) = g(z)  \\ 
& H \begin{bmatrix} 1\\ \hat{z} \\ \hat{z}\bar{\otimes} \hat{z} \\ \vdots \\ \underbrace{\hat{z} \bar{\otimes} \cdots \bar{\otimes}\;\hat{z}}_{p \;\text{times}}\end{bmatrix} =  Gu^{'}(z) \\ 
\end{split}
\label{proof1_sp:eqn1}
\end{equation}

Following the assumptions, $h$ is restricted to be polynomial but $f$ bears no restriction. If $H$ is equal to the matrix $G$ for columns $i$ where $u_i=u_j^{'}$ for some $j$ and zero in other columns and $f=g^{-1}$, we get the ideal solution $\hat{z}=z$, thus a solution to the above identity exists.
Since $G$ has full column rank, we can select $q$ rows of $G$ such that $\tilde{G} \in \mathbb{R}^{q\times q}$ and $\mathsf{rank}(\tilde{G}) = q$. Denote the corresponding matrix $H$ that select the same rows as $\tilde{H}$. We restate the identity in \Cref{proof1_sp:eqn1} in terms of $\tilde{H}$ and $\tilde{G}$ as follows.  For all $z\in \mathcal{Z}\cup \mathcal{Z}^{(i)}$

\begin{equation}
    \begin{split}
        &\tilde{H} \begin{bmatrix} 1 \\ \hat{z} \\ \hat{z}\bar{\otimes} \hat{z} \\ \vdots \\ \underbrace{\hat{z} \bar{\otimes} \cdots \bar{\otimes}\;\hat{z}}_{p \;\text{times}} \end{bmatrix} =  \tilde{G} u^{'}(z) \\ 
        &\tilde{G}^{-1}\tilde{H}\begin{bmatrix} 1 \\ \hat{z} \\ \hat{z}\bar{\otimes} \hat{z} \\ \vdots \\ \underbrace{\hat{z} \bar{\otimes} \cdots \bar{\otimes}\;\hat{z}}_{p \;\text{times}}  \end{bmatrix} =  u^{'}(z) \\ 
        &z = \tilde{A} \begin{bmatrix}1 \\ \hat{z} \\ \hat{z}\bar{\otimes} \hat{z} \\ \vdots \\ \underbrace{\hat{z} \bar{\otimes} \cdots \bar{\otimes}\;\hat{z}}_{p \;\text{times}}\end{bmatrix}  \\
        & z = \tilde{A}_1 \hat{z} + \tilde{A}_2 \; \hat{z}\bar{\otimes} \hat{z} + \cdots \tilde{A}_p\; \underbrace{\hat{z} \bar{\otimes} \cdots \bar{\otimes}\;\hat{z}}_{p \;\text{times}} + c
    \end{split}
    \label{proof1_sp:eqn2}
\end{equation}

In the simplification above, we rely on the fact that $u'(z)$ consists of the first degree term.
Suppose at least one of $\tilde{A}_2, \cdots, \tilde{A}_p$ is non-zero. Among the matrices $\tilde{A}_2, \cdots, \tilde{A}_p$ which are non-zero, pick the matrix  $\tilde{A}_k$ with largest index $k$. Suppose row $i$ of $\tilde{A}_k$ has some non-zero element.  Now consider the element in the row in the RHS of \eqref{proof1_sp:eqn2} corresponding to $z_{i}^{o}$. Observe that $z_{i}^{o}$ is a polynomial of $\hat{z}$ of degree  $ko$, where $k\geq2$. In the LHS, we have a polynomial of degree at most $p$. The equality between LHS and RHS is true for all $\hat{z}\in f(\mathcal{X} \cup \mathcal{X}^{(i)})$. The difference of LHS and RHS is an analytic function.  From \Cref{assm3: h_poly_new} $f(\mathcal{X} \cup \mathcal{X}^{(i)})$ has a measure greater than zero. Therefore, we leverage \citet{mityagin2015zero} to conclude that the LHS is equal to RHS on entire $\mathbb{R}^d$.  
If two polynomials are equal everywhere, then their respective coefficients have to be the same. Based on supposition, RHS has non zero coefficient for terms with degree $p+1$ while LHS has zero coefficient for terms higher than degree $p$. This leads to a contradiction.  As a result, none of $\tilde{A}_2, \cdots, \tilde{A}_p$ can be non-zero. Thus $z = \tilde{A}_1 \hat{z} +c$. Next, we  need to show that $\tilde{A}_1$ is invertible, which follows from Lemma  \ref{lem:abij}. 
\end{proof}

\subsubsection{Extensions to polynomial $g(\cdot)$ with unknown degree}
\label{sec:poly_unkown_degree_app}

 The learner starts with solving the reconstruction identity by setting the degree of $h(\cdot)$ to be $s$; here we assume $H$ has full rank (this implicitly requires  that $n$ is greater than the number of terms in the polynomial of degree~$s$). 

\begin{equation}
     H \begin{bmatrix} 1 \\ \hat{z} \\ \hat{z} \bar{\otimes} \hat{z} \\ \vdots \\ \underbrace{\hat{z} \bar{\otimes} \cdots \bar{\otimes}\;\hat{z}}_{s \;\text{times}} \end{bmatrix} =  G \begin{bmatrix} 1 \\ z \\ z \bar{\otimes} z \\ \vdots \\ \underbrace{z \bar{\otimes} \cdots \bar{\otimes}\;z}_{p \;\text{times}}\end{bmatrix} \\ 
\end{equation}

We can restrict $H$ to rows such that it is a square invertible matrix $\tilde{H}$. Denote the corresponding restriction of $G$ as $\tilde{G}$. The equality is stated as follows. 

\begin{equation}
\begin{bmatrix} 1 \\ \hat{z} \\ \hat{z}\bar{\otimes} \hat{z} \\ \vdots \\ \underbrace{\hat{z} \bar{\otimes} \cdots \bar{\otimes}\;\hat{z}}_{s \;\text{times}} \end{bmatrix} =  \tilde{H}^{-1}\tilde{G} \begin{bmatrix} 1 \\ z \\ z\bar{\otimes} z \\ \vdots \\ \underbrace{z \bar{\otimes} \cdots \bar{\otimes}\;z}_{p \;\text{times}} \end{bmatrix}
\end{equation} 

If $s>p$, then  $\underbrace{\hat{z} \bar{\otimes} \cdots \bar{\otimes}\;\hat{z}}_{s \;\text{times}}$ is a polynomial of degree at least $p+1$. Since the RHS contains a polynomial of degree at most $p$ the two sides cannot be equal over a set of values of $z$ with positive Lebesgue measure in $\mathbb{R}^{d}$. Thus the reconstruction identity will only be satisfied when $s=p$. Thus we can start with the upper bound and reduce the degree of the polynomial on LHS till the identity is satisfied. 

\subsubsection{Extensions from polynomials to $\epsilon$-approximate polynomials}
\label{sec:approx_poly_app}

We now discuss how to extend \Cref{thm1} to settings beyond polynomial $g$. Suppose $g$ is a function that can be $\epsilon$-approximated by a polynomial of degree $p$ on entire $\mathcal{Z} \cup \mathcal{Z}^{(i)}$. In this section, we assume that we continue to use polynomial decoders $h$ of degree $p$ (with full rank matrix $H$) for reconstruction. We state this as follows. 

\begin{constraint}
\label{assm: h_poly_appendix} 
The learned decoder $h$ is a polynomial of degree $p$ and its corresponding coefficient matrix $h$ is determined by $H$ as follows. For all $z\in \mathbb{R}^d$
\begin{equation}
    h(z) = H[1, z, z\bar{\otimes} z, \cdots, \underbrace{z \bar{\otimes} \cdots \bar{\otimes}\;z}_{p \;\text{times}}]^{\top}
\end{equation}
where $ \bar{\otimes} $ represents the Kronecker product  with all distinct entries. $H$ has a full column rank. 
\end{constraint}

Since we use $h$ as a polynomial, then satisfying the exact reconstruction is not possible. Instead, we enforce  approximate reconstruction as follows.  For all $x\in \mathcal{X} \cup \mathcal{X}^{(i)}$, we want 
\begin{equation}
\|    h \circ f(x) - x\| \leq \epsilon,
\label{approx_recons_identity}
\end{equation}
where $\epsilon$ is the tolerance on reconstruction error. 
Recall $\hat{z} = f(x)$. We further simplify it as $\hat{z} = f \circ g(z) = a(z)$. We also assume that $a$ can be $\eta$-approximated on entire $\mathcal{Z}\cup\mathcal{Z}^{(i)}$ with a polynomial of sufficiently high degree say $q$. We write this as follows. For all $z\in \mathcal{Z} \cup \mathcal{Z}^{(i)}$,

\begin{equation}
    \begin{split}
    \Bigg\|    \hat{z} - \Theta \begin{bmatrix} z \\ z\bar{\otimes} z \\ \vdots \\\underbrace{z \bar{\otimes} \cdots \bar{\otimes}\;z}_{q \;\text{times}} \end{bmatrix}  \Bigg\| \leq \eta ,\\
    \Bigg\|    \hat{z} - \Theta_1 z  - \Theta_2 \;z \bar{\otimes} z - \cdots \Theta_p\; \underbrace{z \bar{\otimes} \cdots \bar{\otimes}\;z}_{q \;\text{times}}\Bigg\| \leq \eta.
    \end{split}
\end{equation}

We want to show that the norm of $\Theta_k$ for all $k\geq 2$ is sufficiently small. We state some assumptions needed in theorem below. 

\begin{assumption}
\label{assm6: reg_pol}
Encoder $f$ does not take values near zero, i.e., $f_i(x)\geq \gamma \eta$ for all $x\in \mathcal{X}\cup \mathcal{X}^{(i)}$ and for all $i \in \{1, \cdots, d\}$, where $\gamma>2$.  The absolute value of each element in $\tilde{H}^{-1}\tilde{G}$ is bounded by a fixed constant.  Consider the absolute value of the singular values of $\tilde{H}$; we assume that the smallest absolute value is strictly positive and bounded below by $\zeta$.  
\end{assumption}

\begin{theorem}
\label{thm1_approx_appendix}
Suppose the true decoder $g$ can be approximated by a polynomial of degree $p$ on entire $\mathcal{Z} \cup \mathcal{Z}^{(i)}$ with approximation error $\frac{\epsilon}{2}$. Suppose $a=f\circ g$ can be approximated by polynomials on entire $\mathcal{Z} \cup \mathcal{Z}^{(i)}$ with $\eta$ error. If $ [-z_{\mathsf{max}}, z_{\mathsf{max}}]^d \subseteq \mathcal{Z} \cup \mathcal{Z}^{(i)}$, where $z_{\mathsf{max}}$ is sufficiently large, and  \Cref{assm1: dgp}, \Cref{assm6: reg_pol} hold, then the polynomial approximation of $a$ (recall $\hat{z}=a(z)$) corresponding to solutions of approximate reconstruction identity in \Cref{approx_recons_identity} under \Cref{assm: h_poly_appendix} is approximately linear, i.e., the norms of the weights on higher order terms are sufficiently small. Specifically, the absolute value of the weight associated with term of degree $k$ decays as $\frac{1}{z_{\max}^{k-1}}$. 
\end{theorem}

\begin{proof} 
\label{proof: thm1}
We start by restating the approximate reconstruction identity. We use the fact that $g$ can be approximated with a polynomial of say degree $p$ to simplify the identity below. For all $x\in \mathcal{X} \cup \mathcal{X}^{(i)}$
\begin{equation}
\begin{split}
& \|    h \circ f(x) - x\| \leq \epsilon \\ 
& \Bigg\|  H \begin{bmatrix} \hat{z} \\ \hat{z}\bar{\otimes} \hat{z} \\ \vdots \\  \underbrace{\hat{z} \bar{\otimes} \cdots \bar{\otimes}\;\hat{z}}_{p \;\text{times}} \end{bmatrix} -  G \begin{bmatrix} z \\ z\bar{\otimes} z \\ \vdots \\ \underbrace{z \bar{\otimes} \cdots \bar{\otimes}\;z}_{p \;\text{times}} \end{bmatrix} \Bigg\| - \Bigg\| G \begin{bmatrix} z \\ z\bar{\otimes} z \\ \vdots \\ \underbrace{z \bar{\otimes} \cdots \bar{\otimes}\;z}_{p \;\text{times}} \end{bmatrix} - g(z) \Bigg\|\leq \epsilon 
\end{split}
\end{equation}

To obtain the second step from the first, add and subtract $G[ z, z\bar{\otimes} z, \cdots,\underbrace{\hat{z} \bar{\otimes} \cdots \bar{\otimes}\;\hat{z}}_{p \;\text{times}}]^{\top} $ and use reverse triangle inequality. Since $H$ is full rank, we select rows of $H$ such that $\tilde{H}$ is square and invertible. The corresponding selection for $G$ is denoted as $\tilde{G}$. We write the identity in terms of these matrices as follows. 
\begin{equation}
\begin{split}
& \Bigg\|  \tilde{H} \begin{bmatrix} \hat{z} \\ \hat{z}\bar{\otimes} \hat{z} \\ \vdots \\ \underbrace{\hat{z} \bar{\otimes} \cdots \bar{\otimes}\;\hat{z}}_{p \;\text{times}} \end{bmatrix} - \tilde{G} \begin{bmatrix} z \\ z\bar{\otimes} z \\ \vdots \\ \underbrace{z \bar{\otimes} \cdots \bar{\otimes}\;z}_{p \;\text{times}}\end{bmatrix} \Bigg\| \leq \frac{3\epsilon}{2}  \\
& \Bigg\|  \begin{bmatrix} \hat{z} \\ \hat{z}\bar{\otimes} \hat{z} \\ \vdots \\ \underbrace{\hat{z} \bar{\otimes} \cdots \bar{\otimes}\;\hat{z}}_{p \;\text{times}}\end{bmatrix} - \tilde{H}^{-1} \tilde{G} \begin{bmatrix} z \\ z\bar{\otimes} z \\ \vdots \\ \underbrace{z \bar{\otimes} \cdots \bar{\otimes}\;z}_{p \;\text{times}} \end{bmatrix} \Bigg\| \leq \frac{3\epsilon}{2|\sigma_{\mathsf{min}}(\tilde{H})|} 
\end{split} 
\label{eqn2:proof_cnt}
\end{equation}
where $|\sigma_{\mathsf{min}}(\tilde{H})|$ is the singular value with smallest absolute value corresponding to the matrix $\tilde{H}$.  In the simplification above, we use the assumption that $g$ is $\frac{\epsilon}{2}$-approximated by a polynomial with matrix $G$ and we also use the fact that $|\sigma_{\mathsf{min}}(\tilde{H})|$ is positive. Now we write that the polynomial that approximates $\hat{z}_i = a_i(z)$ as follows. 

\begin{equation}
   | \hat{z}_i - \theta_1^{\top}z - \theta_2^{\top} z \bar{\otimes} z - \cdots  \theta_{q}^{\top} \underbrace{z \bar{\otimes} \cdots \bar{\otimes}\;z}_{q \;\text{times}}|  \leq \eta
\end{equation}

\begin{equation}
\begin{split}
&     \hat{z}_i \geq \theta_1^{\top}z + \theta_2^{\top} z \bar{\otimes} z  +  \cdots  \theta_{q}^{\top} \underbrace{z \bar{\otimes} \cdots \bar{\otimes}\;z}_{q \;\text{times}} - \eta   \\
&    \hat{z}_i \leq \theta_1^{\top}z + \theta_2^{\top} z \bar{\otimes} z  +  \cdots  \theta_{q}^{\top} \underbrace{z \bar{\otimes} \cdots \bar{\otimes}\;z}_{q \;\text{times}} + \eta  
\label{eqn3:proof_cnt}
\end{split}
\end{equation}

From Assumption \ref{assm6: reg_pol} we know that $\hat{z}_i \geq \gamma\eta$, where $\gamma>2$. It follows from the above equation that \begin{equation}
\label{eqn: approx_gamma_bound}
\begin{split}
& \theta_1^{\top}z + \theta_2^{\top} z \bar{\otimes} z  +  \cdots  \theta_{q}^{\top} \underbrace{z \bar{\otimes} \cdots \bar{\otimes}\;z}_{q \;\text{times}} + \eta \geq \gamma \eta  \\
& \implies \theta_1^{\top}z + \theta_2^{\top} z \bar{\otimes} z  +  \cdots  + \theta_{q}^{\top} \underbrace{z \bar{\otimes} \cdots \bar{\otimes}\;z}_{q \;\text{times}}    - (\gamma-1)\eta \geq 0 \\
& \implies \frac{1}{\gamma-1} \geq \frac{\eta}{\theta_1^{\top}z + \theta_2^{\top} z \bar{\otimes} z  +  \cdots  + \theta_{q}^{\top} \underbrace{z \bar{\otimes} \cdots \bar{\otimes}\;z}_{q \;\text{times}} }
\end{split}
\end{equation} 

For $\hat{z}_i \geq \gamma\eta$, we track how $\hat{z}_{i}^{p}$ grows below.
\begin{equation}
\begin{split} 
& \hat{z}_i \geq \theta_1^{\top}z + \theta_2^{\top} z \bar{\otimes} z + \cdots  \theta_{q}^{\top} \underbrace{z \bar{\otimes} \cdots \bar{\otimes}\;z}_{q \;\text{times}}-\eta \geq (\gamma-2)\eta \geq  0 \\ 
&     \hat{z}_i^{p} \geq (\theta_1^{\top}z + \theta_2^{\top} z \bar{\otimes} z  + \cdots  \theta_{q}^{\top} \underbrace{z \bar{\otimes} \cdots \bar{\otimes}\;z}_{q \;\text{times}} - \eta )^p \\
& \hat{z}_i^{p} \geq (\theta_1^{\top}z + \theta_2^{\top} z \bar{\otimes} z  + \cdots  \theta_{q}^{\top} \underbrace{z \bar{\otimes} \cdots \bar{\otimes}\;z}_{q \;\text{times}})^{p}(1-\frac{1}{\gamma-1})^{p}
\end{split}
\label{eqn4:proof_cnt}
\end{equation}
In the last step of the above simplification, we use the condition in \Cref{eqn: approx_gamma_bound}. We consider $z=[z_{\mathsf{max}}, \cdots, z_{\mathsf{max}}]$. Consider the terms $\theta_{ij} z_{\mathsf{max}}^{k}$ inside the polynomial in the RHS above. We assume all components of $\theta$ are positive.  Suppose $\theta_{ij} \geq \frac{1}{z_{\mathsf{max}}^{k-\kappa-1}}$, where $\kappa\in (0,1)$, then the RHS in \Cref{eqn4:proof_cnt} grows at least $z_{\mathsf{max}}^{(1+\kappa)p} \big(\frac{\gamma-2}{\gamma-1}\big)^{p}$. From \Cref{eqn2:proof_cnt}, $\hat{z}_{i}^{p}$ is very close to degree $p$
polynomial in $z$. Under the assumption that the terms in $\tilde{H}^{-1}\tilde{G}$ are bounded by a constant, the polynomial of degree $p$ grows at at most $z_{\mathsf{max}}^{p}$. 
The difference in growth rates the \Cref{eqn2:proof_cnt} is an increasing function of $z_{\mathsf{max}}$ for ranges where $z_{\mathsf{max}}$ is sufficiently large. Therefore, the reconstruction identity in \Cref{eqn2:proof_cnt} cannot be satisfied for points in a sufficiently small neighborhood of $z=[z_{\mathsf{max}}, \cdots, z_{\mathsf{max}}]$. Therefore, $\theta_{ij} < \frac{1}{z_{\mathsf{max}}^{k-\kappa-1}}$.  We can consider other vertices of the hypercube $\mathcal{Z}$ and conclude that $|\theta_{ij}|<\frac{1}{z_{\mathsf{max}}^{k-\kappa-1}}$.

\end{proof}

\subsection{Representation identification under $do$ interventions}
\label{sec: rep_id_do}

\DoItvID*


\begin{proof}

First note that Assumptions \ref{assm1: dgp}-\ref{assm2: g_poly} hold. Since we solve \Cref{eqn: recons} under \Cref{assm3: h_poly_new}, we can continue to use the result from Theorem \ref{thm1}.  From Theorem \ref{thm1}, it follows that the estimated latents $\hat{z}$ are an affine function of the true $z$. $\hat{z}_k = a^{\top}z + b,$ $\forall z \in \mathcal{Z} \cup \mathcal{Z}^{(i)}$, where $a\in \mathbb{R}^{d}, b \in \mathbb{R}$.

 We consider a $z\in \mathcal{Z}^{(i)}$ such that $z_{-i}$ is in the interior of the support of $\mathbb{P}_{Z_{-i}}^{(i)}$.  
We write $z \in \mathcal{Z}^{(i)}$ as $[z^{*}, z_{-i}]$.  We can write $\hat{z}_k = a_iz^{*} + a_{-i}^{\top}z_{-i}+b$, where $a_{-i}$ is the vector of the values of coefficients in $a$ other than the coefficient of $i^{th}$ dimension, $a_i$ is $i^{th}$ component of $a$, $z_{-i}$ is the vector of values in $z$ other than $z_i$. From the constraint in \Cref{eqn: intv_cons} it follows that for all $z \in \mathcal{Z}^{(i)}$, $\hat{z}_k = z^{\dagger}$. We use these expressions to carry out the following simplification.
\begin{equation}
\begin{split}
 a_{-i}^{\top}z_{-i}  = z^{\dagger}- a_iz^{*}-b \label{proof2:eqn1}
\end{split}
\end{equation}

Consider another  data point  $z^{'}\in \mathcal{Z}^{(i)}$ from the same interventional distribution such that $z_{-i}^{'} = z_{-i}+ \theta e_j$ is in the interior of the support of $\mathbb{P}_{Z_{-i}}^{(i)}$, where $e_j$ is vector with one in $j^{th}$ coordinate and zero everywhere else. From Assumption \ref{assm:span}, we know that there exists a small enough $\theta$ such that $z_{-i}^{'}$ is in the interior.  Since the point is from the same interventional distribution $z_{i}^{'}=z^{*}$.  For $z_{-i}^{'}$ we have
\begin{equation}
\begin{split}
         a_{-i}^{\top}z_{-i}' = z^{\dagger}- a_iz^{*}-b\\ 
         \label{proof2:eqn2}
\end{split}
\end{equation}

We take a difference of the two equations \eqref{proof2:eqn1} and \eqref{proof2:eqn2} to get
\begin{equation}
             a_{-i}^{\top}(z_{-i}-z_{-i}^{'}) = \theta a_{-i}^{\top}e_j = 0.
\end{equation}
From the above, we get that the $j^{th}$ component of $a_{-i}$ is zero. We can repeat the above argument for all $j$ and get that $a_{-i}=0$.  Therefore, $\hat{z}_k = a_iz_i +b$ for all possible values of $z_i$ in $\mathcal{Z} \cup \mathcal{Z}^{(i)}$. 
\end{proof}

\subsubsection{Extension of $do$ interventions beyond polynomials}
\label{sec:beyond_do_itv_app}

In the main body of the paper, we studied the setting where $g$ is a polynomial. We relax the constraint on $g$. We consider settings with multiple $do$ interventional distribution on a target latent. 

We write the DGP for intervention $j \in \{1, \cdots, t\}$ on latent $i$ as 

\begin{equation} 
\begin{split}
&    z_i = z^{*,j} \\
&   z_{-i} \sim \mathbb{P}_{Z_{-i}}^{(i,j)}
\label{eqn: intv_multiple}
\end{split}
\end{equation}

Let $\mathcal{T}=\{z^{*,1}, \cdots, z^{*,t}\}$ be the set of $do$ intervention target values.  We extend the constrained representation learning setting from the main body, where the learner leverages the geometric signature of a single $do$ intervention per latent dimension to multiple $do$ interventional distributions per latent dimension.
\begin{equation} 
\begin{split}
&    h \circ f(x) = x, \qquad \forall x \in \mathcal{X} \cup \mathcal{X}^{(i,j)} \\ 
&    f_k(x) = z^{\dagger,j}, \qquad \;\;\;\forall x \in \mathcal{X}^{(i,j)}, \forall j \in \{1, \cdots, t\} 
\end{split}
 \label{eqn: intv_cons_multi}
\end{equation}

Recall that the $\hat{z} = f(x) = f\circ g(z) = a(z)$. Consider the $k^{th}$ component $\hat{z}_k = a_k(z)$. Suppose $a_k(z)$ is invertible and only depends on $z_i$, we can write it as $a_k(z_i)$. If $\hat{z}_k$ only depends on $z_i$, i.e., $\hat{z}_k =a_k(z_i) $ and $a_k$ is invertible, then the $z_i$ is identified up to an invertible transform.  Another way to state the above property is $\nabla_{z_{-i}}a_k(z)=0$ for all $z_{-i}$. In what follows, we show that it is possible to approximately achieve identification up to an invertible transform. We show that if the number of interventions $t$ is sufficiently large, then $\|\nabla_{z_{-i}}a_k(z)\| \leq \epsilon$ for all $z\in \mathcal{Z}$.

\begin{assumption}
\label{assm: supp_multi_do_int}
       The interior of the support of $z$ in the observational data, i.e., $\mathcal{Z}$, is non-empty. The interior of the support of $z_{-i}$ in the interventional data, i.e., $\mathcal{Z}_{-i}^{(i,j)}$, is equal to the support in observational data, i.e.,  $\mathcal{Z}_{-i}$, for all $j\in \{1, \cdots, t\}$. Each intervention $z^{*,j}$ is sampled from a distribution $\mathbb{Q}$. The support of $\mathbb{Q}$ is equal to the support of $z_i$ in the observational data, i.e., $\mathcal{Z}_i$. The density of $\mathbb{Q}$ is greater than $\varrho$ ($\varrho>0$) on the entire support. 
\end{assumption}

The above assumption states the restrictions on the support of the latents underlying the observational data and the latents underlying the interventional data. 
\begin{assumption}
\label{assm: bdd_second_order}
    $\|\frac{\partial^{2} a(z)}{\partial z_i \partial z_j}\| $ is bounded by $L <\infty$ for all $z \in \mathcal{Z}$ and for all $i,j \in \{1, \cdots, d\}$.  
\end{assumption}

\begin{lemma}
\label{lemma:cov}
    If the number of interventions $t \geq \log (\frac{\delta \epsilon }{2(\beta_{\sup}^i + \beta^{i}_{\inf})}) / \log(1-\varrho\frac{\epsilon}{2})$, then $\max_{z_i\in \mathcal{Z}_i} \min_{z^{*,j} \in \mathcal{T}} \|z_i-z^{*,j}\| \leq \epsilon$ with probability $1-\delta$. 
\end{lemma}

\begin{proof}
Consider the interval $[-\beta^{i}_{\inf}, \beta_{\sup}^i]$, where $\beta^{i}_{\inf}$ and $\beta^{i}_{\sup}$ are the infimum and supremum of $\mathcal{Z}_i$. Consider an $\frac{\epsilon}{2}$ covering of $[-\beta^{i}_{\inf}, \beta_{\sup}^i]$. This covering consists of $\frac{2(\beta_{\sup}^i + \beta^{i}_{\inf})}{\epsilon}$ equally spaced points at a separation of $\epsilon/2$.  Consider a point $z_i$, its nearest neighbor in the cover is denoted as $z_l^{'}$, and the nearest neighbor of $z_i$ in the set of interventions $\mathcal{T}$ is $z^{*,j}$.  The nearest neighbor of $z_l^{'}$ in the set of interventions is $z^{*,r}$. Since $  \|z_i-z^{*,j}\| \leq \|z_i-z^{*,q}\| $ for all $q \in \{1, \cdots, t\}$ we can write 
\begin{equation}
    \begin{split}
        \|z_i-z^{*,j}\| \leq \|z_i-z^{*,r}\| \leq \|z_i - z_l^{'}\| + \| z_l^{'}-z^{*,r}\|  \leq \frac{\epsilon}{2} + \| z_l^{'}-z^{*,r}\| 
    \end{split}
\end{equation}

Observe that if $\| z_l^{'}-z^{*,r}\|$ is less than $\frac{\epsilon}{2}$ for all $z_l^{'}$ in the cover, then for all $z_i$ in $\mathcal{Z}_i$, $ \|z_i-z^{*,j}\| $ is less than $\epsilon$.
We now show that $\| z_l^{'}-z^{*,r}\|$ is sufficiently small provided $t$ is sufficiently large. Observe that 

$$\mathbb{P}(\| z_l^{'}-z^{*,r}\| >\frac{\epsilon}{2}) \leq (1-\varrho\frac{\epsilon}{2})^{t}$$

We would like that $(1-\varrho\frac{\epsilon}{2})^{t} \leq \delta$, which implies $t\geq \log (\delta) / \log(1-\varrho\frac{\epsilon}{2})$. Therefore, if $t\geq \log (\delta) / \log(1-\varrho\frac{\epsilon}{2})$, then $\mathbb{P}(\| z_l^{'}-z^{*,r}\| \leq \frac{\epsilon}{2}) $ with a probability at least $1-\delta$. If we set $\delta = \frac{\delta \epsilon }{2(\beta_{\sup}^i + \beta^{i}_{\inf})}$, then we obtain that for all $l$, $\mathbb{P}(\| z_l^{'}-z^{*,r}\| \leq \frac{\epsilon}{2}) $  with probability at least $1-\delta$. 
The final expression for $t \geq \log (\frac{\delta \epsilon }{2(\beta_{\sup}^i + \beta^{i}_{\inf})}) / \log(1-\varrho\frac{\epsilon}{2}) $
\end{proof}
\begin{theorem}
\label{thm_do_appendix} Suppose the observational data and interventional data are generated from  \Cref{eqn: dgp} and \Cref{eqn: intv_multiple} respectively. If  the number of interventions $t$ is sufficiently large, i.e., $t\geq \log (\frac{\delta \epsilon }{2L(\beta_{\sup}^i + \beta^{i}_{\inf})}) / \log(1-\varrho\frac{\epsilon}{2L})$, \Cref{assm: supp_multi_do_int} and \Cref{assm: bdd_second_order} are satified,  then the solution to \Cref{eqn: intv_cons_multi} identifies the intervened latent $z_{i}$ approximately up to an invertible tranform, i.e., $\|\nabla_{z_{-i}}a_k(z)\|_{\infty} \leq \epsilon$ for all $z\in \mathcal{Z}$. 
\end{theorem}

\begin{proof}
    Recall $\hat{z} = f(x) = f \circ g (z) = a(z)$, where $a: \cup_{j}\mathcal{Z}^{(i,j)} \cup \mathcal{Z} \rightarrow \cup_{j}\hat{\mathcal{Z}}^{(i,j)} \cup \hat{\mathcal{Z}}$. Consistent with the notation used earlier in the proof of Theorem \ref{thm1}, $\hat{\mathcal{Z}}^{(i,j)} = f(\mathcal{X}^{(i,j)})$. In Lemma \ref{lem:abij}, we had shown that $a$ is bijective, we can use the same recipe here and show  that $a$ is bijective. 
    
    Owing to the constraint in \Cref{eqn: intv_cons_multi}, we claim that $\nabla_{z_{-i}}a_k(z)=0$ for all $z_{-i}$ in the interior of $\mathcal{Z}_{-i}$ with $z_i=z^{*,j}$.     Consider a ball around $z_{-i}$ that is entirely contained in $\mathcal{Z}_{-i}$, denote it as $\mathcal{B}_z$.  From \Cref{eqn: intv_cons_multi}, it follows that $f_k(x)$ takes the same value on this neighborhood. As a result,  $a_k(z)$ is equal to a constant on the ball $\mathcal{B}_z$.  Therefore, it follows that $\nabla_{z_{-i}}a_k(z) = 0$ on the ball $\mathcal{B}_z$. We can extend this argument to all the points in the interior of the support of $z_{-i}$. As a result, $\nabla_{z_{-i}}a_k(z) = 0$ on the interior of the support of $z_{-i}$. Further, $\nabla_{z_{-i}}a_k(z) = 0$ for all $z=[z^{*,j}, z_{-i}]$  in  $\cup_{j}\mathcal{Z}^{(i,j)}$.  Define $\aleph(z)=\nabla_{z_{-i}}a_k(z)$. Consider the $j^{th}$ component of $\aleph(z)$ denoted as $\aleph_j(z)$. 
Consider a point $z\in \mathcal{Z}$ and find its nearest neighbor in $\cup_{j}\mathcal{Z}^{(i,j)}$ and denote it as $z^{'}$. Following the assumptions, $z^{'}_{-i}=z_{-i}$. We expand $\aleph_j(z)$ around $z^{'}$ as follows
\[\aleph_j(z)= \aleph_{j}(z^{'}) + \nabla_{z}\aleph_j(z^{''})^{\top}(z-z^{'})\]
\[\aleph_j(z)=   \frac{\partial \aleph_j(z^{''})}{\partial z_i}(z_{i}-z_{i}^{'})\]
In the above, we use the fact that $\aleph_{j}(z^{'})=0$. 
\[|\aleph_j(z)| =   \bigg| \frac{\partial \aleph_j(z^{''})}{\partial z_i}(z_{i}-z_{i}^{'})\bigg| \leq  \bigg| \frac{\partial \aleph_j(z^{''})}{\partial z_i}\bigg| \frac{\epsilon}{L}  \leq \epsilon \] 
To see the last inequality in the above, use \Cref{lemma:cov} with $\epsilon$ as $\epsilon/L$ and \Cref{assm: bdd_second_order}.  
\end{proof}

In the discussion above, we showed that multiple $do$ interventional distribution on target latent dimension help achieve approximate identification of a latent up to an invertible transform. The above argument extends to all latents provided we have data with multiple do interventional distributions per latent. We end this section by giving some intuition as to why multiple interventions are necessary in the absence of much structure on $g$. 

\paragraph{Necessitating multiple interventions}
We consider the case with one $do$ intervention. Consider the set of values achieved under intervention, where $z_{-i}$ is from the interior of $\tilde{\mathcal{Z}}^{(i)}_{-i}$. We call this set $\tilde{\mathcal{Z}}^{(i)}$
Suppose $a$ is a bijection of the following form. 
\begin{equation}
    a = \begin{cases} \mathsf{I}, \text{if} \; z \; \text{is in} \; \tilde{\mathcal{Z}}^{(i)} \\ 
    \tilde{a}\;\; \text{otherwise}
    \end{cases}
\end{equation}
where $\mathsf{I}$ is identity function and $\tilde{a}$ is an arbitrary bijection with bounded second order derivative (satisfying \Cref{assm: bdd_second_order}). Define $f = a\circ g^{-1} $ and $h=g\circ a^{-1}$. Observe that these $f$ and $h$ satisfy both the constraints in the representation learning problem in \Cref{eqn: intv_cons}. In the absence of any further assumptions on $g$ or structure of support of $\mathcal{Z}$, each intervention enforces local constraints on $a$.

\subsection{Representation identification under general perfect and imperfect interventions}
\label{sec: rep_id_imp}
Before proving Theorem \ref{thm3}, we prove a simpler version of the theorem, which we leverage to prove Theorem \ref{thm3}. We start with the case when the set $\mathcal{S}$ has one element say $\mathcal{S}=\{j\}$.  

\begin{assumption}
\label{assm5}
Consider the $Z$ that follow the interventional distribution $\mathbb{P}_{Z}^{(i)}$.   The joint support of $z_i,z_j$  satisfies factorization of support, i.e., 
\begin{equation}
    \mathcal{Z}_{i,j}^{(i)} =  \mathcal{Z}_{i}^{(i)} \times   \mathcal{Z}_{j}^{(i)}
\end{equation}
For all $j \in [d]$, $-\infty <\alpha_{\inf}^{j} \leq \alpha_{\sup}^{j} < \infty$. There exists a $\zeta>0$ such that the all the points in $\Pi_{j \in [d]} (\alpha_{\sup}^{j} -\zeta, \alpha_{\sup}^{j}) \cup (\alpha_{\inf}^{j}, \alpha_{\inf}^{j}+\zeta)$ are in $\mathcal{Z}^{(i)}$.
\end{assumption}

The above assumption only requires support independence for two random variables $Z_i$ and $Z_j$. 

We now describe a constraint, where the learner enforces support independence between $\hat{z}_i$ and $\hat{z}_j$. 
\begin{constraint}
     \label{eqn: intv_cons_soft_1}
The pair $(\hat{z}_i, \hat{z}_j)$ satisfies support independence on interventional data, i.e.,  
 $$ \hat{\mathcal{Z}}^{(i)}_{i,j} = \hat{\mathcal{Z}}^{(i)}_{i} \times \hat{\mathcal{Z}}^{(i)}_{j}$$
\end{constraint}

In the above \Cref{eqn: intv_cons_soft_1}, we use same indices $i$ and $j$ as in \Cref{assm5} for convenience, the arguments extend to the case where we use a different pair.

\begin{theorem}
\label{thm3_pair}
 Suppose the observational data and interventional data are generated from  \Cref{eqn: dgp} and \Cref{eqn: dgp_int} respectively under Assumptions \ref{assm1: dgp}, \ref{assm2: g_poly}, \ref{assm5}. The autoencoder that solves \Cref{eqn: recons} under Constraint \ref{assm3: h_poly_new}, \ref{eqn: intv_cons_soft_1} achieves block affine identification, i.e., $\forall z \in \mathcal{Z}, \hat{z} = Az + c$, where $\hat{z}$ is the output of the encoder $f$ and $z$ is the true latent and $A$ is an invertible $d\times d$ matrix and $c\in \mathbb{R}^{d}$. Further, the matrix $A$ has a special structure, i.e., the row $a_i$ and $a_j$ do not have a non-zero entry in the same column. Also, each row $a_i$ and $a_j$ has at least one non-zero entry.
\end{theorem}

\begin{proof}
Let us first verify that there exists a solution to \Cref{eqn: recons} under Constraint \ref{assm3: h_poly_new}, \ref{eqn: intv_cons_soft_1}. If $\hat{Z} =  Z$  and $h = g $, then that suffices to guarantee that a solution  exists. 

First note that since Assumptions \ref{assm1: dgp}, \ref{assm2: g_poly} holds and we are solving \Cref{eqn: recons} under \Cref{assm3: h_poly_new}, we can continue to use the result from Theorem \ref{thm1}.  From Theorem \ref{thm1}, $\forall z \in \mathcal{Z} \cup \mathcal{Z}^{(i)}, \hat{z} = Az + c$, where $\hat{z}$ is the output of the encoder $f$ and $z$ is the true latent and $A$ is an invertible $d\times d$ matrix and $c\in \mathbb{R}^{d}$. 

From Assumption \ref{assm5} we know each component $k \in \{1, \cdots, d\}$ of $z$, $z_k$ is bounded above and below. Suppose the minimum and maximum value achieved by $z_k \in \mathcal{Z}^{(i)}_k$ is $\alpha^{k}_{\inf}$ and the maximum value achieved by $z_k \in \mathcal{Z}^{(i)}_k$ is $\alpha_{\sup}^{k}$ . 

Define a new latent $$z_{k}^{'} = 2\bigg(\frac{z_{k} - \frac{\alpha^{k}_{\sup} + \alpha^{k}_{\inf}}{2}}{\alpha^{k}_{\sup}-\alpha^{k}_{\inf}}\bigg), \forall k \in \{1, \cdots, d\}$$ 

Notice post this linear operation, the new latent takes a maximum value of $1$ and a minimum value of $-1$. 

We start with $\hat{z} = Az^{'} + c$, where $z^{'}$ is element-wise transformation of $z$ that brings its maximum and minimum value of each component to $1$ and $-1$. Following the above transformation, we define the left most interval for $z_i^{'}$ as $[-1, -1+\eta_{i}]$ and the rightmost interval is $[1-\zeta_i, 1]$, where $\eta_i>0$ and $\zeta_i>0$. Such an interval exists owing to the Assumption \ref{assm5}.  

Few remarks are in order. i) Here we define intervals to be closed from both ends. Our arguments also extend to the case if these intervals are open from both ends or one end, 
ii) We assume all the values in the interval $[-1, -1+\eta_{i}]$ are in the support. The argument presented below extends to the case when all the values in $[-1, -1+\eta_{i}]$ are assumed by  $z_i^{'}$ except for a set of measure zero, iii) The assumption \ref{assm5} can be relaxed by replacing supremum and infimum with essential supremum and infimum.

For a sufficiently small $\kappa$, we claim that the marginal distribution of $\hat{z}_i$ and $\hat{z}_j$ contain the sets defined below. Formally stated
\begin{equation}
    (-\|a_i\|_1+c_i, -\|a_i\|_1+c_i +\kappa) \cup (\|a_i\|_1+c_i -\kappa, \|a_i\|_1+c_i) \subseteq \hat{\mathcal{Z}}^{(i)}_{i}
    \label{eqn: int1}
\end{equation}
\begin{equation}
    (-\|a_j\|_1+c_j, -\|a_j\|_1+c_j +\kappa) \cup (\|a_j\|_1+c_j -\kappa, \|a_j\|_1+c_j) \subseteq \hat{\mathcal{Z}}^{(i)}_{j}
       \label{eqn: int2}
\end{equation}
where $a_i$ and $a_j$ are $i^{th}$ and $j^{th}$ row in matrix $A$.  We justify the above claim next. Suppose all elements of  $a_{i}$ are positive. We set $\kappa$ sufficiently small such that $\frac{\kappa}{|a_{ik}|d} \leq \eta_k$ for all $k\in \{1, \cdots, d\}$. Since $\kappa$ is sufficiently small,  $[-1,-1 + \frac{\kappa}{ |a_{ik}|d}]$ in the support $z^{'}_k$, this holds for all $k \in \{1, \cdots, d\}$. As a result,  $(-\|a_i\|_1+c_i , -\|a_i\|_1+c_i +\kappa)$ is in the support of $\hat{z}_i$.  We can repeat the same argument when the signs of $a_{i}$ are not all positive by adjusting the signs of the elements $z^{'}$.  This establishes $(-\|a_i\|_1+c_i, -\|a_i\|_1+c_i +\kappa) \subseteq \hat{\mathcal{Z}}^{(i)}_{i}$.  Similarly, we can also establish that $(\|a_i\|_1+c_i -\kappa, \|a_i\|_1+c_i) \subseteq \hat{\mathcal{Z}}^{(i)}_{i}$.

Suppose the two rows $a_i$ and $a_j$ share at least $q\geq 1$ non-zero entries. 
Without loss of generality assume that $a_{i1}$ is non-zero and $a_{j1}$ is non-zero. Pick an $0<\epsilon<\kappa$
\begin{itemize}
    \item Suppose $a_{i1}$ and $a_{j1}$ are both positive. In this case, if $\hat{z}_i<-\|a_i\|_1+c_i + \epsilon$, then $$z_{1}^{'}<-1 + \frac{2\epsilon}{|a_{i1}|}$$
    To see why is the case, substitute $z_{1}^{'}=-1 + \frac{2\epsilon}{|a_{i1}|}$ and observe that $\hat{z}_i>-\|a_i\|_1+c_i + \epsilon$. 
    \item  Suppose $a_{i1}$ and $a_{j1}$ are both positive. In this case, if $\hat{z}_j>\|a_j\|_1+c_j - \epsilon$, then $$z_{1}^{'}>1 - \frac{2\epsilon}{|a_{j1}|}$$
\end{itemize}
For sufficiently small $\epsilon$ ($\epsilon < \frac{1}{1/|a_{i1}| + |a_{j1}|} $) both $z_{1}^{'}<-1 + \frac{2\epsilon}{a_{i1}}$ and $z_{1}^{'}>1 - \frac{2\epsilon}{a_{j1}}$ cannot be true simultaneously. 

Therefore, $\hat{z}_i<-\|a_i\|_1+c_i + \epsilon$ and $\hat{z}_j>\|a_j\|_1+c_j - \epsilon$ cannot be true simultaneously. Individually,  $\hat{z}_i<-\|a_i\|_1+c_i + \epsilon$ occurs with a probability greater than zero; see \Cref{eqn: int1}. Similarly, $\hat{z}_j>\|a_j\|_1+c_j - \epsilon$ occurs with a probability greater than zero; see \Cref{eqn: int2}. This contradicts the support independence constraint. For completeness, we present the argument for other possible signs of $a$. 

\begin{itemize}
    \item Suppose $a_{i1}$ is positive and $a_{j1}$ is negative. In this case, if $\hat{z}_i<-\|a_i\|_1+c_i + \epsilon$, then $$z_{1}^{'}<-1 + \frac{2\epsilon}{|a_{i1}|}$$
    \item  Suppose $a_{i1}$ is positive and $a_{j1}$ is negative. In this case, if $\hat{z}_j<-\|a_j\|_1+c_j + \epsilon$, then $$z_{1}^{'}>1 - \frac{2\epsilon}{|a_{j1}|}$$
\end{itemize}

Rest of the above case is same as the previous case.  We can apply the same argument to any shared non-zero component. Note that a row $a_i$ cannot have all zeros or all non-zeros (then $a_j$ has all zeros). If that is the case, then matrix $A$ is not invertible.  This completes the proof.
\end{proof}

We now use the result from Theorem~\ref{thm3_pair} to prove the Theorem~\ref{thm3}.

\GeneralItvID*


\begin{proof}
Let us first verify that there exists a solution to  \Cref{eqn: recons} under Constraint \ref{assm3: h_poly_new}, \ref{eqn: intv_cons_soft} (with $|\mathcal{S}^{'}| \leq |\mathcal{S}|$). 

We write $\hat{Z} =  \Pi Z$, where $\Pi$ is a permutation matrix such that  $\hat{Z}_k = Z_i$. For each $m \in \mathcal{S}^{'}$  there exists a unique $j \in \mathcal{S}$ such that $\hat{Z}_m = Z_j$. Suppose $h = g \circ \Pi^{-1}$. Observe that this construction satisfies the constraints in  \Cref{eqn: intv_cons_soft}. 

To show the above claim, we leverage \Cref{thm3_pair}. We apply \Cref{thm3_pair} to all the pairs in $\{(k,m), \forall m \in \mathcal{S}^{'}\}$, we obtain the following. 
We write $\hat{z}_k = a_{k}^{\top} z + c_k$. Without loss of generality, assume $a_{k}$ is non-zero in first $s$ elements. Now consider any $\hat{z}_m= a_{m}^{\top}z  +c_m$, where $m \in \mathcal{S}^{'}$. From \Cref{thm3_pair} it follows that $a_{m}[1:s]=0$. This holds true for all $m \in \mathcal{S}^{'}$.  Suppose $s\geq d-|\mathcal{S}^{'}| + 1$. In this case, the first $s$ columns cannot be full rank. Consider the submatrix formed by the first $s$ columns. In this submatrix $|\mathcal{S}^{'}|$ rows are zero. The maximum rank of this matrix is $d-|\mathcal{S}^{'}|$. If $s\geq d-|\mathcal{S}^{'}|+1$, then this submatrix would not have a full column rank, which contradicts the fact that $A$ is invertible. Therefore, $1 \leq s\leq d-|\mathcal{S}^{'}| $.
\end{proof}
We can relax the assumption that $|\mathcal{S}^{'}| \leq |\mathcal{S}|$ in the above theorem. We follow an iterative procedure. We start by solving \Cref{eqn: intv_cons_soft} with $|\mathcal{S}^{'}|=d-1$. If a solution exists, then we stop. If a solution does not exist, then we reduce the size of $|\mathcal{S}^{'}|$ by one and repeat the procedure till we find a solution. As we reach $|\mathcal{S}^{'}|=|\mathcal{S}|$ a solution has to exist.

\subsection{Representation identification with observational data under independent support}
\label{sec: rep_id_obs}

\ObsSuppID*


\begin{proof}

We will leverage Theorem \ref{thm3_pair} to show this claim.  Consider $\hat{z}_i = a_i^{\mathsf{T}}z + c_i$. We know that the $a_i$ has at least one non-zero element. Suppose it has at least $q\geq 2$ non-zero elements. Without loss of generality assume that these correspond to the first $q$ components. We apply Theorem \ref{thm3_pair} to each pair $\hat{z}_i, \hat{z}_j$ for all $j\not=i$. Note here $i$ is kept fixed and then Theorem \ref{thm3_pair} is applied to every possible pair. From the theorem we get that $a_{j}[1:q]$ is zero for all $j \not=i$. If $q\geq 2$, then the span of first $q$ columns will be one dimensional and as a result $A$ cannot be invertible. Therefore, only one element of row $i$ is non-zero. We apply the above argument to all $i\in \{1,\cdots, d\}$. We write a function $\pi:\{1, \cdots, d\} \rightarrow \{1, \cdots, d\}$, where $\pi(i)$ is the index of the element that is non-zero in row $i$, i.e.,  $\hat{z}_i = a_{i\pi(i)}z_{\pi(i)} + c_i$. Note that $\pi$ is injective, if two indices map to the same element, then that creates shared non-zero coefficients, which violates Theorem \ref{thm3_pair}. This completes the proof. 
\end{proof}

\newpage

\section{Supplementary Materials for Empirical Findings}
\label{sec:emp_find_append}

\subsection{Method details}
\label{sec:method_det_append}
\begin{algorithm}[th]
\begin{algorithmic}[1]
\STATE \textbf{\COMMENT{Step 1: Training autoencoder ($f, h$)}} 
\STATE Sample data: $X \sim \mathcal{X} \cup \mathcal{X}^{I}$ where $\mathcal{X}^{I}= \cup_{i=1}^{d} \mathcal{X}^{(i)}$
\STATE Minimize reconstruction loss: $f^{\dagger},h^{\dagger} =  \argmin_{f,h} \mathbb{E} \big[\|h \circ f (X) -X\|^2\big] $
\STATE
\STATE \textbf{\COMMENT{Step 2: Learning transformation $\Gamma$ with Independence of Support (IOS) objective}}
\STATE Sample data: $\hat{Z} \sim f^{\dagger}(\mathcal{X})$ where $f^{\dagger}$ is the encoder learnt in Step 1
\STATE Minimize reconstruction + Hausdorff loss: $ \min_{\Gamma}  \mathbb{E} \big[\| \Gamma^{'} \circ \Gamma (\hat{Z}) - \hat{Z} \|^2\big]  + \lambda \times \sum_{k\not=m } \mathsf{HD}\big(\hat{\mathcal{Z}}_{k,m}(\Gamma), \hat{\mathcal{Z}}_{k}(\Gamma) \times \hat{\mathcal{Z}}_{m}(\Gamma)\big) $
\STATE Return transformed latents: $\Gamma (\hat{Z})$
\STATE
\STATE \textbf{\COMMENT{Step 2: Learning transformation $\Gamma = [\gamma_i]_{i=1:d}$ using do-interventions}}
\FOR{$i$ in $\{1, \cdots, d\}$}
\STATE Sample data: $\hat{Z} \sim f^{\dagger}(\mathcal{X}^{(i)})$ where $f^{\dagger}$ is the encoder learnt in Step 1s
\STATE Fix intervention targets at random $\hat{Y}^{(i)} \sim \text{Uniform}(0,1)$
\STATE Minimize MSE loss: $\min_{\gamma_{i}} \mathbb{E}_{\hat{Z}}\big[\big\|\gamma_{i} ( \hat{Z} )  - \hat{Y}^{(i)} \big\|^2\big] $
\ENDFOR
\STATE Return transformed latents: $\Gamma (\hat{Z})$
\end{algorithmic}
\caption{Summarizing our two step approach for both the independence of support (IOS) and interventional data case.}
\label{main-algo-app}
\end{algorithm}

We provide details about our training procedure in Algorithm~\ref{main-algo-app}. For learning with the independence of support (IOS) objective in Step 2, we need to ensure that the map $\Gamma$ is invertible, hence we minimize a combination of reconstruction loss with Hausdorff distance, i.e., 
\begin{equation}
\min_{\Gamma}  \mathbb{E} \big[\| \Gamma^{'} \circ \Gamma (\hat{Z}) - \hat{Z} \|^2\big]  + \lambda \times \sum_{k\not=m } \mathsf{HD}\big(\hat{\mathcal{Z}}_{k,m}(\Gamma), \hat{\mathcal{Z}}_{k}(\Gamma) \times \hat{\mathcal{Z}}_{m}(\Gamma)\big) 
\label{eqn: ae_hd}
\end{equation}
where $\hat{Z}$ denotes the output from the encoder learnt in Step 1, i.e., $\hat{Z}= f^{\dagger} (X)$.

If we have data with multiple interventional distributions per latent dimension, then we sample a new target for each interventional distribution.  In our polynomial decoder experiments, we use a linear $\gamma_i$. In our image based experiments, in Step 2, we use a non-linear map $\gamma_i$.

\subsection{Experiment setup details: Polynomial decoder ($g$)}
\label{sec:setup_poly_decoder_app}

\paragraph{Basic setup.} We sample data following the DGP described in Assumption~\ref{assm2: g_poly} with the following details:

\begin{itemize}
    \item Latent dimension: $d \in \{ 6, 10 \}$
    \item Degree of decoder polynomial ($g$): $p \in \{2, 3\}$
    \item Data dimension: $n= 200$
    \item Decoder polynomial coefficient matrix $G$: sample each element of the matrix iid from a standard normal distribution.  
\end{itemize}

\paragraph{Latent distributions.} Recall $z_i$ is the $i^{th}$ component of the latent vector $z \in \mathbb{R}^{d}$. The various latent distributions ($\mathbb{P}_Z$) we use in our experiments are as follows:

\begin{itemize}
    \item \textbf{Uniform:} Each latent component $z_i$ is sampled from Uniform(-5, 5). All the latents ($z_i$) are independent and identically distributed. \\
    
    \item \textbf{Uniform-Correlated:}  Consider a pair of latent variables $z_i, z_{i+1}$ and sample two confounder variables $c_1, c_2$ s.t. $c_1 \sim \text{Bernoulli}(p=0.5)$, and $c_2 \sim \text{Bernoulli}(p=0.9)$. Now we sample $z_i, z_{i+1}$ using $c_1, c_2$ as follows:
    $$     
    z_{i} \sim \begin{cases}
        \text{Uniform}(0.0, 0.5) & \text{if} \;  c_1 = 1  \\
        \text{Uniform}(-0.5, 0.0) & \text{if} \;  c_1 = 0
    \end{cases}, 
    $$
    $$     
    z_{i+1} \sim \begin{cases}
        \text{Uniform}(0.0, 0.3) & \text{if} \;  c_1 \oplus c_2 =1  \\
        \text{Uniform}(-0.3, 0.0) & \text{if} \;  c_1 \oplus c_2 = 0
    \end{cases}, 
    $$
where $\oplus$ is the xor operation.  Hence, $c_1$ acts as a confounder as it is involved in the generation process for both $z_i, z_{i+1}$, which leads to correlation between them. Due to the xor operation, the two random variables satisfy independence of support condition.    Finally, we follow this generation process to generate the latent vector $z$ by iterating over different pairs ($i \in \{\ 1, \cdots, d\}$ with step size 2 ).

   \item \textbf{Gaussian-Mixture:} Each $z_i$ is sampled from a Gaussian mixture model with two components and equal probability of sampling from the components, as described below:
    $$     
    z_{i} \sim \begin{cases}
        \mathcal{N}(0, 1) & \text{with prob.} \;  0.5  \\
        \mathcal{N}(1, 2) & \text{with prob.} \;  0.5
    \end{cases}
    $$
 All latents in this case are independent and identically distributed like the Uniform case; though we have mixture distribution instead of single mode distribution. \\

    \item \textbf{SCM-S:} The latent variable $z$ is sampled as a DAG with $d$ nodes using the Erdős–Rényi scheme with linear causal mechanism and Gaussian noise~\citep{brouillard2020differentiable} \footnote{\url{https://github.com/slachapelle/dcdi}} and set the expected density (expected number of edges per node) to be 0.5.  
    \item \textbf{SCM-D:}  The latent variable $z$ is sampled as a DAG with $d$ nodes using the Erdős–Rényi scheme with linear causal mechanism and Gaussian noise~\citep{brouillard2020differentiable} and set the expected density (expected number of edges per node) to be 1.0. 
    
\end{itemize}

\begin{table}[h]
    \centering
    \begin{tabular}{c|c|c|c}
        \toprule
        Case & Train & Validation & Test \\
        \midrule
        Observational ($\mathcal{D}$) & 10000 & 2500 & 20000  \\
        Interventional ($\mathcal{D}^{(I)}$) & 10000 & 2500 & 20000 \\
        \bottomrule
    \end{tabular}
    \caption{Statistics for the synthetic poly-DGP experiments}
    \label{tab:data-stats-syn-ploy-dgp}
\end{table}

\paragraph{Further details on dataset and evaluation.} For experiments in Table~\ref{table1_results}, we only use observational data ($\mathcal{D}$); while for experiments in Table~\ref{table2_results}, we use both observational and interventional data ($\mathcal{D} \cup  \mathcal{D}^{(i)}$), with details regarding the train/val/test split described in Table~\ref{tab:data-stats-syn-ploy-dgp}.

We carry out $do$ interventions on each latent with $\mathcal{D}^{(i)}$ corresponding to data from interventions on $z_i$.  The union of data from interventions across all latent dimensions is denoted as $\mathcal{D}^{(I)}= \cup_{i=1:d} \mathcal{D}^{(i)}$.  The index of the variable to be intervened is sampled from $\text{Uniform}(\{1,\dots, d\})$. The selected latent variable to be intervened is set to value $2.0$. 

Further, note that for learning the linear transformation ($\gamma_i$) in Step 2 (\Cref{eqn: cons_int}), we only use the corresponding interventional data ($\mathcal{D}^{(i)}$) from do-intervention on the latent variable $i$. Also, all the metrics ($R^2$, MCC (IOS), MCC, MCC (IL)) are computed only on the test split of observational data ($\mathcal{D}$) (no interventional data used).

\paragraph{Model architecture.} We use the following architecture for the encoder $f$ across all the experiments with polynomial decoder $g$ (Table~\ref{table1_results}, Table~\ref{table2_results}) to minimize the reconstruction loss; 

\begin{itemize}
    \item Linear Layer ($n$, $h$); LeakyReLU($0.5$),
    \item Linear Layer ($h$, $h$); LeakyReLU($0.5$),
    \item Linear Layer ($h$, $d$),
\end{itemize}

where $n$ is the input data dimension and $h$ is hidden units and $h=200$ in all the experiments.  For the architecture for the decoder ($h$) in Table~\ref{table1_results}, Table~\ref{table2_results}, we use the polynomial decoder ($h(z) = H[1, z, z\bar{\otimes} z, \cdots, \underbrace{z \bar{\otimes} \cdots \bar{\otimes}\;z}_{p \;\text{times}}]^{\top}$); where $p$ is set to be same as that of the degree of true decoder polynomial ($g(z)$) and the coefficient matrix $H$ is modeled using a single fully connected layer. 

For the independence of support (IOS) experiments in Table~\ref{table1_results}, we model both $\Gamma, \Gamma^{'}$ using a single fully connected layer. 

For the interventional data results (Table~\ref{table2_results}), we learn the mappings $\gamma_i$ from the corresponding interventional data ($\mathbb{P}_X^{(i)}$) using the default linear regression class from scikit-learn~\citep{scikit-learn} with the intercept term turned off.

Finally, for the results with NN Decoder $h$ (Table~\ref{app:tab-obs-nn}, Table~\ref{app:tab-itv-nn}), we use the following architecture for the decoder with number of hidden nodes $h=200$.

\begin{itemize}
    \item Linear layer ($d$, $h$); LeakyReLU($0.5$)
    \item Linear layer ($h$, $h$); LeakyReLU($0.5$)
    \item Linear layer ($h$, $n$)
\end{itemize}

\paragraph{Hyperparameters.} We use the Adam optimizer with hyperparameters defined below. We also use early stopping strategy, where we halt the training process if the validation loss does not improve over 10 epochs consecutively.

\begin{itemize}
    \item Batch size: $16$
    \item Weight decay: $5\times 10^{-4}$
    \item Total epochs: $200$
    \item Learning rate: optimal value chosen from grid: $\{10^{-3}, 5\times 10^{-4}, 10^{-4}\}$
\end{itemize}

For experiments with independence of support (IOS) objective in Step 2 (Table~\ref{table1_results}), we train with $\lambda= 10$ as the relative weight of Hausdorff distance in the reconstruction loss (Equation~\ref{eqn: ae_hd}).

\subsection{Additional results: Polynomial decoder ($g$)}
\label{sec:res_poly_decoder_app}

Table~\ref{app:tab-obs-poly} presents additional details about Table~\ref{table1_results} in main paper.  We present additional metrics like mean squared loss for autoencoder reconstruction task (Recon-MSE) and MCC computed using representations from Step 1. Note that training with independence of support objective in Step 2 leads to better MCC scores than using the representations from Step 1 on distributions that satisfy independence of support. Also, the Uniform Correlated (Uniform-C) latent case can be interpreted as another sparse SCM with confounders between latent variables. For this case, the latent variables are not independent but their support is still independent, therefore we see improvement in MCC with IOS training in Step 2. Similarly, Table~\ref{app:tab-itv-poly} presents the extended results for the interventional case using polynomial decoder (Table~\ref{table2_results} in main paper); with additional metrics like mean squared loss for autoencoder reconstruction task (Recon-MSE) and $R^2$ to test for affine identification using representations from Step 1. We notice the same pattern for all latent distributions, that training on interventional data on Step 2 improves the MCC metric.

Further, we also experiment with using a neural network based decoder to have a more standard autoencoder architecture where we do not assume access to specific polynomial structure or the degree of the polynomial. Table~\ref{app:tab-obs-nn} presents the results with NN decoder for the observational case, where we see a similar trend to that of polynomial decoder case (Table~\ref{app:tab-obs-poly}) that the MCC increase with IOS training in Step 2 for Uniform and Uniform-C latent distributions. Similarly, Table~\ref{app:tab-itv-nn} presents the results with NN decoder for the interventional case, where the trend is similar to that of polynomial decoder case (Table~\ref{app:tab-itv-poly}); though the MCC (IL) for the SCM sparse and SCM dense case are lower compared to that with polynomial decoder case.

\begin{table}[!h]
\small
    \centering
\begin{tabular}{ccccccc}
\toprule
      $\mathbb{P}_Z$ &   $d$ &  $p$ &      Recon-MSE      &        $R^2$ &      MCC &    MCC (IOS)\\
\midrule

         Uniform &           $6$ &            $2$ &  $1.59 \pm 0.40$ &   $1.00 \pm 0.00$ & $66.91 \pm 2.45$ & $99.31 \pm 0.07$ \\
         Uniform &           $6$ &            $3$ &  $1.81 \pm 0.40$ &   $1.00 \pm 0.00$ & $75.14 \pm 3.93$ & $99.39 \pm 0.06$ \\
         Uniform &          $10$ &            $2$ & $2.04 \pm 0.76$ &   $1.00 \pm 0.00$ & $58.49 \pm 2.26$ & $90.73 \pm 2.92$ \\
         Uniform &          $10$ &            $3$ & $8.59 \pm 2.15$ &  $0.99 \pm 0.00$ &  $56.77 \pm 0.60$ &  $94.62 \pm 1.50$ \\  
         \midrule
       Uniform-C &           $6$ &            $2$ & $0.36 \pm 0.07$ &   $1.00 \pm 0.00$ & $71.19 \pm 2.29$ & $96.81 \pm 0.11$ \\
       Uniform-C &           $6$ &            $3$ & $1.72 \pm 0.67$ &   $1.00 \pm 0.00$ & $70.53 \pm 1.1$ &  $96.29 \pm 0.05$ \\
       Uniform-C &          $10$ &            $2$ & $0.86 \pm 0.27$ &   $1.00 \pm 0.00$ & $64.58 \pm 1.81$ & $85.31 \pm 2.35$ \\
       Uniform-C &          $10$ &            $3$ & $2.42 \pm 0.47$ &   $1.00 \pm 0.00$ & $62.69 \pm 0.92$ & $87.20 \pm 1.77$ \\
         \midrule
Gaussian-Mixture &           $6$ &            $2$ & $0.86 \pm 0.27$ &   $1.0 \pm 0.0$ & $70.53 \pm 1.25$ & $67.43 \pm 2.01$ \\
Gaussian-Mixture &           $6$ &            $3$ & $0.86 \pm 0.32$ &   $0.99 \pm 0.0$ & $66.19 \pm 1.38$ & $67.94 \pm 1.42$ \\
Gaussian-Mixture &          $10$ &            $2$ & $1.38 \pm 0.51$ &   $1.0 \pm 0.0$ &  $59.5 \pm 2.22$ &  $58.3 \pm 0.67$ \\
Gaussian-Mixture &          $10$ &            $3$ & $4.12 \pm 1.70$ &    $0.99 \pm 0.0$ & $57.15 \pm 0.43$ & $59.08 \pm 1.11$ \\
         \midrule
           SCM-S &           $6$ &            $2$ & $1.52 \pm 0.70$ & $0.96 \pm 0.02$ & $71.77 \pm 1.43$ & $72.61 \pm 1.48$ \\
           SCM-S &           $6$ &            $3$ & $2.25 \pm 0.51$ & $0.87 \pm 0.07$ & $73.14 \pm 3.44$ & $70.56 \pm 1.54$ \\
           SCM-S &          $10$ &            $2$ & $4.23 \pm 1.13$ & $0.99 \pm 0.0$ &  $64.35 \pm 2.0$ & $65.86 \pm 1.32$ \\
           SCM-S &          $10$ &            $3$ & $2.83 \pm 0.85$ & $0.90 \pm 0.05$ & $61.95 \pm 0.98$ & $58.77 \pm 1.27$ \\
         \midrule
           SCM-D &           $6$ &            $2$ & $1.34 \pm 0.26$ & $0.97 \pm 0.01$ & $75.25 \pm 2.85$ & $61.61 \pm 4.36$ \\
           SCM-D &           $6$ &            $3$ & $1.20 \pm 0.55$ & $0.81 \pm 0.11$ & $82.9 \pm 3.11$ &  $65.19 \pm 2.70$ \\
           SCM-D &          $10$ &            $2$ & $2.89 \pm 0.79$ & $0.83 \pm 0.10$ & $67.49 \pm 2.32$ & $69.64 \pm 3.09$ \\
           SCM-D &          $10$ &            $3$ & $1.55 \pm 0.39$ & $0.72 \pm 0.15$ & $66.4 \pm 1.86$ &  $60.1 \pm 1.16$ \\
\bottomrule
\end{tabular}

\caption{Observational data with Polynomial Decoder: Mean ± S.E. (5 random seeds). $R^2$ and MCC (IOS) achieve high values (for Uniform \& Uniform-C) as predicted \Cref{thm1} and \Cref{thm4} respectively. }
    \label{app:tab-obs-poly}
\end{table}

\begin{table}[!h]
\small
    \centering
\begin{tabular}{ccccccc}
\toprule
      $\mathbb{P}_Z$ &   $d$ &  $p$ &      Recon-MSE      &        $R^2$ &      MCC &    MCC (IL)\\
\midrule
         Uniform &           $6$ &            $2$ & $0.29 \pm 0.08$ &   $1.0 \pm 0.0$ & $69.11 \pm 1.11$ & $100.0 \pm 0.0$ \\
         Uniform &           $6$ &            $3$ & $0.97 \pm 0.36$ &   $1.0 \pm 0.0$ & $73.42 \pm 0.49$ & $100.0 \pm 0.0$ \\
         Uniform &          $10$ &            $2$ & $2.29 \pm 0.85$ &   $1.0 \pm 0.0$ & $59.96 \pm 2.03$ & $100.0 \pm 0.0$ \\
         Uniform &          $10$ &            $3$ & $2.74 \pm 0.36$ &   $1.0 \pm 0.0$ & $65.94 \pm 0.80$ & $99.85 \pm 0.03$ \\
        \midrule         
       Uniform-C &           $6$ &            $2$ & $0.29 \pm 0.11$ &   $1.0 \pm 0.0$ &  $71.2 \pm 2.46$ & $100.0 \pm 0.0$ \\
       Uniform-C &           $6$ &            $3$ & $1.50 \pm 0.62$ &   $1.0 \pm 0.0$ &  $70.21 \pm 1.90$ & $99.97 \pm 0.01$ \\
       Uniform-C &          $10$ &            $2$ & $0.79 \pm 0.24$ &   $1.0 \pm 0.0$ & $61.02 \pm 1.03$ & $100.0 \pm 0.0$ \\
       Uniform-C &          $10$ &            $3$ & $1.72 \pm 0.45$ &   $1.0 \pm 0.0$ & $61.16 \pm 1.59$ & $99.91 \pm 0.01$ \\
        \midrule                
Gaussian-Mixture &           $6$ &            $2$ & $0.75 \pm 0.27$ &  $1.0 \pm 0.0$ &  $67.72 \pm 2.20$ & $99.99 \pm 0.01$ \\
Gaussian-Mixture &           $6$ &            $3$ & $0.57 \pm 0.20$ &  $0.99 \pm 0.0$ & $70.21 \pm 2.74$ & $99.39 \pm 0.05$ \\
Gaussian-Mixture &          $10$ &            $2$ & $0.61 \pm 0.16$ &  $1.0 \pm 0.0$ &  $60.77 \pm 1.60$ & $99.98 \pm 0.01$ \\
Gaussian-Mixture &          $10$ &            $3$ & $2.29 \pm 0.72$ &  $0.99 \pm 0.0$ & $57.81 \pm 1.16$ & $99.46 \pm 0.05$ \\
        \midrule         
           SCM-S &           $6$ &            $2$ & $0.21 \pm 0.04$ &  $0.99 \pm 0.0$ &  $68.41 \pm 0.90$ & $99.53 \pm 0.38$ \\
           SCM-S &           $6$ &            $3$ & $0.93 \pm 0.18$ &  $0.99 \pm 0.0$ & $74.12 \pm 2.32$ & $99.25 \pm 0.34$ \\
           SCM-S &          $10$ &            $2$ & $0.63 \pm 0.17$ &  $1.0 \pm 0.0$ & $68.01 \pm 2.36$ & $99.92 \pm 0.03$ \\
           SCM-S &          $10$ &            $3$ & $1.29 \pm 0.31$ & $0.97 \pm 0.01$ &  $66.81 \pm 1.10$ &  $98.8 \pm 0.13$ \\
        \midrule         
           SCM-D &           $6$ &            $2$ & $0.81 \pm 0.05$ & $0.99 \pm 0.01$ &  $71.8 \pm 3.77$ & $99.64 \pm 0.12$ \\
           SCM-D &           $6$ &            $3$ & $0.75 \pm 0.26$ & $0.98 \pm 0.01$ & $79.48 \pm 3.45$ & $98.22 \pm 1.07$ \\
           SCM-D &          $10$ &            $2$ & $0.76 \pm 0.15$ & $0.98 \pm 0.01$ & $70.78 \pm 1.89$ & $95.3 \pm 2.24$ \\
           SCM-D &          $10$ &            $3$ & $0.96 \pm 0.22$ & $0.97 \pm 0.0$ &  $70.08 \pm 2.80$ & $97.24 \pm 0.88$ \\
\bottomrule
\end{tabular}
\caption{Interventional data with Polynomial Decoder:  Mean ± S.E. (5 random seeds). MCC(IL) is high as predicted by \Cref{thm2}.}
\label{app:tab-itv-poly}
\end{table}

\begin{table}[!h]
\small
    \centering
\begin{tabular}{ccccccc}
\toprule
      $\mathbb{P}_Z$ &   $d$ &  $p$ &      Recon-MSE      &        $R^2$ &      MCC &    MCC (IOS)\\
\midrule

         Uniform &           $6$ &            $2$ &  $1.22 \pm 0.19$ &  $0.98 \pm 0.0$ & $73.75 \pm 2.85$ & $99.05 \pm 0.02$ \\
         Uniform &           $6$ &            $3$ &  $2.79 \pm 0.20$ &  $0.92 \pm 0.0$ & $63.29 \pm 1.06$ & $95.74 \pm 0.12$ \\
         Uniform &          $10$ &            $2$ &  $3.66 \pm 0.39$ &  $0.99 \pm 0.0$ & $61.71 \pm 1.16$ & $94.25 \pm 2.13$ \\
         Uniform &          $10$ &            $3$ & $33.16 \pm 3.34$ &  $0.94 \pm 0.0$ & $59.27 \pm 1.06$ & $91.24 \pm 4.99$ \\
         \midrule
       Uniform-C &           $6$ &            $2$ &   $0.65 \pm 0.10$ & $0.96 \pm 0.02$ & $68.46 \pm 1.94$ & $94.95 \pm 1.83$ \\
       Uniform-C &           $6$ &            $3$ &   $1.39 \pm 0.30$ &  $0.91 \pm 0.0$ & $68.09 \pm 1.56$ & $89.14 \pm 2.38$ \\
       Uniform-C &          $10$ &            $2$ &  $1.78 \pm 0.09$ &  $0.99 \pm 0.0$ & $62.63 \pm 2.05$ & $88.88 \pm 3.28$ \\
       Uniform-C &          $10$ &            $3$ &  $12.0 \pm 1.59$ &  $0.91 \pm 0.01$ & $59.91 \pm 1.75$ & $81.76 \pm 3.67$ \\
         \midrule
Gaussian-Mixture &           $6$ &            $2$ &  $0.49 \pm 0.12$ &  $0.95 \pm 0.0$ & $72.59 \pm 2.03$ & $65.33 \pm 1.11$ \\
Gaussian-Mixture &           $6$ &            $3$ &  $0.79 \pm 0.16$ &  $0.84 \pm 0.01$ & $66.25 \pm 2.86$ & $63.43 \pm 1.27$ \\
Gaussian-Mixture &          $10$ &            $2$ &  $1.38 \pm 0.18$ &  $0.95 \pm 0.0$ & $57.12 \pm 1.52$ & $54.76 \pm 1.26$ \\
Gaussian-Mixture &          $10$ &            $3$ &  $7.22 \pm 1.23$ &  $0.83 \pm 0.01$ & $55.41 \pm 1.40$ & $52.87 \pm 0.86$ \\
         \midrule
           SCM-S &           $6$ &            $2$ &  $2.24 \pm 1.11$ & $0.59 \pm 0.18$ & $69.77 \pm 3.87$ & $66.04 \pm 1.34$ \\
           SCM-S &           $6$ &            $3$ &  $2.45 \pm 0.18$ & $0.74 \pm 0.05$ & $73.72 \pm 1.63$ & $67.66 \pm 2.18$ \\
           SCM-S &          $10$ &            $2$ &  $6.41 \pm 1.71$ & $0.78 \pm 0.08$ & $65.99 \pm 1.14$ & $63.52 \pm 1.11$ \\
           SCM-S &          $10$ &            $3$ &  $4.32 \pm 1.37$ & $0.11 \pm 0.43$ & $66.96 \pm 2.60$ & $62.11 \pm 1.36$ \\
         \midrule
           SCM-D &           $6$ &            $2$ &   $2.7 \pm 0.39$ & $0.63 \pm 0.22$ & $75.19 \pm 2.62$ & $61.89 \pm 4.0$ \\
           SCM-D &           $6$ &            $3$ &  $1.89 \pm 0.73$ & $0.47 \pm 0.25$ & $77.83 \pm 3.49$ & $65.85 \pm 1.58$ \\
           SCM-D &          $10$ &            $2$ &  $4.46 \pm 0.76$ & $0.46 \pm 0.11$ & $69.81 \pm 1.43$ & $65.35 \pm 2.72$ \\
           SCM-D &          $10$ &            $3$ &  $3.53 \pm 0.69$ & $0.10 \pm 0.29$ & $65.89 \pm 2.56$ & $61.92 \pm 1.95$ \\
\bottomrule
\end{tabular}

\caption{Observational data with Neural Network Decoder: Mean ± S.E. (5 random seeds). $R^2$ achieves high values in many cases but  MCC (IOS) achieve high values (for Uniform \& Uniform-C). }
    \label{app:tab-obs-nn}
\end{table}

\begin{table}[!h]
\small
    \centering
\begin{tabular}{ccccccc}
\toprule
      $\mathbb{P}_Z$ &   $d$ &  $p$ &      Recon-MSE      &        $R^2$ &      MCC &    MCC (IL)\\
\midrule

         Uniform &           $6$ &            $2$ &  $0.35 \pm 0.08$ &  $0.98 \pm 0.0$ & $68.39 \pm 1.21$ & $99.09 \pm 0.02$ \\
         Uniform &           $6$ &            $3$ &  $2.02 \pm 0.28$ &  $0.91 \pm 0.0$ &  $63.2 \pm 1.33$ & $91.67 \pm 2.50$ \\
         Uniform &          $10$ &            $2$ &  $3.89 \pm 0.50$ &  $0.99 \pm 0.0$ & $60.54 \pm 1.81$ & $99.59 \pm 0.04$ \\
         Uniform &          $10$ &            $3$ & $29.21 \pm 2.33$ &  $0.95 \pm 0.0$ & $61.0 \pm 1.48$ & $93.73 \pm 0.45$ \\
        \midrule         
       Uniform-C &           $6$ &            $2$ &  $0.42 \pm 0.15$ &  $0.94 \pm 0.02$ & $65.91 \pm 0.53$ &  $96.43 \pm 1.47$ \\
       Uniform-C &           $6$ &            $3$ &  $1.05 \pm 0.19$ &  $0.91 \pm 0.0$ &  $67.92 \pm 3.48$ &  $94.8 \pm 0.28$ \\
       Uniform-C &          $10$ &            $2$ &  $1.32 \pm 0.09$ &  $0.99 \pm 0.0$ &  $60.02 \pm 1.83$ &  $99.42 \pm 0.01$ \\
       Uniform-C &          $10$ &            $3$ & $10.46 \pm 1.27$ &  $0.92 \pm 0.0$ &  $61.68 \pm 1.20$ &  $93.83 \pm 0.78$ \\
        \midrule         
Gaussian-Mixture &           $6$ &            $2$ &  $0.45 \pm 0.13$ &  $0.94 \pm 0.0$ & $70.64 \pm 3.83$ & $96.87 \pm 0.14$ \\
Gaussian-Mixture &           $6$ &            $3$ &  $0.62 \pm 0.12$ &  $0.83 \pm 0.01$ & $64.43 \pm 2.36$ & $84.53 \pm 2.60$ \\
Gaussian-Mixture &          $10$ &            $2$ &  $0.87 \pm 0.15$ &  $0.94 \pm 0.0$ & $57.35 \pm 1.62$ & $97.06 \pm 0.16$ \\
Gaussian-Mixture &          $10$ &            $3$ &  $5.98 \pm 0.93$ &  $0.83 \pm 0.0$ & $57.89 \pm 2.06$ & $80.14 \pm 1.77$ \\
        \midrule         
           SCM-S &           $6$ &            $2$ &  $0.27 \pm 0.07$ & $0.94 \pm 0.02$ & $74.68 \pm 2.28$ & $93.07 \pm 2.16$ \\
           SCM-S &           $6$ &            $3$ &   $0.9 \pm 0.18$ & $0.89 \pm 0.02$ & $71.56 \pm 3.18$ & $88.66 \pm 2.71$ \\
           SCM-S &          $10$ &            $2$ &  $0.93 \pm 0.23$ & $0.98 \pm 0.0$ &  $66.08 \pm 1.04$ & $94.14 \pm 0.39$ \\
           SCM-S &          $10$ &            $3$ &  $1.99 \pm 0.36$ & $0.88 \pm 0.01$ & $63.35 \pm 1.44$ & $76.62 \pm 6.15$ \\
        \midrule         
           SCM-D &           $6$ &            $2$ &  $0.69 \pm 0.07$ & $0.95 \pm 0.02$ & $76.99 \pm 2.53$ &  $91.63 \pm 1.90$ \\
           SCM-D &           $6$ &            $3$ &  $0.87 \pm 0.25$ & $0.88 \pm 0.01$ & $75.72 \pm 1.69$ &  $88.19 \pm 3.63$ \\
           SCM-D &          $10$ &            $2$ &  $1.05 \pm 0.29$ & $0.95 \pm 0.01$ & $68.71 \pm 2.16$ &  $90.14 \pm 4.35$ \\
           SCM-D &          $10$ &            $3$ &  $1.68 \pm 0.34$ & $0.86 \pm 0.01$ & $68.52 \pm 2.11$ &  $81.82 \pm 3.0$ \\

\bottomrule
\end{tabular}
\caption{Interventional data with Neural Network Decoder:  Mean ± S.E. (5 random seeds). MCC(IL) is high.}
\label{app:tab-itv-nn}
\end{table}

\clearpage 

\subsection{Experiment setup details: Synthetic image experiments }
\label{sec:setup_images_app}

The latent variable comprises of two balls and their ($x$, $y$) coordinates; hence we have $d=4$ dimensional latent variable. We use PyGame~\citep{pygame} rendering engine final images of dimension $64 \times 64 \times 3$.

\paragraph{Latent Distributions.} We denote the $(x, y)$ coordinates of the Ball 1 as ($x_1$, $y_1$), and for Ball 2 as ($x_2$, $y_2$). We have the following three cases for the latent distributions in case of synthetic image experiments:

\begin{itemize}
    \item \textbf{Uniform:} Each coordinate of Ball 1 ($x_1$, $y_1$) and Ball 2 ($x_2$, $y_2$) are sampled from $\text{Uniform}(0.1, 0.9)$. 
    
    \item \textbf{SCM (linear):} The coordinates of Ball 1 ($x_1$, $y_1$) are sampled from $\text{Uniform}(0.1, 0.9)$, which are used to sample the coordinates of Ball 2 as follows:
    $$     
    x_2 \sim \begin{cases}
        \text{Uniform}(0.1, 0.5) & \text{if} \; x_1 + y_1 \geq 1.0 \\
        \text{Uniform}(0.5, 0.9) & \text{if} \; x_1 + y_1 < 1.0 
    \end{cases} 
    $$

    $$ 
    y_2 \sim \begin{cases}
        \text{Uniform}(0.5, 0.9) & \text{if} \; x_1 + y_1 \geq 1.0 \\
        \text{Uniform}(0.1, 0.5) & \text{if} \; x_1 + y_1 < 1.0 
    \end{cases} 
    $$

    \item \textbf{SCM (non-linear):} The coordinates of Ball 1 ($x_1$, $y_1$) are sampled from $\text{Uniform}(0.1, 0.9)$, which are used to sample the coordinates of Ball 2 as follows:
    $$     
    x_2 \sim \begin{cases}
        \text{Uniform}(0.1, 0.5) & \text{if} \; 1.25 \times  (x_1^2 + y_1^2 ) \geq 1.0 \\
        \text{Uniform}(0.5, 0.9) & \text{if} \; 1.25 \times (x_1^2 + y_1^2 ) < 1.0 
    \end{cases} 
    $$

    $$ 
    y_2 \sim \begin{cases}
        \text{Uniform}(0.5, 0.9) & \text{if} \; 1.25  \times (x_1^2 + y_1^2 ) \geq 1.0 \\
        \text{Uniform}(0.1, 0.5) & \text{if} \; 1.25 \times  (x_1^2 + y_1^2 )  < 1.0 
    \end{cases} 
    $$

\end{itemize}

\begin{table}[h]
    \centering
    \begin{tabular}{c|c|c|c}
        \toprule
        Case & Train & Validation & Test \\
        \midrule
        Observational ($\mathcal{D}$) & 20000 & 5000 & 20000  \\
        Interventional ($\mathcal{D}^{(I)}$) & 20000 & 5000 & 20000 \\
        \bottomrule
    \end{tabular}
    \caption{Statistics for the synthetic image experiments}
    \label{tab:data-stats-syn-image}
\end{table}

\paragraph{Further details on dataset and evaluation.} For experiments in Table~\ref{table4_results}, the details regarding the train/val/test split are described in Table~\ref{tab:data-stats-syn-image}. 

Note that the interventional data ($\mathcal{D}^{(I)}$) is composed of do interventions on each latent variable ($\mathcal{D}^{(I)}= \cup_{i=1:d} \mathcal{D}^{i}$), where latent variable to be intervened is sampled from $\text{Uniform}(\{1, \cdots, d\})$. Hence, each latent variable has equal probability to be intervened.

While performing do-interventions on any latent variable ($\mathcal{D}^{(i)}$), we control for the total number of distinct values the latent takes under the intervention (\#interv, each distinct value correpsonds to sampling data from one interventional distribution). When \#interv $=1$, then we set the latent variable $i$ to value 0.5. For the case when \#interv $> 1$, we sample the values corresponding to different do-interventions on latent variable $i$ as total of \#interv equally distant points from $S= [0.25, 0.75]$. Eg, when \#interv $=3$, then the possible values after do-intervention on latent variable $i$ are $\{0.25, 0.50, 0.75\}$. Note that we uniformly at random sample the value of intervention  from the set of intervention values.

Note that we only use the observational data ($\mathcal{D}$) for training the autoencoder in Step 1. while the non-linear transformations $\gamma_i$ in Step 2 (\Cref{eqn: cons_int}) are learnt using the corresponding interventional data ($\mathcal{D}^{(i)}$). Further, the metrics (MCC, MCC (IL)) are computed only on the test split of observational data ($\mathcal{D}$) (no interventional data used).

\paragraph{Model architecture.} We use the following architecture for encoder $f$ across all experiments (Table~\ref{table4_results}) in Step 1 of minimizing the reconstruction loss.

\begin{itemize}
    \item ResNet-18 Architecture (No Pre Training): Image ($64 \times 64 \times 3$) $\rightarrow$ Penultimate Layer Output ($512$ dimensional)
    \item Linear Layer $(512, 128)$; BatchNorm();  LeakyReLU()
    \item Linear Layer $(128, 25)$; BatchNorm()
\end{itemize}

We use the following architecture for decoder $h$ across all experiments (Table~\ref{table4_results}) in Step 1 of minimizing the reconstruction loss. Our architecture for decoder is inspired from the implementation in widely used works~\citep{locatello2019challenging}.

\begin{itemize}
    \item Linear Layer $(25, 128)$; LeakyReLU()
    \item Linear Layer $(128, 1024)$; LeakyReLU()
    \item DeConvolution Layer ($c_{in}$: $64$, $c_{out}$: $64$, kernel: $4$; stride: $2$; padding: $1$); LeakyReLU()
    \item DeConvolution Layer ($c_{in}$: $64$, $c_{out}$: $32$, kernel: $4$; stride: $2$; padding: $1$); LeakyReLU()
    \item DeConvolution Layer ($c_{in}$: $32$, $c_{out}$: $32$, kernel: $4$; stride: $2$; padding: $1$); LeakyReLU()
    \item DeConvolution Layer ($c_{in}$: $32$, $c_{out}$: $3$, kernel: $4$; stride: $2$; padding: $1$); LeakyReLU()
\end{itemize}

\textbf{Note:} Here the latent dimension of the encoder ($25$) is not equal to the true latent dimension ($d=4$) as that would lead issues with training the autoencoder itself. Also, this choice is more suited towards practical scenarios where we do not know the dimension of latent beforehand. 

For learning the mappings $\gamma_i$ from the corresponding interventional data ($\mathbb{P}_X^{(i)}$), we use the default MLP Regressor class from scikit-learn~\citep{scikit-learn} with 1000 max iterations for convergence.

\paragraph{Hyperparameters.} We use Adam optimizer with hyperparameters defined below. We also use early stopping strategy, where we halt the training process if the validation loss does not improve over 100 epochs consecutively.

\begin{itemize}
    \item Batch size: $64$
    \item Weight decay: $5 \times 10^{-4}$
    \item Total epochs: $1000$
    \item Learning rate: $5\times 10^{-4}$
\end{itemize}

\subsection{Additional Results: Synthetic Image Experiments}
\label{sec:results_images_app}

\begin{table}[t]
\small
    \centering
\begin{tabular}{ccccccc}
\toprule
      $\mathbb{P}_Z$ &   \#interv &      Recon-RMSE      &        $R^2$ &  MCC (IL)\\
\midrule

     Uniform &                    $1$ & $0.04 \pm 0.0$ & $0.51 \pm 0.0$ & $34.18 \pm 0.24$ \\
     Uniform &                    $3$ & $0.04 \pm 0.0$ & $0.51 \pm 0.0$ & $73.94 \pm 0.38$ \\
     Uniform &                    $5$ & $0.04 \pm 0.0$ & $0.51 \pm 0.0$ & $73.62 \pm 0.21$ \\
     Uniform &                    $7$ & $0.04 \pm 0.0$ & $0.51 \pm 0.0$ & $72.54 \pm 0.34$ \\
     Uniform &                    $9$ & $0.04 \pm 0.0$ & $0.51 \pm 0.0$ & $73.14 \pm 0.47$ \\
    \midrule
    SCM (linear) &                    $1$ & $0.03 \pm 0.0$ &  $0.8 \pm 0.0$ & $12.81 \pm 0.28$ \\
    SCM (linear) &                    $3$ & $0.03 \pm 0.0$ &  $0.8 \pm 0.0$ & $73.21 \pm 0.33$ \\
    SCM (linear) &                    $5$ & $0.03 \pm 0.0$ &  $0.8 \pm 0.0$ & $83.38 \pm 0.21$ \\
    SCM (linear) &                    $7$ & $0.03 \pm 0.0$ &  $0.8 \pm 0.0$ & $84.22 \pm 0.25$ \\
    SCM (linear) &                    $9$ & $0.03 \pm 0.0$ &  $0.8 \pm 0.0$ & $86.16 \pm 0.17$ \\
    \midrule
    SCM (non-linear) &                    $1$ & $0.04 \pm 0.0$ & $0.69 \pm 0.0$ &  $19.70 \pm 0.31$ \\
    SCM (non-linear) &                    $3$ & $0.04 \pm 0.0$ & $0.69 \pm 0.0$ & $59.68 \pm 0.28$ \\
    SCM (non-linear) &                    $5$ & $0.04 \pm 0.0$ & $0.69 \pm 0.0$ &  $62.79 \pm 0.20$ \\
    SCM (non-linear) &                    $7$ & $0.04 \pm 0.0$ & $0.69 \pm 0.0$ & $69.31 \pm 0.34$ \\
    SCM (non-linear) &                    $9$ & $0.04 \pm 0.0$ & $0.69 \pm 0.0$ & $71.37 \pm 0.26$ \\
    
\bottomrule
\end{tabular}
\caption{Interventional data in image-based experiments: Mean ± S.E. (5 random seeds). MCCs increase with the number of interventions per latent dimension as predicted by \Cref{thm_do_appendix}.}
\label{app:tab-image-full}
\end{table}

Table~\ref{app:tab-image-full} presents more details about Table~\ref{table4_results} in the main paper, with additional metrics like mean squared loss for autoencoder reconstruction task (Recon-MSE) and and $R^2$ to test for affine identification using representations from Step 1. Note that Recon-RMSE and $R^2$ are computed using the autoencoder trained from Step 1, hence the results are not affected by training on varying \#interv per latent in Step 2. We get high $R^2$ values across different latent distributions indicating the higher dimensional latents ($\hat{d}=25$) learned by the encoder are related to the small dimensional true latents ($d=4$) by a linear function.

\begin{figure}[t]
\centering
    \includegraphics[scale=0.7]{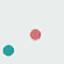}
    \quad
    \includegraphics[scale=0.7]{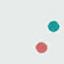}
    \quad
    \includegraphics[scale=0.7]{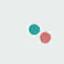} 
    \quad
    \includegraphics[scale=0.7]{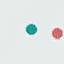} 
    \quad
    \includegraphics[scale=0.7]{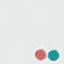} 
    \\
    \includegraphics[scale=0.7]{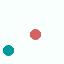} 
    \quad
    \includegraphics[scale=0.7]{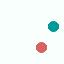} 
    \quad
    \includegraphics[scale=0.7]{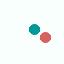} 
   \quad
    \includegraphics[scale=0.7]{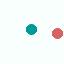} 
    \quad
    \includegraphics[scale=0.7]{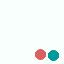} 
    \caption{Reconstructed images (top row) for the corresponding real images (bottom row) for the uniform latent case.}
    \label{fig:recon-uniform}
\end{figure}

\begin{figure}[th]
    \centering
    \includegraphics[scale=0.7]{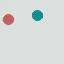}
    \quad
    \includegraphics[scale=0.7]{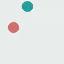}
    \quad
    \includegraphics[scale=0.7]{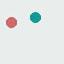} 
    \quad
    \includegraphics[scale=0.7]{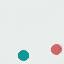} 
    \quad
    \includegraphics[scale=0.7]{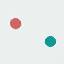} 
    \\
    \includegraphics[scale=0.7]{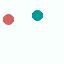} 
    \quad
    \includegraphics[scale=0.7]{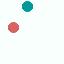} 
    \quad
    \includegraphics[scale=0.7]{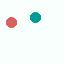} 
   \quad
    \includegraphics[scale=0.7]{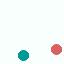} 
    \quad
    \includegraphics[scale=0.7]{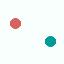} 
    \caption{Reconstructed images (top row) for the corresponding real images (bottom row) for the SCM (linear) latent case.}
    \label{fig:recon-scm-linear}
\end{figure}

\begin{figure}[h]
    \centering
    \includegraphics[scale=0.7]{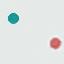}
    \quad
    \includegraphics[scale=0.7]{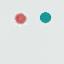}
    \quad
    \includegraphics[scale=0.7]{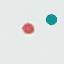} 
    \quad
    \includegraphics[scale=0.7]{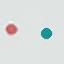} 
    \quad
    \includegraphics[scale=0.7]{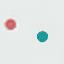} 
    \\
    \includegraphics[scale=0.7]{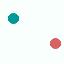} 
    \quad
    \includegraphics[scale=0.7]{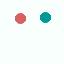} 
    \quad
    \includegraphics[scale=0.7]{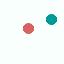} 
   \quad
    \includegraphics[scale=0.7]{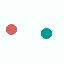} 
    \quad
    \includegraphics[scale=0.7]{real_image_3_nlin.jpg} 
     \\
    \caption{Reconstructed images (top row) for the corresponding real images (bottom row) for the SCM (non-linear) latent case.}
    \label{fig:recon-scm-non-linear}
\end{figure}

We also report a batch of reconstructed images from the trained autoencoder for the different latent distributions; Uniform (Figure~\ref{fig:recon-uniform}), SCM Linear (Figure~\ref{fig:recon-scm-linear}), and SCM Non-Linear (Figure~\ref{fig:recon-scm-non-linear}). In all the cases the position and color of both the balls is accurately reconstructed.

\clearpage

\subsection{Experiments with independence penalty from $\beta$-VAE}
In this section, we provide some additional comparisons with models trained with independence prior on the latents  used in $\beta$-VAEs \citep{burgess2018understanding}. We take a standard autoencoder that uses a reconstruction penalty and add to it the $\beta$-VAE penalty. We carry out the comparisons for both polynomial data generation experiments and also for the image-based experiments. For the polynomial data generation experiments, we  use the same MLP based encoder-decoder architecture that we used earlier for  Table~\ref{app:tab-obs-nn} and Table~\ref{app:tab-itv-nn}. In Table~\ref{beta-ae1} and Table~\ref{beta-ae2}, we show the results for autoencoder trained with $\beta$-VAE penalty for the  same setting as was used in Table~\ref{app:tab-obs-nn} and Table~\ref{app:tab-itv-nn}  respectively.
For the image-based experiments, we use the same ResNet-based encoder-decoder architecture that we used earlier for Table~\ref{table4_results}.   In Table~\ref{beta-ae3}, we show results for the image-based experiments using the same setting as Table~\ref{table4_results} focusing on the case with nine interventions.  

\begin{table}[!h]
\small
    \centering
\begin{tabular}{ccccccc}
\toprule
      $\mathbb{P}_Z$ &   $d$ &  $p$ &    MCC ($\beta=0.1$) &  MCC ($\beta=1.0$) &   MCC ($\beta=10.0$) &  MCC (IOS)\\
\midrule

         Uniform &           $6$ &            $2$ &  $67.35 \pm 2.7$ & $68.73 \pm 2.88 $  & $72.38 \pm 3.4 $ & $99.05 \pm 0.02$ \\
         Uniform &           $6$ &            $3$ &  $70.98 \pm 2.57$ & $69.43 \pm 1.82$  & $71.46 \pm 3.13$  & $95.74 \pm 0.12$ \\
         Uniform &          $10$ &            $2$ &  $58.94 \pm 2.04$ & $57.8 \pm 1.35$ & $60.14 \pm 1.33 $ & $94.25 \pm 2.13$ \\
         Uniform &          $10$ &            $3$ &  $59.29 \pm 2.45$ & $60.94 \pm 2.17$  & $59.22 \pm 1.24 $ & $91.24 \pm 4.99$ \\
         \midrule
           SCM-S &           $6$ &            $2$ & $65.37 \pm 2.28$  & $61.98 \pm 4.52$  & $65.63 \pm 4.15$  & $66.04 \pm 1.34$ \\
           SCM-S &           $6$ &            $3$ & $64.53 \pm 2.38 $ & $65.23 \pm 0.99$  & $68.61 \pm 2.74$ & $67.66 \pm 2.18$ \\
           SCM-S &          $10$ &            $2$ & $62.54 \pm 1.33$  & $62.23 \pm 1.65$  & $63.43 \pm 0.87$ & $63.52 \pm 1.11$ \\
           SCM-S &          $10$ &            $3$ & $62.44 \pm 0.56 $ & $58.04 \pm 1.64 $ & $59.5 \pm 0.89$ & $62.11 \pm 1.36$ \\
         \midrule
           SCM-D &           $6$ &            $2$ & $60.86 \pm 2.63$ & $59.36 \pm 2.71$ & $62.26 \pm 1.66$ & $61.89 \pm 4.0$ \\
           SCM-D &           $6$ &            $3$ & $66.32 \pm 1.36$ & $65.49 \pm 1.97$ & $66.43 \pm 1.4$ & $65.85 \pm 1.58$ \\
           SCM-D &          $10$ &            $2$ & $61.13 \pm 1.42$ & $62.23 \pm 1.32$ & $61.38 \pm 2.25$ & $65.35 \pm 2.72$ \\
           SCM-D &          $10$ &            $3$ & $60.39 \pm 1.83$ & $58.64 \pm 2.01 $ & $58.43 \pm 1.08$ & $61.92 \pm 1.95$ \\
\bottomrule
\end{tabular}
\caption{Observational data with Neural Network Decoder: Mean ± S.E. (5 random seeds). }
\label{beta-ae1}
\end{table}

\begin{table}[!h]
\small
    \centering
\begin{tabular}{ccccccc}
\toprule
      $\mathbb{P}_Z$ &   $d$ &  $p$ &    MCC ($\beta=0.1$) &  MCC ($\beta=1.0$) &   MCC ($\beta=10.0$) &    MCC (IL)\\
\midrule

         Uniform &           $6$ &            $2$ & $69.79 \pm 1.83$  & $68.62 \pm 2.99$  & $69.59 \pm 1.76 $ & $99.09 \pm 0.02$ \\
         Uniform &           $6$ &            $3$ &  $68.44 \pm 1.88$ & $71.86 \pm 0.42$ & $68.76 \pm 2.13$ & $91.67 \pm 2.50$ \\
         Uniform &          $10$ &            $2$ & $60.46 \pm 1.46$  & $58.93 \pm 0.91 $  & $58.95 \pm 0.89$ & $99.59 \pm 0.04$ \\
         Uniform &          $10$ &            $3$ &  $59.85 \pm 1.46$ & $62.92 \pm 2.98$  & $ 61.34 \pm 1.66$ & $93.73 \pm 0.45$ \\
        \midrule         
           SCM-S &           $6$ &            $2$ & $71.18 \pm 2.11$  & $71.01 \pm 1.32$ & $67.6 \pm 2.07$ & $93.07 \pm 2.16$ \\
           SCM-S &           $6$ &            $3$ &  $72.67 \pm 1.29$ & $70.9 \pm 3.63$ & $75.6 \pm 3.04$ & $88.66 \pm 2.71$ \\
           SCM-S &          $10$ &            $2$ &  $65.04 \pm 1.46 $ & $64.47 \pm 1.49$ & $65.84 \pm 2.1$ & $94.14 \pm 0.39$ \\
           SCM-S &          $10$ &            $3$ &  $63.2 \pm 1.83$ & $62.31 \pm 1.1$ & $62.16 \pm 1.68$ & $76.62 \pm 6.15$ \\
        \midrule         
           SCM-D &           $6$ &            $2$ & $72.6 \pm 2.01 $ & $76.58 \pm 2.95$ & $71.92 \pm 2.85 $ &  $91.63 \pm 1.90$ \\
           SCM-D &           $6$ &            $3$ & $67.79 \pm 0.97$ & $72.93 \pm 1.81 $ & $ 72.98 \pm 1.27$ &  $88.19 \pm 3.63$ \\
           SCM-D &          $10$ &            $2$ & $69.78 \pm 3.85 $ & $69.19 \pm 2.78$ & $66.53 \pm 1.28$ &  $90.14 \pm 4.35$ \\
           SCM-D &          $10$ &            $3$ & $64.9 \pm 1.68 $ & $66.86 \pm 2.61$ & $64.25 \pm 1.61 $ &  $81.82 \pm 3.0$ \\

\bottomrule
\end{tabular}
\caption{Interventional data with Neural Network Decoder:  Mean ± S.E. (5 random seeds).}
\label{beta-ae2}
\end{table}

\begin{table}[t]
\small
    \centering
\begin{tabular}{ccccccc}
\toprule
      $\mathbb{P}_Z$ &  MCC ($\beta= 0.1$)  &     MCC ($\beta= 1.0$)      &    MCC ($\beta= 10.0$)   &  MCC (IL)\\
\midrule
    Uniform &     $42.6 \pm 4.23$ &  $36.5 \pm 2.45$ &   $38.3 \pm 2.22$   & $73.1 \pm 0.47$ \\
     \midrule
    SCM (linear) &     $60.8 \pm 2.52$ &  $59.5 \pm 2.47$   &   $61.6 \pm 1.06$   & $86.2 \pm 0.17$ \\
    \midrule
    SCM (non-linear) &  $62.5 \pm 1.88$  &  $60.7 \pm 2.41$   &   $59.7 \pm 1.29$   & $71.4 \pm 0.26$ \\

\bottomrule
\end{tabular}
\caption{Interventional data in image-based experiments: Mean ± S.E. (5 random seeds).}
\label{beta-ae3}
\end{table}

\end{document}